\documentclass[10pt,twocolumn,letterpaper]{article}

\usepackage{iccv}
\usepackage{times}
\usepackage{epsfig}
\usepackage{graphicx}
\usepackage{amsmath}
\usepackage{amssymb}
\usepackage{bbm}
\usepackage{booktabs}
\usepackage{xcolor}
\usepackage{multirow}
\usepackage{multicol}
\usepackage{makecell}
\usepackage[percent]{overpic}
\usepackage{subcaption}
\usepackage{pifont}

\usepackage[pagebackref=true,breaklinks=true,letterpaper=true,colorlinks,bookmarks=false]{hyperref}

\usepackage[capitalize]{cleveref}
\crefname{section}{Sec.}{Secs.}
\Crefname{section}{Section}{Sections}
\Crefname{table}{Table}{Tables}
\crefname{table}{Tab.}{Tabs.}

\iccvfinalcopy 

\def\iccvPaperID{10776} 

\begin{document}

\title{GlueStick: Robust Image Matching by Sticking Points and Lines Together}


\author{Rémi Pautrat\footnotemark~~${}^1$
\and
Iago Su\'arez\footnotemark[\value{footnote}]~~${}^2$
\and
Yifan Yu${}^1$
\and
Marc Pollefeys${}^{1, 3}$
\and
Viktor Larsson${}^4$
\and
${}^1$ \normalsize{Department of Computer Science, ETH Zurich}
\and
${}^2$ \normalsize{Qualcomm XR Labs Europe}
\and
${}^3$ \normalsize{Microsoft Mixed Reality and AI Zurich lab}
\and
${}^4$ \normalsize{Lund University}
}

\maketitle
\ificcvfinal\thispagestyle{empty}\fi

\footnotetext{* Authors contributed equally.}

\begin{abstract}
   Line segments are powerful features complementary to points. They offer structural cues, robust to drastic viewpoint and illumination changes, and can be present even in texture-less areas. However, describing and matching them is more challenging compared to points due to partial occlusions, lack of texture, or repetitiveness.  
   This paper introduces a new matching paradigm, where points, lines, and their descriptors are unified into a single wireframe structure. We propose GlueStick, a deep matching Graph Neural Network (GNN) that takes two wireframes from different images and leverages the connectivity information between nodes to better glue them together.
   In addition to the increased efficiency brought by the joint matching, we also demonstrate a large boost of performance when leveraging the complementary nature of these two features in a single architecture.
   We show that our matching strategy outperforms the state-of-the-art approaches independently matching line segments and points for a wide variety of datasets and tasks.
   The code is available at \url{https://github.com/cvg/GlueStick}.
\end{abstract}
\begin{figure}
    \centering
    \small
    \newcommand{\sz}{0.9}
    \begin{tabular}{cc}
        \rotatebox{90}{(a) SuperGlue~\cite{sarlin_2020_superglue}} & \includegraphics[width=\sz\columnwidth]{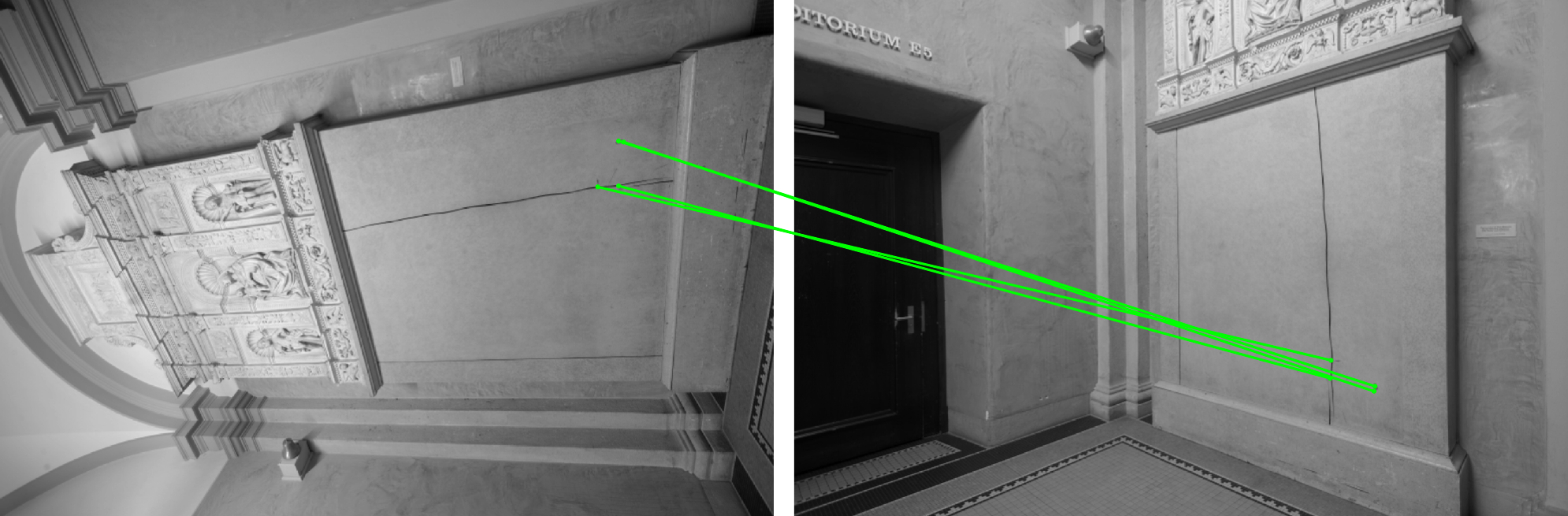} \\
        \rotatebox{90}{\phantom{xxxx}(b) Ours} & \includegraphics[width=\sz\columnwidth]{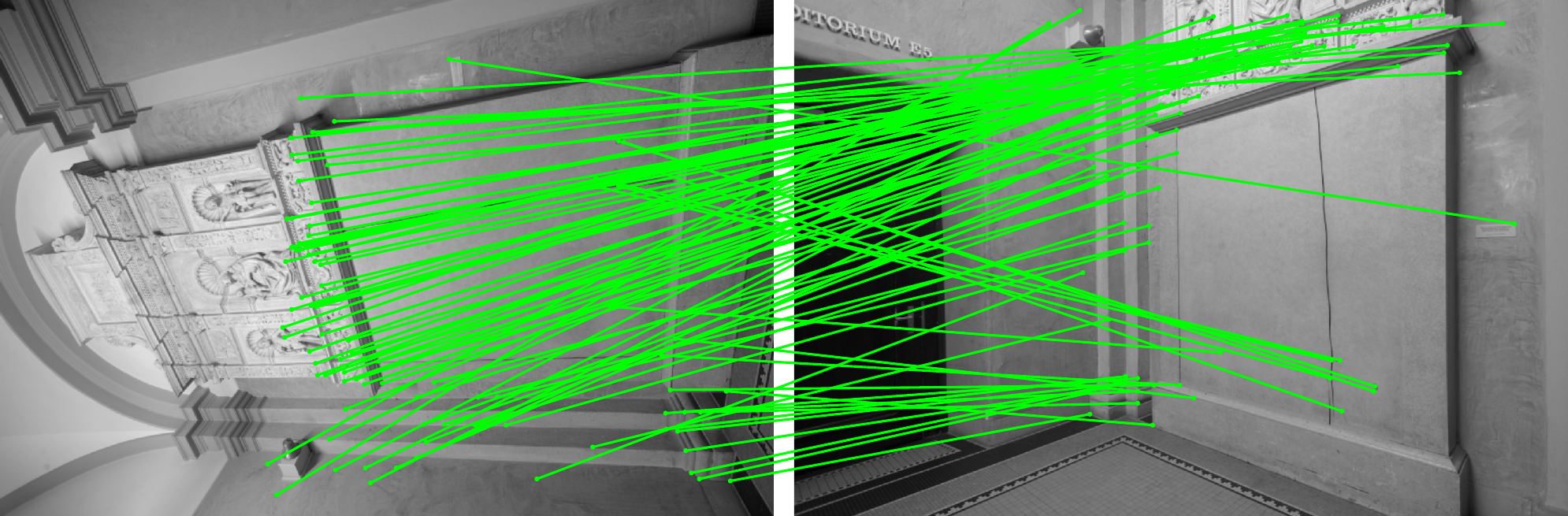} \\
        \rotatebox{90}{\phantom{x}(c) LineTR~\cite{syoon_2021_linetr}} & \includegraphics[width=\sz\columnwidth]{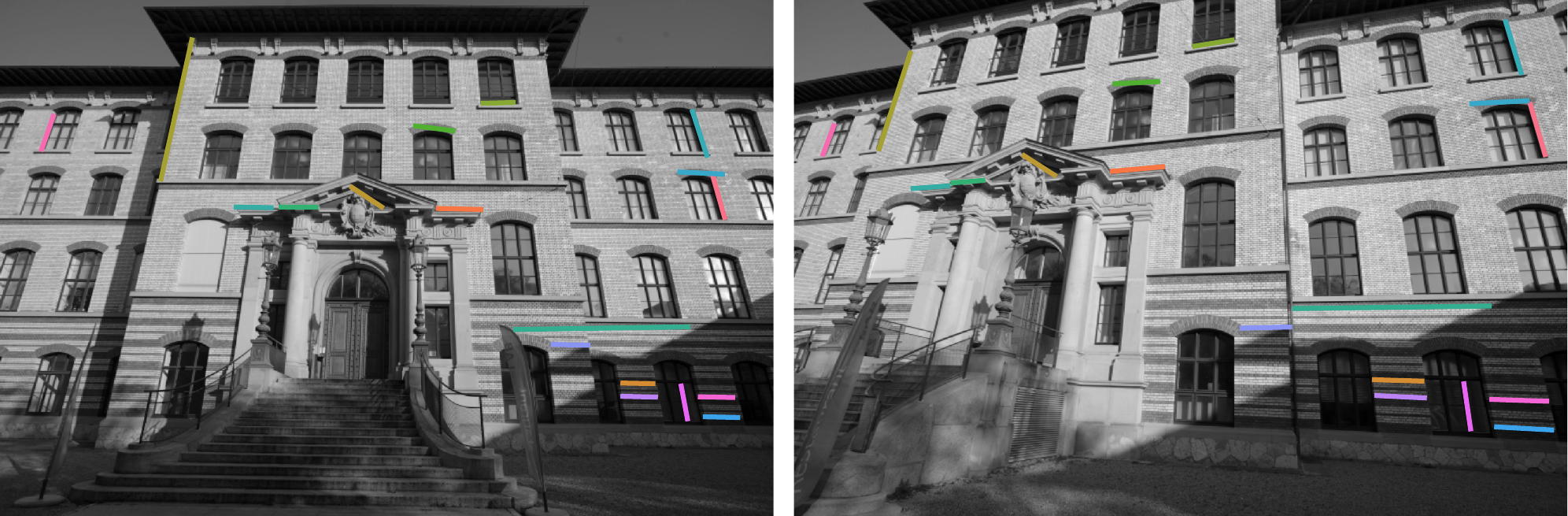} \\
        \rotatebox{90}{\phantom{xxxx}(d) Ours} & \includegraphics[width=\sz\columnwidth]{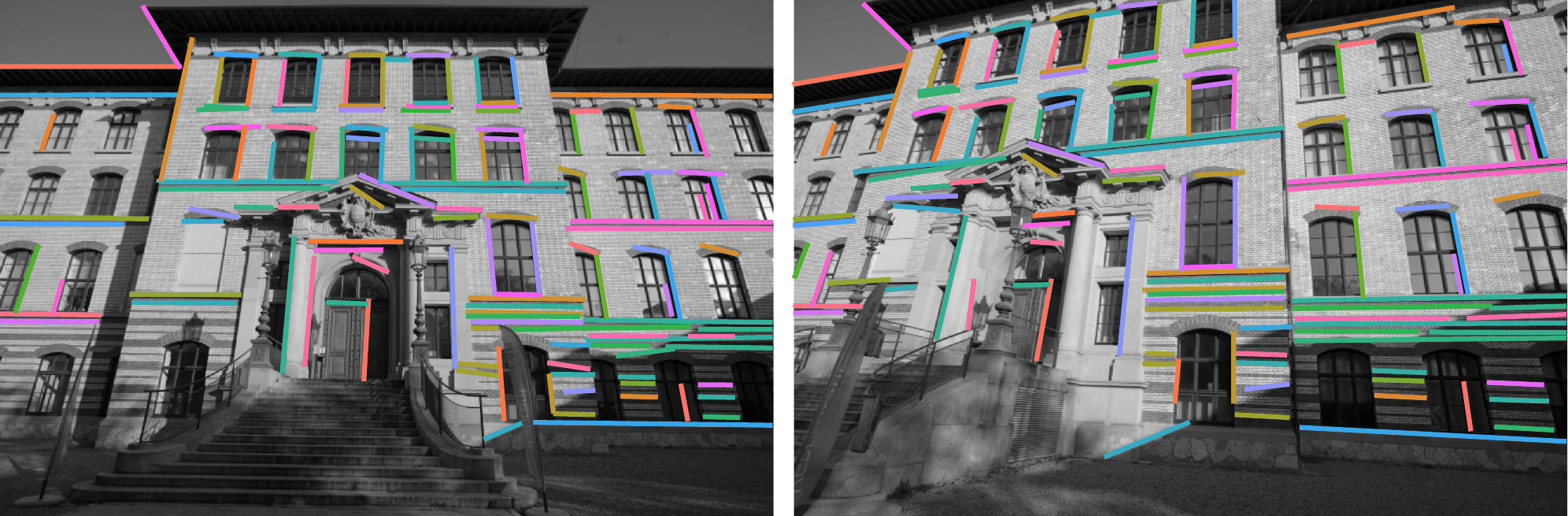}
    \end{tabular}
    \caption{\textbf{Joint matching of points and lines.} Matching feature points often fails in textureless areas (a), while current line matching methods struggle with large viewpoint changes (c). We propose GlueStick, a network jointly matching points and lines. While none of the methods were trained on the rotations of (a)(b), line matches can guide GlueStick while SuperGlue~\cite{sarlin_2020_superglue} fails, and vice-versa in (d) where points can complement the line matching.
    For clarity reasons, we show here only the correct matches.\vspace{-0.2cm}}
    \label{fig:teaser}
\end{figure}
\section{Introduction}
\label{sec:intro}

Line segments are high-level geometric structures useful in a wide range of computer vision tasks such as SLAM~\cite{gomez_2019_plslam, zuo_2017_robust, pumarola_2017_plslam}, pose estimation~\cite{xu_2017_pose}, construction monitoring~\cite{Kropp_2018_interior, asadi_2019_real}, and 3D reconstruction~\cite{hofer_2017,Zeng_2020_bundle, Zhou_2019_learning}. Lines are ubiquitous in structured scenes and offer stronger constraints than feature points. In particular, lines shine in low-textured scenes where point-based approaches struggle.

However, compared to keypoints, line segments are often poorly localized in the image and suffer from lower repeatability.
Line segments are also more challenging to describe since they can cover a large spatial extent in the image and suffer from occlusions and perspective effects due to viewpoint changes. 
Furthermore, lines often appear as part of repetitive structures in human-made environments, making classical descriptor-based matching fail. 
For this reason, typical matching heuristics such as mutual nearest neighbor and Lowe's ratio test~\cite{lowe_2004_distinctive} 
are often less effective for lines.

Recently, deep learning has ushered in a new paradigm for feature point matching using Graph Neural Networks (GNNs)~\cite{sarlin_2020_superglue,sun_2021_loftr}. This new approach bypasses the need for matching heuristics or even outlier removal techniques, thanks to the high precision of the predicted matches~\cite{sarlin_2020_superglue}. A key component to achieve this is to leverage the positional encoding of keypoints directly in the network and to let it combine visual features with geometric information~\cite{sarlin_2020_superglue, sun_2021_loftr, truong_2021_learning, jiang_2021_cotr}. Letting the GNN reason with all features simultaneously brings in additional context and can disambiguate repetitive structures (\cref{fig:teaser}).

Even though recent advances leveraged similar ideas to enrich line descriptors~\cite{syoon_2021_linetr}, directly transferring this GNN approach to line matching is not trivial. The large extent of lines and their lack of repeatability make it hard to find a good feature representation for them.
In this paper, we take inspiration from SuperGlue~\cite{sarlin_2020_superglue} and introduce GlueStick, to jointly match keypoints and line segments.
Our goal is to leverage their complementary nature in the matching process. By processing them together in a single GNN, the network can learn to resolve ambiguous line matches by considering nearby distinctive keypoints, and vice versa.
We propose to leverage the connectivity between points and lines via a unified wireframe structure, effectively replacing previous handcrafted heuristics for line matching~\cite{zhang_2013_lbd, schmid_1997_automatic, li_2016_ljl} by a data-driven approach.

Our network takes as input sparse keypoints, lines, and their descriptors extracted from an image pair, and outputs a set of locally discriminative descriptors enriched with the context from all features in both images, before establishing the final matches. Inside the network, keypoints and line endpoints are represented as nodes of a wireframe. The network is composed of self-attention layers between nodes, cross-attention layers exchanging information across the two images, and a new line message passing module enabling communication between neighboring nodes of the wireframe.
After the GNN, points and lines are split into two separate matching matrices and a dual-softmax is used to find the final assignment of the features.
Overall, our contributions are as follows:
\begin{enumerate} \itemsep0pt
	\item We replace heuristic geometric strategies for line matching with a data-driven approach, by jointly matching points and lines within a single network.
	\item We offer a novel architecture exploiting the local connectivity of the features within an image.
	\item We experimentally show large improvements of our method over previous state-of-the-art point and line matchers on a wide range of datasets and tasks.
\end{enumerate}

\section{Related Work}

\begin{figure*}[ht]
	\centering
	\includegraphics[width=0.9\linewidth]{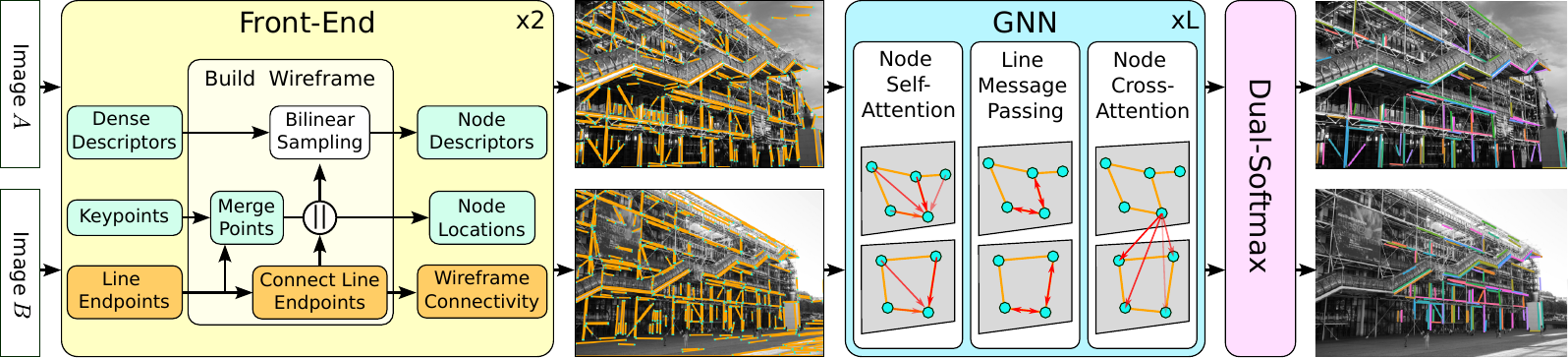}
	\caption{\textbf{Overview of GlueStick.} Keypoints, dense descriptors, and lines are extracted from two images, and unified into two wireframes (front-end). We then enrich the features of the nodes of both wireframes via self, line, and cross-attention inside a Graph Neural Network (GNN). Finally, points and lines are matched separately via two dual-softmax modules.}
	\label{fig:overview}
\end{figure*}

\textbf{Line segment detection} is a classical problem in computer vision that can be traced back to the Hough Transform~\cite{hough_1962_method} and its improvements~\cite{matas_2000_robust, fernandes_2008_real, zhao_2021_deep}. Local line segment detectors~\cite{gioi_2010_lsd, akinlar_2011_edlines, suarez_2018_fsg, suarez_2022_elsed,Pautrat_2023_DeepLSD} are efficient alternatives that fit segments to local regions with a prominent gradient. With more computational cost, deep line segment detectors offer better detection results, in particular for specialized tasks such as wireframe parsing~\cite{Huang_2018_wireframe, Zhou_2019_lcnn, Xue_2019_afm, Zhang_2019_ppgnet, Xue_2020_hawp, Xu_2021_letr}. In this work, we train our network with the LSD detector~\cite{gioi_2010_lsd}, due to its high accuracy and versatility.

\textbf{Line segment description} is classically performed by using the image gradients to describe the texture around each segment locally~\cite{wang2009wide, wang2009msld, zhang_2013_lbd, verhagen2014scale, li_2016_ljl}.  
More recently, deep learning models have emerged. 
Early works mimic keypoint patch descriptors by extracting a patch around each line and describing it via a neural network~\cite{lange_2019_dld,Kruger_2020_wld,l2d2}. An alternative approach is to sample points along the line and to describe them separately~\cite{vakhitov_2019_lld,pautrat_2021_sold2}. SOLD${}^2$~\cite{pautrat_2021_sold2} introduces a joint detection and description of line segments, as well as a mechanism to handle the partial occlusion of lines during matching.
In this paper, we use the (point-based) SuperPoint~\cite{detone_2018_superpoint} dense descriptors, interpolated at the two line endpoints. While these might not capture the full visual context of the line, having comparable descriptors for both points and lines is crucial in our network.

Since descriptor-based matching for line segments is generally more difficult than for points, several methods in the literature complemented the descriptor matching with geometric scene information~\cite{li_2016_line}:  global rotation between images~\cite{zhang_2013_lbd}; properties of pairs of matched lines like the angle between segments, intersection ratios or projection ratios~\cite{zhang_2013_lbd, wang2009wide}; line-point invariants~\cite{fan_2012_robust}; cross-ratio~\cite{ramalingam_2015_line} or consistency with a fundamental matrix estimated from points~\cite{schmid_1997_automatic, li_2016_ljl}. 
However, estimating the fundamental matrix to perform matching generates a chicken-and-egg problem, and these heuristics often fail in realistic scenarios. For this reason, recent point matchers are learning the geometric relationships between the points of two images, thus implicitly learning the underlying epipolar geometry~\cite{sarlin_2020_superglue}.

\textbf{Matching with transformers}. SuperGlue~\cite{sarlin_2020_superglue} uses a GNN to process keypoints and their descriptors from two input images, adding a positional encoding to better disambiguate repetitive patterns. Several variations of this method have been proposed later, with higher efficiency~\cite{Chen_2021_ICCV,Shi2022ClusterGNNCC} and with dense predictions~\cite{sun_2021_loftr,truong_2021_learning,jiang_2021_cotr,wang2022matchformer,Chen2022ASpanFormerDI,edstedt2023dkm}.

WGLSM~\cite{ma_2021_wglsm} combines a CNN and a GNN to match line segments, but without feature points. In the GNN, each line is represented with a single node, and the assignation is solved using a single Sinkhorn matrix. LineTR~\cite{syoon_2021_linetr} proposes to use attention inside points sampled for each line to deal with the line scale changes and occlusions. HDPL~\cite{guo_2021_hdpl} mixes points and lines in the same GNN, each line being represented with a single node in the GNN. They only use a single Sinkhorn matrix, allowing point-line assignments.

In contrast to these methods, we model each line segment endpoint as a separate node in the GNN. The endpoints are, in most cases, consistent with the underlying epipolar geometry, allowing the network to leverage both points and lines to disambiguate the matching. In our ablation study, we show that matching points and line endpoints together already greatly improves the matching performance.

\section{GlueStick}

In this section, we show how to combine points and lines within the same network. The motivation for this is that each feature can leverage cues from the neighbouring features to improve the matching performance. For example, a line using the surrounding points or vice-versa. Furthermore, the network can automatically discover combinations of points and lines that are useful for matching, instead of heuristically mining them as in previous works~\cite{li_2016_ljl}.
Our architecture, displayed in \cref{fig:overview}, consists in three blocks:
\begin{enumerate}
	\item \textbf{Front-End}: We extract points, lines, and their descriptors with common feature detectors, then combine them into a single wireframe (Sec.~\ref{sec:preprocessing}).
	\item \textbf{GNN}: The goal of this block, described in Sec.~\ref{sec:gnn}, is to combine the visual and spatial information of each feature, and to allow interaction between all features, regardless of their original receptive field. The output is a set of updated descriptors, enriched by the knowledge of relevant features within and across images, as well as within nodes connected by a line segment.
	\item \textbf{Dual-Softmax}: The final assignation is solved separately for points and lines, using two independent dual-softmax modules~\cite{Rocco18b,sun_2021_loftr}, as detailed in Sec.~\ref{sec:dual_softmax}.
\end{enumerate}

\subsection{From Points and Lines to Wireframes}
\label{sec:preprocessing}
The input to our GNN is a set of points, their associated local descriptors, and a connectivity matrix indicating which points are connected by a line. The first step is to establish this connectivity and build the wireframe graph.

We use SuperPoint (SP)~\cite{detone_2018_superpoint} to predict keypoints and a dense descriptor map, and we detect segments with the general-purpose LSD~\cite{gioi_2010_lsd} detector. Keypoints located close to line endpoints are redundant, so we remove SP keypoints that are within a small distance $d$ to existing line endpoints.

Furthermore, line segments generated by generic detectors such as LSD are usually disconnected. To give more structure to the input and to explicitly encourage the network to reason in terms of line connectivity, we merge close-by endpoints, again with a distance threshold $d$. This process lifts the unstructured line cloud into an interconnected wireframe. After this step, each keypoint and line endpoint is represented as a node in the wireframe, with different connectivities for each node: 0 for an isolated keypoint, 2 for a corner, etc. We then interpolate the dense SP feature map at the node locations to equip them with a visual descriptor.
Note that this endpoint merging is modifying the position of the endpoints but not the number of lines. For downstream tasks requiring high precision, we use the original position of the endpoints, to keep the sub-pixel accuracy of the original detector.

\begin{figure*}[t]
	\centering
	\includegraphics[width=0.9\linewidth]{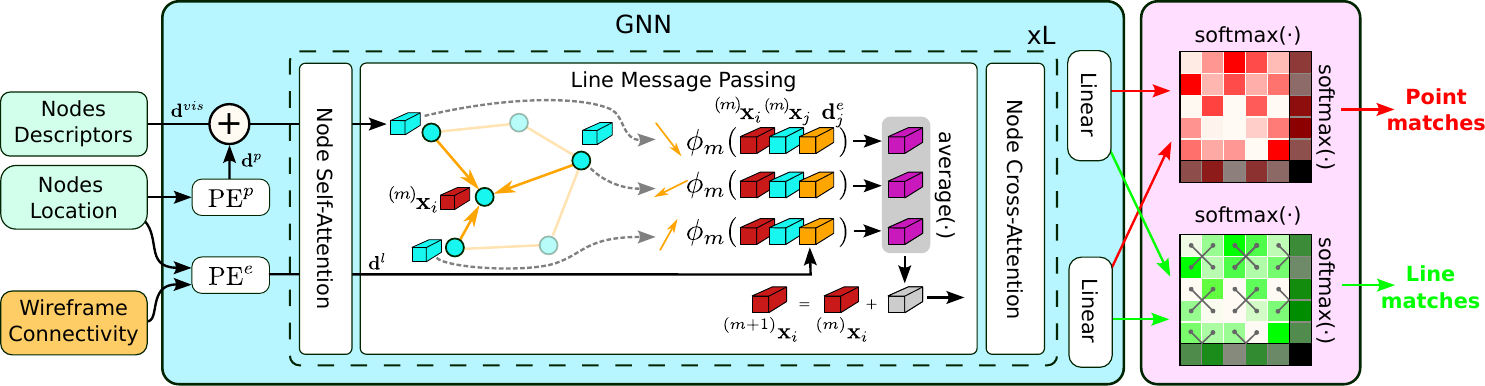}
	\caption{\textbf{Graph Neural Network (GNN) architecture.} Node features of the wireframe are enriched via several communication layers. Our Line Message Passing exchanges information between neighboring nodes that are connected together.}
	\label{fig:gnn}
\end{figure*}

\subsection{Attention-based Graph Neural Network (GNN)}
\label{sec:gnn}
A key part of our method is the GNN, which aggregates visual and spatial information to  predict a set of \textit{enriched} feature descriptors, that are used to establish the final matches via descriptor similarity.
Within the network, each node (either a keypoint, or a line endpoint) is associated with an initial descriptor that is based on the visual appearance as well as the position in the image. 

Let $A$ and $B$ be a pair of images. For each image, the inputs of the network are: a set of nodes $\mathbf{p}$, with coordinates ($x_p$, $y_p$), confidence score $s_p$ and visual descriptors $\mathbf{d}^{vis} \in \mathbb{R}^{D}$; and a set of line segments $\mathbf{l}$ defined as a pair of nodes ($x_p$, $y_p$) and ($x'_p$, $y'_p$), and with a line score $s_l$. This line score can be any value returned by the line detector indicating the quality of the line, or simply the length of the line to put more emphasis on longer lines.
The node score $s_p$ is either coming from the keypoint detector, or is equal to $s_l$ when it is a line endpoint.

\noindent\textbf{Positional and Directional Encoding.}
The first step is to encode the spatial information of each feature. To this end, we learn two positional encoders ($\text{PE}^p$ and $\text{PE}^e$) with Multi-layer Perceptron (MLP) that generate a \textit{spatial} descriptor $\mathbf{d}^p$ for each node and an \textit{edge}-descriptor $\mathbf{d}^e$ for each line segment originating from this node. A node with connectivity 3 will for instance get assigned one $\mathbf{d}^p$ and 3 $\mathbf{d}^e$ (one for each outgoing line segment). The edge-positional encoding takes as additional information the offset to the other endpoint of its line segment, allowing it to have access to the angle and length of the line segment:
\begin{equation}
	\begin{aligned}
		\mathbf{d}^p & = \text{PE}^p([x_p, y_p, s_p]^{\top}) \\
		\mathbf{d}^e & = \text{PE}^e([x_p, y_p, x'_p - x_p, y'_p - y_p, s_l]^{\top}) .
	\end{aligned}
\end{equation}
The spatial-descriptor $\mathbf{d}^p$ is used to initialize the node features, while the edge-descriptors $\mathbf{d}^e$ are used in the line message passing (see below).

\noindent\textbf{Network Architecture.}
Our GNN is a complete graph with three types of undirected edges (See \cref{fig:overview}). 
Self-attention edges $\mathcal{E}_{\text{self}}$, connect nodes of one image with all the nodes of the same image. 
Line edges $\mathcal{E}_{\text{line}}$, connect nodes that are endpoints of the same line. 
Cross attention edges $\mathcal{E}_{\text{cross}}$, connect nodes of one image to the other image nodes.

A node $i$ is initially assigned a feature descriptor fusing its spatial and visual information: ${}^{(0)}\mathbf{x}_i = \mathbf{d}^p_i + \mathbf{d}^{vis}_i$.
This node descriptor is then iteratively enriched and refined with the context of all the other descriptors in $L$ iterations of Self, Line, and Cross layers. Finally, the features of each node are linearly projected to obtain the output features. The next paragraphs detail each type of layer.

~

\noindent\textbf{Self and Cross Layers.}
$\mathcal{E}_{\text{self}}$ and $\mathcal{E}_{\text{cross}}$ edges are similarly defined as in \cite{sarlin_2020_superglue}. The $m$-th feature update is defined by a residual message passing:

\begin{equation}
	{}^{(m+1)}\mathbf{x}_{i}={}^{(m)} \mathbf{x}_{i}+
	\psi_m \left(\left[^{(m)} \mathbf{x}_{i} || a_m({}^{(m)} \mathbf{x}_{i}; \mathcal{E}) 
	\right]\right) ,
\end{equation}
where $||$ denotes concatenation, the function $\psi_m$ is modeled with an MLP, and $a_m({}^{(m)} \mathbf{x}_{i}; \mathcal{E})$ is the Multi-Head Attention mechanism from \cite{vaswani_2017_transformers} applied to the set of edges $\mathcal{E}$:
\begin{equation}
	a_m(\mathbf{x}_{i}; \mathcal{E}) =\sum_{j:(i, j) \in \mathcal{E}} \operatorname{softmax}_{j}\left(\frac{\mathbf{q}_{i}^{\top} \mathbf{k}_{j}}{\sqrt{D}}\right) \mathbf{v}_{j} ,
\end{equation}
where the keys $\mathbf{k}_{j}$, queries $\mathbf{q}_{i}$, and values $\mathbf{v}_{j}$ are computed as linear projections of the node features $\mathbf{x}_{i}$ and $\mathbf{x}_{j}$. In self-attention layers, $\mathbf{k}_{j}$ and $\mathbf{v}_{j}$ will come from the same image, whereas in cross-attention they will come from the other image. Self-attention allows the network to leverage the context of the full image, and to resolve repetitive structures. Cross-attention moves corresponding features closer in descriptor space and can search for similar node structures in the other image to fully leverage spatial information.

~

\noindent\textbf{Line Message Passing.}
We describe here our novel Line Message Passing (LMP) transferring information across the line edges $\mathcal{E}_{\text{line}}$.
By connecting line segments in a wireframe structure, we allow the $i$-th node to leverage the local edge connectivity to the set $\mathcal{N}_i$ of neighboring nodes, and to look for the same type of connectivities in the other image.
This mechanism is enabled by the $m$-th LMP update which aggregates the information contained in the two endpoint features ${}^{(m)} \mathbf{x}_{i}$ and ${}^{(m)} \mathbf{x}_{j}$ and the corresponding endpoint positional encoding $\mathbf{d}^e_j$:
\begin{equation}
	{}^{(m+1)} \mathbf{x}_{i} = {}^{(m)} \mathbf{x}_{i} + \sum_{j \in \mathcal{N}_{i}} \frac{\phi_m ([{}^{(m)} \mathbf{x}_{i} || {}^{(m)} \mathbf{x}_{j} || \mathbf{d}^e_j ])}{|\mathcal{N}_{i}|}  ,
\end{equation}
where $\phi_m$ denotes again an MLP and $|\mathcal{N}_{i}|$ is the number of neighbors of node $i$. We use here a simple average across all neighbors.
An attention mechanism could also have been applied, but we empirically found that it only increased the complexity of the model, for no gain in performance.

\subsection{Dual-Softmax for Points and Lines}
\label{sec:dual_softmax}
\begin{figure}
	\centering
	\includegraphics[width=0.9\linewidth]{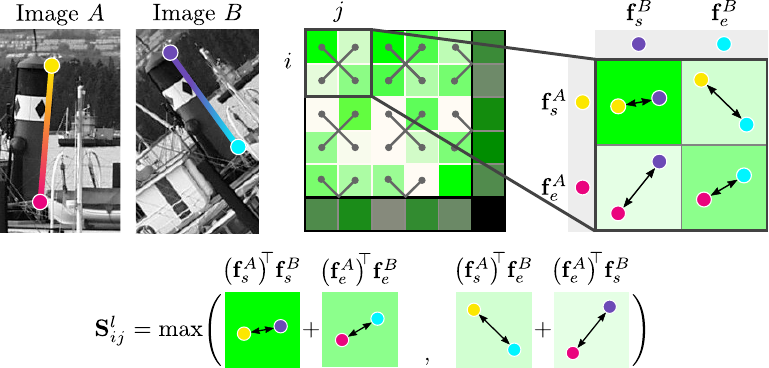}
	\caption{\textbf{Line matching with order-agnostic endpoints.} We consider the maximum score assignment between the two possible configurations of endpoint matching.}
	\label{fig:structural-similarity-diagram}
\end{figure}
Recent works~\cite{Rocco18b,sun_2021_loftr} show that dual-softmax approach obtains similar or better results than the usual Sinkhorn algorithm~\cite{sinkhorn_1967_concerning,sarlin_2020_superglue}, being also more efficient. We observed similar behaviour in our experiments and opted for the dual-softmax assignment.
GlueStick provides both point and line matches in a single forward pass. We match nodes and lines separately through two independent dual-softmax assignments. On the one hand, all nodes (keypoints and line endpoints) are matched against each other using the final features output by the GNN: $\mathbf{f}^A_i \in \mathbb{R}^D$ for node $i$ in image $A$ and $\mathbf{f}^B_j \in \mathbb{R}^D$ for node $j$ in image $B$. Each element of the assignment matrix $\mathbf{S}^{p}$ is formed by:
\begin{equation}
	\mathbf{S}_{ij}^p = (\mathbf{f}^A_i)^{\top} \mathbf{f}^B_j .
\end{equation}
We add a dustbin row and column at the end of $\mathbf{S}^p$, filled with a learnable parameter representing the threshold below which a node is considered unmatched, as \cite{sarlin_2020_superglue} also does. We then apply softmax on all rows and all columns, and compute their geometric mean:
\begin{equation}
	\mathbf{S}_{\mathrm{final}}^p = \sqrt{\mathrm{softmax}_{\mathrm{row}}(\mathbf{S}^p) \odot \mathrm{softmax}_{\mathrm{col}}(\mathbf{S}^p)}.
	\label{eq:dual_softmax}
\end{equation}
Where $\odot$ means the element-wise product. Given this final assignation matrix, we keep the mutual nearest neighbors that have a matching score above a given threshold $\eta$.

On the other hand, lines are matched in a similar way, except that each line is represented by its two endpoints features $\textbf{f}_s \in \mathbb{R}^D$ and $\textbf{f}_e \in \mathbb{R}^D$. To make the matching agnostic of the endpoint ordering, we take the maximum of the two configurations in the line assignation matrix (see \cref{fig:structural-similarity-diagram}):
\begin{align}
	\begin{split}
		\mathbf{S}_{ij}^l = \max\bigl(&\left(\mathbf{f}^A_s\right)^{\top} \mathbf{f}^B_s + \left(\mathbf{f}^A_e\right)^{\top} \mathbf{f}^B_e,\\
		&\left(\mathbf{f}^A_s\right)^{\top} \mathbf{f}^B_e + \left(\mathbf{f}^A_e\right)^{\top} \mathbf{f}^B_s\bigl) .
	\end{split}
\end{align}
Finally, we get $\mathbf{S}_{\mathrm{final}}^l$ by applying the dual-softmax of Eq.~\ref{eq:dual_softmax} and match lines with mutual nearest neighbors.

\subsection{Ground Truth Generation}
\label{sec:gt_generation}
A challenging task in line matching is to generate high-quality labels handling line fragmentation, assignation, and partial visibility. To obtain the Ground Truth (GT) point matches $\mathcal{M}^p$, we use the same methodology as in \cite{sarlin_2020_superglue}.
In a nutshell, we leverage camera poses and depth to re-project keypoints from one image to another, and we add a new match whenever a re-projection falls within a small neighborhood of an existing keypoint.

For lines, we also leverage depth, but with a more complex setup. Let images $A$ and $B$ contain $M$ and $N$ line segments indexed by $\mathcal{A}:=\{1, \ldots, M\}$ and $\mathcal{B}:=\{1, \ldots, N\}$. We will denote the generated GT line matches
$\mathcal{M}^l=\{(i, j)\} \subset \mathcal{A} \times \mathcal{B}$.
For each segment $\mathbf{l}_i^{A}$ detected on image $A$, we sample K points  $\left[\mathbf{x}^{A}_{i,1}, \dotsc, \mathbf{x}^{A}_{i,K} \right]$ along it. A point is considered invalid if it has either no depth or its projection $\mathbf{x}^{B}_i$ in the other image has no depth.
A point is also considered non-valid if it is occluded.
We detect these cases by comparing the depth $d(\mathbf{X}_i)$ of the unprojection $\mathbf{X}_i$ in 3D of point $\mathbf{x}^{A}_i$ with its expected depth $d^{B}$ in image $B$:
\begin{equation}
	\mathrm{Occluded} = \frac{|d(\mathbf{X}_i) - d^{B}|}{d^{B}} > T_{\text{occlusion}} ,
\end{equation}
where $T_{\text{occlusion}}$ defines the tolerance threshold of depth variations.
Segments with more than 50\% of invalid points are labeled as \texttt{IGNORE} and will not affect the loss function.

Next, we generate a closeness matrix $\mathbf{C}^{B} \in \mathbb{N}^{M \times N}$ keeping track of how many sampled points of line $i$ in $A$ are reprojected close to a line $j$ in $B$:
\begin{equation}
	\mathbf{C}^B_{i, j} = \sum_{k = 1}^{K} \mathbbm{1}\left( \operatorname{valid}(\mathbf{x}^{B}_{i,k}) \land d_{\perp}\left(\mathbf{x}^{B}_{i,k}, \mathbf{l}_j^{B} \right) < T_{\text{dist}} \right) ,
\end{equation}
where $\mathbbm{1}(\cdot)$ is the indicator function and $d_{\perp}(\cdot, \cdot)$ the perpendicular point-line distance. $T_{\text{dist}}$ is a distance threshold in pixels that controls how demanding the GT is. 
$\mathbf{C}^A$ is defined analogously, and thus, we can define a cost matrix $\mathbf{C}$ with a minimum overlap threshold $T_{\text{overl}}$:
\begin{equation}
	\mathbf{C}_{i, j} =  \begin{cases}
		\infty,              &\hspace{-8pt} \text{if } \mathbf{C}^A_{i, j} < T_{\text{overl}} \lor \mathbf{C}^B_{j, i} < T_{\text{overl}} \\
		-\mathbf{C}^A_{i, j} \mathbf{C}^B_{j, i}, &\hspace{-8pt} \text{otherwise}.
	\end{cases}
\end{equation}
Last, we solve the assignation problem defined by $\mathbf{C}$ with the Hungarian algorithm~\cite{kuhn_1955_hungarian}. The resulting assignations $(i, j) \in \mathcal{M}^l$ are the \texttt{MATCHED} features, whereas all the valid entries $\mathcal{I} \subseteq \mathcal{A}$ and $\mathcal{J} \subseteq \mathcal{B}$ that were not assigned are labeled as \texttt{UNMATCHED}.

\subsection{Loss Function}

A classical approach for descriptor learning is to apply the triplet-ranking-loss~\cite{balntas_2016_tfeat, suarez_2021_revisiting} with hard negative mining~\cite{mishchuk_2017_hardnet}.
However, repetitive structures are often present along lines, which may produce detrimental hard negatives.
We resort instead to minimizing the negative log-likelihood of point and line assignments $\mathbf{S}_{\mathrm{final}}^p$ and $\mathbf{S}_{\mathrm{final}}^l$:

\begin{equation}
	\mathcal{L} = \frac{\operatorname{NLL}(\mathbf{S}_{\mathrm{final}}^p, \mathcal{M}^p) + \operatorname{NLL}(\mathbf{S}_{\mathrm{final}}^l, \mathcal{M}^l)}{2} ,
	\label{eq:global_loss}
\end{equation}
where for an assignment matrix $\mathbf{A}$ and GT matches $\mathcal{M}$:
\begin{align}
	\operatorname{NLL}(\mathbf{A}, \mathcal{M}) =& -\sum_{(i, j) \in \mathcal{M}} \log \mathbf{A}_{i, j} \\
	&-\sum_{i \in \mathcal{I}} \log \mathbf{A}_{i, N+1}-\sum_{j \in \mathcal{J}} \log \mathbf{A}_{M+1, j} . \nonumber
\end{align}

\section{Experiments}
We pre-train our model on pairs of images synthetically warped by a homography, using the one million distractor images of \cite{radenovic2018}, increasing the difficulty of the homographies gradually and speeding up convergence. We then fine-tune the model on MegaDepth~\cite{Li_2018_MegaDepth} that contains 195 scenes of outdoor landmarks. We select image pairs with a minimum overlap of 10\% of 3D points and resize each to $640\times 640$ px. The wireframe threshold $d$ to merge nodes is set to 3 pixels, and to generate the GT: $T_{\text{occlusion}} = 0.1$, $T_{\text{dist}} = 5$, and $T_{\text{overl}} = 0.2$. Our GNN contains 9 blocks of [self-attention, line message passing, cross-attention], and the matching threshold is set to $\eta=0.2$. Features inside the network have size $D = 256$. We optimize our network using Adam with learning rate $10^{-4}$ for the homography pre-training and $10^{-5}$ for MegaDepth. To limit computational cost during training, we set a maximum number of 1000 keypoints and 250 line segments per image. Training takes 10 days on 2 NVIDIA RTX2080 GPUs.

\subsection{Baselines}

In the following, we compare GlueStick with several state-of-the-art line matchers: the handcrafted Line Band Descriptor (LBD)\footnote{We use the authors' code instead of the binary version from OpenCV.}~\cite{zhang_2013_lbd}, the self-supervised SOLD${}^2$~\cite{pautrat_2021_sold2}, the transformer-based LineTR~\cite{syoon_2021_linetr}, and the learned L2D2~\cite{l2d2} descriptors. SOLD${}^2$ uses its own detector since it is integrated with the descriptor. For all the other methods we use LSD~\cite{gioi_2010_lsd}. We also compare to PL-Loc~\cite{syoon_2021_linetr}, the point-line matcher combining SuperPoint~\cite{detone_2018_superpoint} and LineTR~\cite{syoon_2021_linetr}. Whenever possible, we also compare to two additional point-based matchers: ClusterGNN~\cite{Shi2022ClusterGNNCC}\footnote{We reuse the numbers of the paper as the code is not publicly available.} and LoFTR~\cite{sun_2021_loftr}.

\subsection{Ablation Study}

Line segments are especially challenging to match in 3D due to occlusions, background changes, or partial visibility. We advocate for a proper evaluation of line matching covering these scenarios. Our ablation study is thus led on the ETH3D~\cite{Schops_2017_eth3d} dataset, an indoor-outdoor dataset of multiple scenes with GT LiDAR depth, and poses. We use the 13 scenes of the training set 
of the high-resolution multi-view images (downsampled by a factor of 8)
, and sample all pairs of images with at least 500 GT keypoints in common, similarly as in~\cite{pautrat_2021_sold2}. We apply the same methodology as in Sec.~\ref{sec:gt_generation} to compute the GT line matches. Given this GT, we can compute the precision-recall curve of the line matching by ordering lines by decreasing matching score.

\begin{figure}
	\centering
	\begin{subfigure}{0.54\linewidth}
		\scriptsize
		\setlength{\tabcolsep}{3pt}
		\renewcommand{\arraystretch}{1.3}
		\begin{tabular}{lccccc}
			\toprule
			& P & L & W & LMP & AP ($\uparrow$) \\
			\midrule
			SG + Endpts & \checkmark & \checkmark &  &  & 54.5 \\
			SG + W & \checkmark & \checkmark & \checkmark & & 67.6 \\
			SG + LMP & \checkmark & \checkmark &  & \checkmark & 69.9 \\
			GlueStick-L &  & \checkmark & \checkmark & \checkmark & 64.0 \\
			GlueStick & \checkmark & \checkmark  & \checkmark & \checkmark & \textbf{72.6} \\
			\bottomrule
		\end{tabular}
		\vspace{3pt}
		\caption{Ablation study}
		\label{fig:eth3d:ablation}
	\end{subfigure}
	\hfill
	\begin{subfigure}{0.44\linewidth}
		\includegraphics[width=\columnwidth]{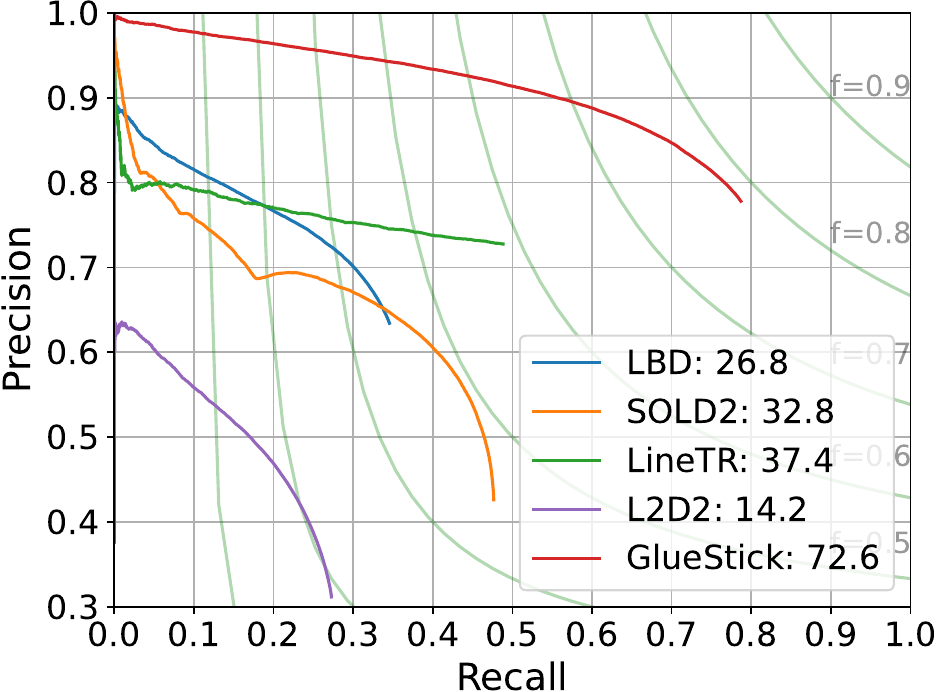}
		\caption{Comparison to SOTA}
		\label{fig:eth3d:sota}
	\end{subfigure}
	\caption{\textbf{Ablation study and comparison to the state of the art (SOTA) on the ETH3D dataset~\cite{Schops_2017_eth3d}.} We compute the line matching precision-recall curves and average precision (AP), displayed in the legend. (a) We compare several variations of our method using points (P), lines (L), wireframe connectivity (W), and Line Message Passing (LMP). (b) GlueStick surpasses all SOTA line matchers.}
	\label{fig:eth3d}
\end{figure}

We compare several variations of our method in \cref{fig:eth3d:ablation}. \textit{SG + Endpts} refers to the pre-trained outdoor model of SuperGlue~\cite{sarlin_2020_superglue} to match the line endpoints, and use our proposed line association of Sec.~\ref{sec:dual_softmax} agnostic to the ordering of endpoints. \textit{SG + W} is similar, but with our proposed wireframe preprocessing connecting line segments together. \textit{SG + LMP} represents a SuperGlue backbone with the addition of our Line Message Passing (LMP), but no wireframe preprocessing. Finally, \textit{GlueStick-L} is our proposed model without keypoints and matching lines only. The average precision (AP) shows that both the wireframe pre-processing and LMP bring a large boost of performance on the SuperGlue baseline. Their combination - our proposed model GlueStick - obtains the highest performance. \textit{GlueStick-L} loses performance, but remains competitive, showing that the line matching is not relying only on points.

\subsection{Line Matching Evaluation on ETH3D}

We compare our method with previous state-of-the-art line matchers on the ETH3D dataset~\cite{Schops_2017_eth3d}, and show the blatant superiority of GlueStick in \cref{fig:eth3d:sota}. It recovers almost 80\% of the GT line matches, whereas the best previous methods do not manage to reach 50\% of recall. At equivalent recalls, it outperforms the previous best method, LineTR, by more than 15\% in precision, and is almost doubling the AP. This major improvement is due to the possibility of leveraging points in the matching, and to the rich signal provided by the wireframe structure. Note that GlueStick without points (in \cref{fig:eth3d:ablation}), is still significantly better than other pure line matchers. It obtains good results thanks to the inclusion of a graph matching strategy combining appearance similarities and geometric consistencies. Despite having powerful descriptors, L2D2 and SOLD$^2$ obtain worse results, because they neither use scene points nor geometric consistency between matches.

In terms of run time, GlueStick is also competitive. It runs in 258 ms on average on the images of ETH3D (around 775$\times$515 pixels), which is similar to the execution time of SuperGlue of 235 ms. Other line matchers are even slower, with 419 ms for SOLD${}^2$ and 304 ms for LineTR.

\subsection{Homography Estimation}

We evaluate our method on the task of homography estimation. While HPatches~\cite{hpatches_2017_cvpr} is the most popular dataset, it is now very saturated~\cite{sarlin_2020_superglue,sun_2021_loftr}, and contains few structural lines that would be necessary to properly estimate a homography. Thus, lines do not help much to improve the current performance obtained by point methods. Nevertheless, GlueStick ranks first among all considered methods on HPatches. We show these results in the supplementary material.
To circumvent this, we implement two meaningful experiments evaluating the homography estimation task in real-world scenarios: relative pose from planar surfaces (Sec.~\ref{sec:scannet_rel_pose}), and relative pose with pure rotations (Sec.~\ref{sec:rotation_rel_pose}).

\subsubsection{Dominant Plane on ScanNet}
\label{sec:scannet_rel_pose}

ScanNet~\cite{dai2017scannet} is a large-scale RGB-D indoor dataset with GT camera poses, which pictures some hard cases for feature points with low texture, and where lines are expected to provide better constraints. We use the same test set of 1500 images as in \cite{sarlin_2020_superglue}, where the overlap between image pairs is computed from GT poses and depth. For each image pair, we 
match them with different state-of-the-art point, line, and point-line matchers. We then use a hybrid RANSAC~\cite{Sattler2019Github,Camposeco2018CPVR} to estimate a homography from these feature correspondences. 
This is a common way to initialize SLAM systems~\cite{mur2015orb-slam}. 
Since the GT homography is not known, we rely on the GT relative pose to evaluate the quality of the retrieved homography, as was done in previous works~\cite{barath_23_homographies}. The relative pose corresponding to the predicted homography can be extracted using \cite{malis:inria-00174036}. We report the pose error, computed as the maximum of the angular error in translation and rotation~\cite{Yi2018LearningTF,Brachmann2019NeuralGuidedRL,sarlin_2020_superglue}, as well as the corresponding pose AUC at error thresholds 10 / 20 / 30 degrees error. 
Note that this evaluation is valid regardless of the plane selected by each method to estimate the homography: all planes lead to the same relative pose.

The results are shown in Tab.~\ref{tab:scannet_rel_pose}. It can be seen first that GlueStick matching points only obtains better results than SuperGlue. This shows that our re-trained network is able to match and even outperform SuperGlue network for keypoint matching. Secondly, when matching lines only, GlueStick significantly exceeds the previous state of the art for line matching.
This demonstrates that leveraging context from neighboring lines and being aware of their interconnection is highly beneficial. Finally, we obtain the best results overall when combining points and lines. The network can leverage both kinds of features and may rely more on the accurate points on well-textured images, while using lines in scenarios with scarce points.

\begin{table}
	\centering
	\scriptsize
	\setlength{\tabcolsep}{5pt}
	\begin{tabular}{clcc}
		\toprule
		& & Pose error ($\downarrow$) & Pose AUC ($\uparrow$) \\
		\midrule
		\multirow{3}{*}{Points} & SuperGlue (SG)~\cite{sarlin_2020_superglue} & 18.1 & 15.6 / 29.8 / 39.4 \\
		& LoFTR~\cite{sun_2021_loftr} & 16.8 & 15.8 / 30.9 / 41.4 \\
		& GlueStick & \textbf{15.7} & \textbf{17.4 / 32.8 / 42.9} \\
		\midrule
		\multirow{6}{*}{Lines} & LBD~\cite{zhang_2013_lbd} & 49.2 & \phantom{1}3.7 / \phantom{1}8.2 / 13.4 \\
		& SOLD${}^2$~\cite{pautrat_2021_sold2} & 55.6 & \phantom{1}4.9 / 10.8 / 16.1 \\
		& LineTR~\cite{syoon_2021_linetr} & 51.6 & \phantom{1}4.5 / 11.0 / 16.8 \\
		& L2D2~\cite{l2d2} & 60.0 & \phantom{1}2.8 / \phantom{1}6.5 / 10.5 \\
		& SG + Endpts (no KP) & 36.0 & \phantom{1}7.1 / 15.0 / 22.2 \\
		& GlueStick & \textbf{27.6} & \textbf{\phantom{1}9.4 / 20.0 / 28.6} \\
		\midrule
		\multirow{3}{*}{\makecell{Points\\+ Lines}} & PL-Loc~\cite{syoon_2021_linetr} & 26.2 & 12.2 / 24.1 / 32.2 \\
		& SG + Endpts & 17.1 & 17.5 / 31.8 / 41.2 \\
		& GlueStick & \textbf{14.1} & \textbf{19.3 / 35.4 / 46.0} \\
		\bottomrule
	\end{tabular}
	\caption{\textbf{Homography estimation on ScanNet~\cite{dai2017scannet}.} We first estimate a homography based on point-only, line-only or points+lines matches, then decompose it into the corresponding relative pose. We report the median pose error in degrees, as well as the AUC at 10$^\circ$ / 20$^\circ$ / 30$^\circ$ error.}
	\label{tab:scannet_rel_pose}
\end{table}

\subsubsection{Pure Rotations on SUN360}
\label{sec:rotation_rel_pose}
We also evaluate our method in estimating pure camera rotations, which are the cornerstone of some applications such as image stitching, or visual-guided sensor fusion.
We use the SUN360~\cite{xiao2012recognizing} dataset containing 360º images. From each original 360º image, we crop 10 pairs of perspective images (640$\times$480 pixels) with a field-of-view of 80º. Pairs are randomly sampled with an angular difference in range $\pm\left[50^{\circ}, 70^{\circ}\right]$ for yaw and $\pm\left[0^{\circ}, 30^{\circ}\right]$ for pitch. 
We first extract the local feature matches, then we estimate a rotation using Hybrid RANSAC~\cite{Sattler2019Github,Camposeco2018CPVR}.
We evaluate the angular error between the predicted and GT relative rotations.

\begin{figure}
	\centering
	\small
	\includegraphics[width=0.9\linewidth]{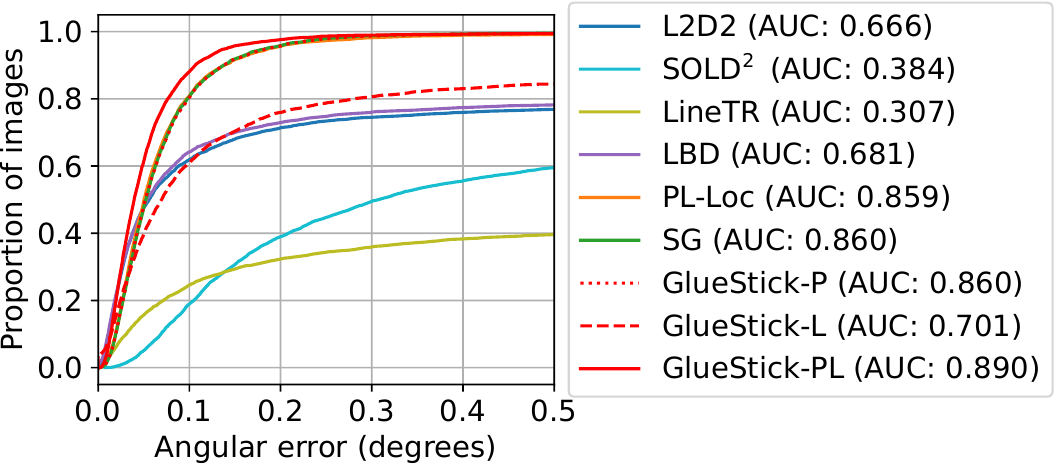}
	\vspace{-0.2cm}
	\caption{\textbf{Camera rotation estimation in SUN360~\cite{xiao2012recognizing}.} We show the cumulative angular error for all pairs of images. We report the AUC up to an error threshold of 0.5$^\circ$.
	}
	\label{fig:pure_rotation}
\end{figure}

In \cref{fig:pure_rotation}, GlueStick-PL (with points and lines) obtains the best results because lines help to match pairs where there is not enough texture. Specifically, long lines can be detected very precisely, thus contributing to an accurate estimation.
Point-based methods (SG and GlueStick-P) obtain already 3 points less of AUC.
PL-Loc~\cite{syoon_2021_linetr} is ranked fourth because it effectively uses points and lines, though independently and without spatial reasoning for point matches.

\subsection{Visual Localization}

We introduce here the downstream task of localizing a query image, given the known poses of database images. We follow the pipeline of LIMAP~\cite{Liu_2023_LIMAP}, which integrates line features into \emph{hloc}~\cite{sarlin2019coarse,hloc}.
We use NetVLAD~\cite{arandjelovic2016netvlad} for image retrieval, detect SuperPoint~\cite{detone_2018_superpoint} feature points and LSD~\cite{gioi_2010_lsd} lines, and match these features with either SuperGlue~\cite{sarlin_2020_superglue} + a line matcher, or with GlueStick.
We use the GT depth to back-project lines in 3D: points are sampled along each line, un-projected to 3D, and a 3D line is fit to these un-projected points.
We use the solvers of \cite{kukelova2016efficient,Zhou2018ASA,PoseLib} to generate poses from a minimal set of 3 features (3 points, 2 points and 1 line, 1 point and 2 lines, or 3 lines), then combine them in a hybrid RANSAC~\cite{Sattler2019Github,Camposeco2018CPVR} to recover the query camera poses.

\textbf{Datasets.}
We compare our method to other baselines on two datasets. The 7Scenes dataset~\cite{7scenes} is a famous RGB-D dataset for visual localization, displaying 7 indoor scenes with GT poses and depth. It is however limited in scale, and most scenes are already saturated for point-based localization. One scene remains extremely challenging for feature points: the Stairs scene, as illustrated in \cref{fig:visual_examples}. Due to the lack of texture and repeated steps of the stairs, current point-based methods are still struggling on this scene~\cite{dsacstar}.  We report median translation and rotation error, as well as the percentage of successfully recovered poses under a 5 cm / 5$^\circ$ threshold.
InLoc~\cite{wijmans17rgbd,taira2018inloc} is a large-scale indoor dataset with GT poses and depth, with two test scenes: DUC1 and DUC2. It is challenging for point-based methods due to images with low texture and large viewpoint changes. We report the pose AUC at 0.25m / 0.5m / 1m and 10$^\circ$.

\textbf{Results.}
The results can be found in Tab.~\ref{tab:visual_localization}. GlueStick with points only is able to surpass SuperGlue, confirming the strong matching performance of isolated keypoints already. In particular, GlueStick obtains an improvement of 44\% in pose accuracy over SuperGlue on Stairs. Adding line features significantly improves the performance for Stairs and brings a small improvement on InLoc as well. This demonstrates the importance of lines in texture-less areas and with repeated structures. Combining points and lines in a single network allows GlueStick to reason about neighboring features and can thus beat the other methods that are independently matching points and lines.

\begin{table}
	\centering
	\scriptsize
	\setlength{\tabcolsep}{3.8pt}
	\begin{tabular}{llcccc}
		\toprule
		& & \multicolumn{2}{c}{7Scenes~\cite{7scenes}} & \multicolumn{2}{c}{InLoc~\cite{taira2018inloc}} \\
		\cmidrule(r{5pt}l{5pt}){3-4} \cmidrule(r{5pt}l{5pt}){5-6}
		& & T / R err. & Acc. & DUC 1 & DUC 2 \\
		\midrule
		\multirow{4}{*}{P} & SuperGlue~\cite{sarlin_2020_superglue} & 4.7 / 1.25 & 53.4 & 48.5 / 68.2 / 80.3 & 53.4 / 75.6 / 82.4 \\
		& ClusterGNN~\cite{Shi2022ClusterGNNCC} & - & - & 47.5 / 69.7 / 79.8 & 53.4 / 77.1 / 84.7 \\
		& LoFTR~\cite{sun_2021_loftr} &  \textbf{4.4 / 0.95} & 53.9 & 47.5 / \textbf{72.2} / \textbf{84.8} & 54.2 / 74.8 / 85.5 \\
		& GlueStick & \textbf{4.4} / 1.21 & \textbf{55.4} & \textbf{49.0} / 70.2 / 84.3 & \textbf{55.0 / 83.2 / 87.0} \\
		\midrule
		\multirow{5}{*}{P+L}
		& SOLD${}^2$~\cite{pautrat_2021_sold2} & 3.2 / 0.83 & 75.8 & 44.9 / 69.7 / 79.8 & 54.2 / 75.6 / 80.2 \\
		& LineTR~\cite{syoon_2021_linetr} & 3.7 / 1.02 & 66.6 & 46.0 / 67.2 / 76.3 & 53.4 / 77.1 / 80.9 \\
		& L2D2~\cite{l2d2} & 4.1 / 1.15 & 55.8 & 46.5 / 68.7 / 80.3 & 51.9 / 75.6 / 79.4 \\
		& SG + Endpts & 3.1 / 0.81 & 75.6 & 45.5 / 71.2 / 81.8 & 45.5 / 71.2 / 81.8 \\
		& GlueStick & \textbf{2.9 / 0.79} & \textbf{79.7} & \textbf{47.5 / 73.7 / 85.9} &\textbf{57.3 / 83.2 / 87.0} \\
		\bottomrule
	\end{tabular}
	\vspace{-0.2cm}
	\caption{\textbf{Visual localization on 7Scenes~\cite{7scenes} and InLoc~\cite{taira2018inloc}.} We report the median translation (cm), rotation error (deg), and pose accuracy at a 5 cm / 5$^\circ$ threshold for the scene Stairs of 7Scenes, and the pose AUC at 0.25m / 0.5m / 1m and 10$^\circ$ error for InLoc. GlueStick ranks first both for points-only (P) and point-line (P+L) results.}
	\label{tab:visual_localization}
\end{table}

\begin{figure}
	\centering
	\scriptsize
	\newcommand{\sz}{0.45}
	\setlength{\tabcolsep}{1pt}
	\begin{tabular}{cc@{\hskip5pt}cc}
		\rotatebox{90}{SuperGlue~\cite{sarlin_2020_superglue}} & \includegraphics[width=\sz\columnwidth]{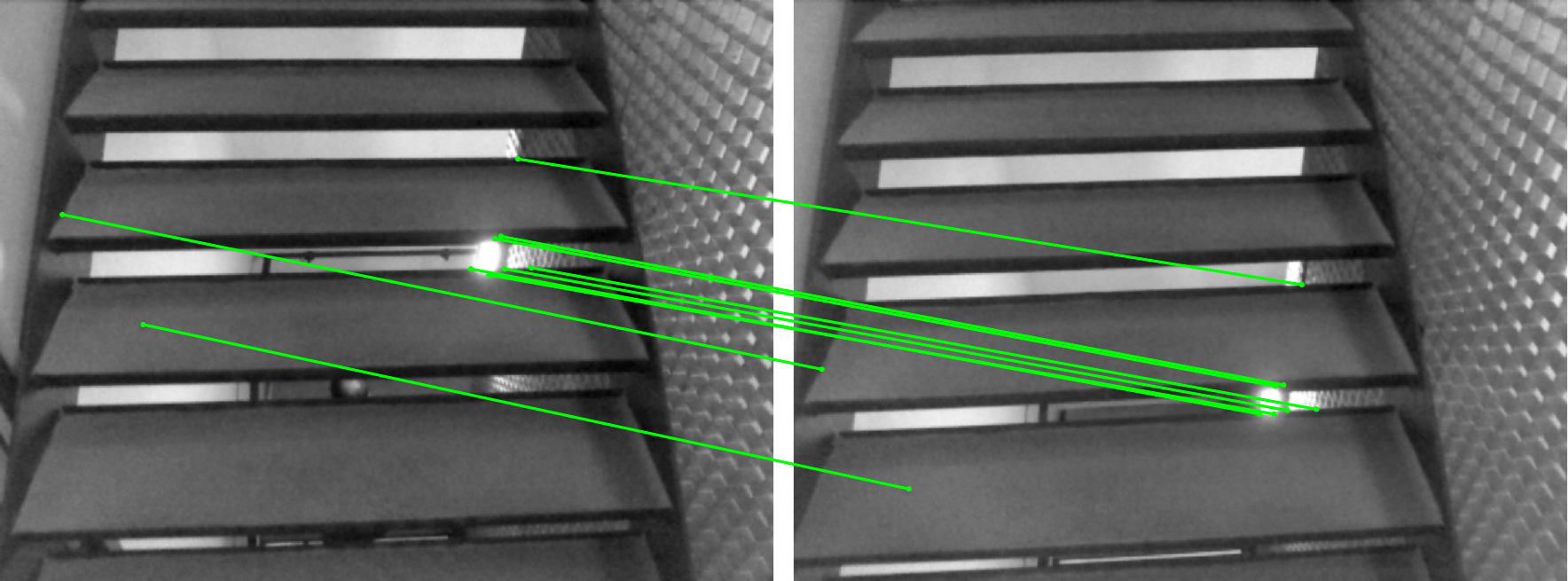} & \rotatebox{90}{\phantom{x}LineTR~\cite{syoon_2021_linetr}} & \includegraphics[width=\sz\columnwidth]{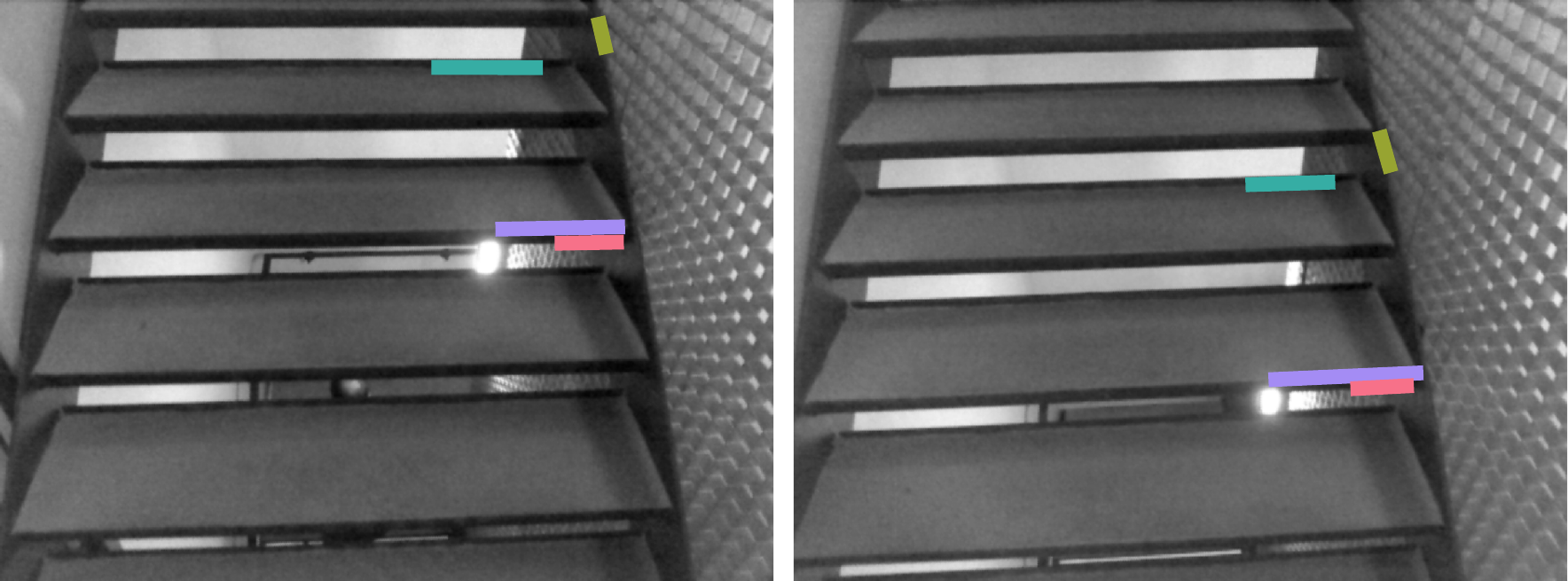} \\
		\rotatebox{90}{\phantom{x}GlueStick} & \includegraphics[width=\sz\columnwidth]{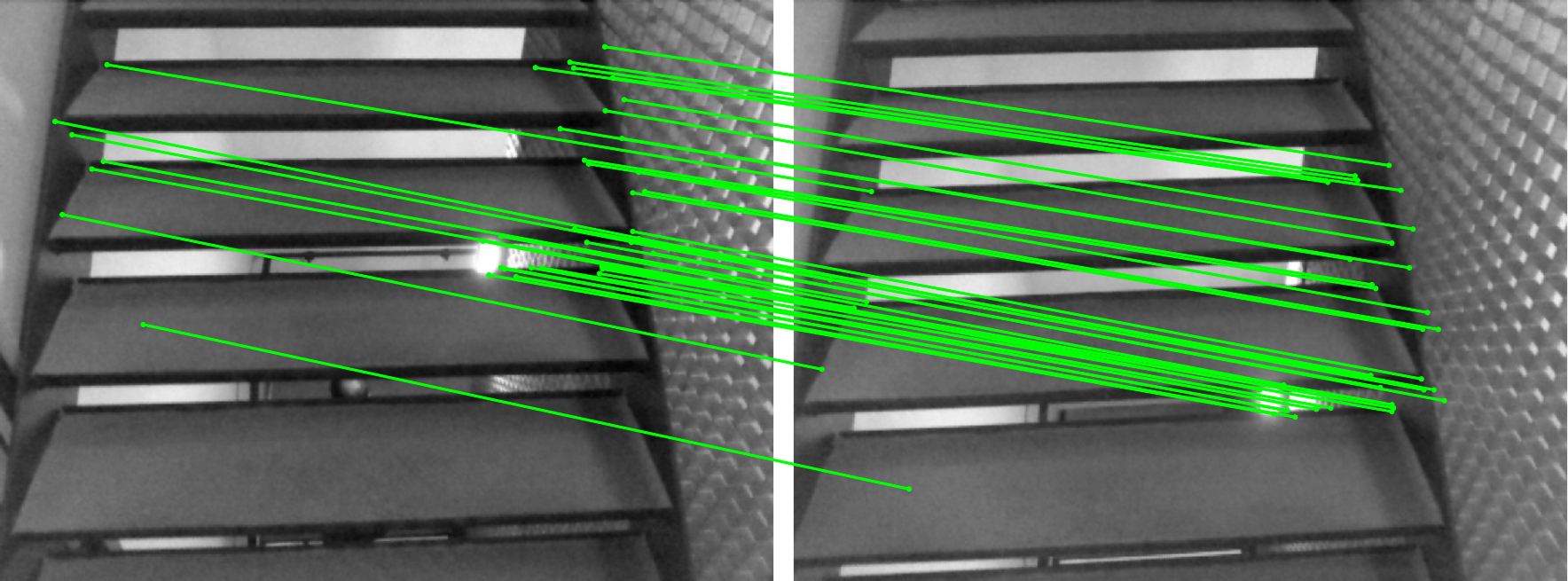} & \rotatebox{90}{\phantom{x}GlueStick} & \includegraphics[width=\sz\columnwidth]{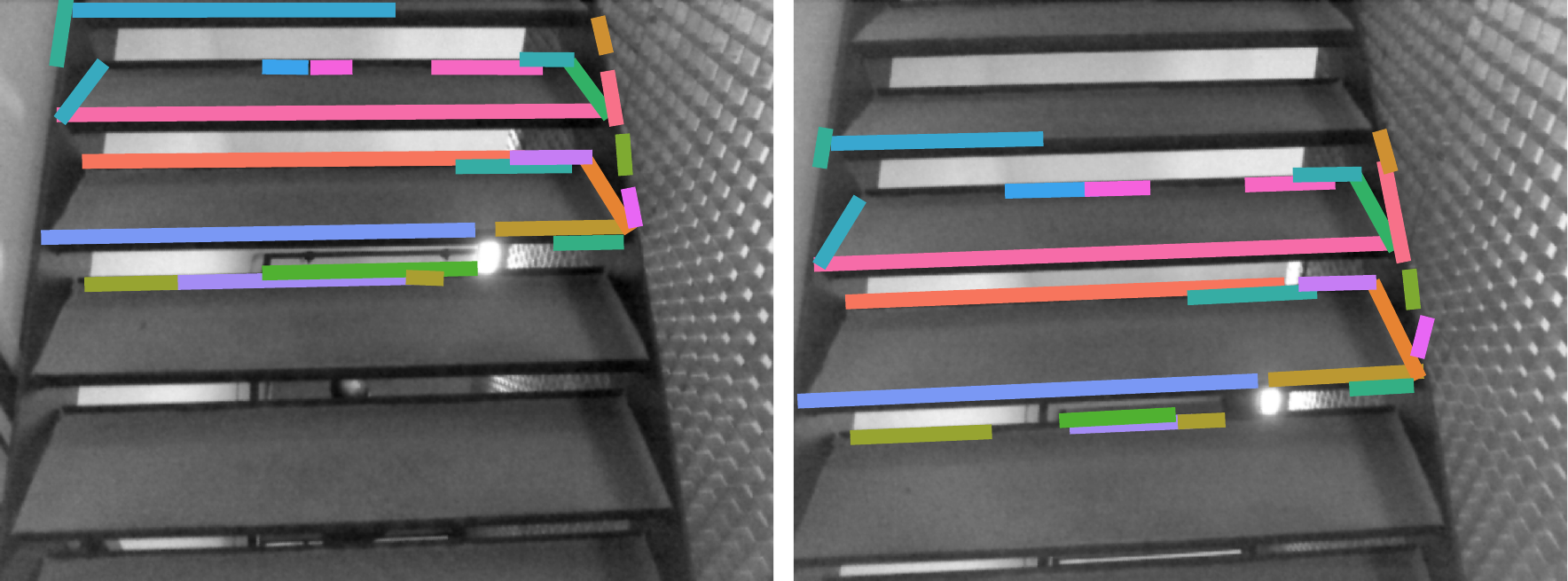} \\
	\end{tabular}
	\vspace{-0.2cm}
	\caption{\textbf{Correct matches on 7Scenes Stairs~\cite{7scenes}.} Lines can guide the point matching in very challenging scenarios with low texture and repeated patterns.}
	\label{fig:visual_examples}
\end{figure}

\section{Conclusion}
Matching points across two views and matching line segments are traditionally treated as two separate independent problems. In this work, we challenge this paradigm and present GlueStick, a learned matcher that jointly establishes correspondences between points and lines. By processing both types of features together, the network is able to propagate strong matches of either type to neighboring features that might have less discriminative appearance.

In our experiments, we show an improved matching performance across the board, for both points and lines. In particular for line matching, GlueStick provides a significant leap forward compared to the current descriptor-based state of the art. The key insight in our work is that line segments do not appear randomly scattered in the image, but rather form connected structures. This connectivity is explicitly encoded and exploited in our network architecture. Finally, we show that the improved matches we obtain directly translate to better results in downstream tasks such as homography and camera pose estimation.

\section*{Acknowledgement}
We would like to thank P.E. Sarlin and P. Lindenberger for their great support, as well as L. Cavalli, J.M. Buenaposada, and L. Baumela for helping to review this paper. V. Larsson was supported by the strategic research project ELLIIT.

\clearpage

\appendix

\twocolumn[
    {\centering \Large Supplementary Material \\[1ex]}
    \vspace*{3ex}
]

In the following, we provide additional details regarding GlueStick, our point-line matcher. \cref{sec:gt_generation_app} offers a visualization of the generation of our line ground truth. \cref{sec:additional_insights} gives additional insights and ablation studies motivating our choices. \cref{sec:experimental_details} specifies some experimental details to reproduce our experiments and brings additional results. \cref{sec:qualitative_examples} shows matching results, as well as failure cases of our method. Finally, \cref{sec:attention_visualization} provides visualizations of the attention for various kinds of nodes.

\section{Ground Truth Generation}
\label{sec:gt_generation_app}
Designing a line matching ground truth (GT) is challenging, due to partial occlusions and lack of repeatability of line detectors. We provide here some visualizations of the GT generation process.

\begin{figure}
    \centering
    \begin{subfigure}{\linewidth}
        \centering
        \includegraphics[width=0.50\linewidth]{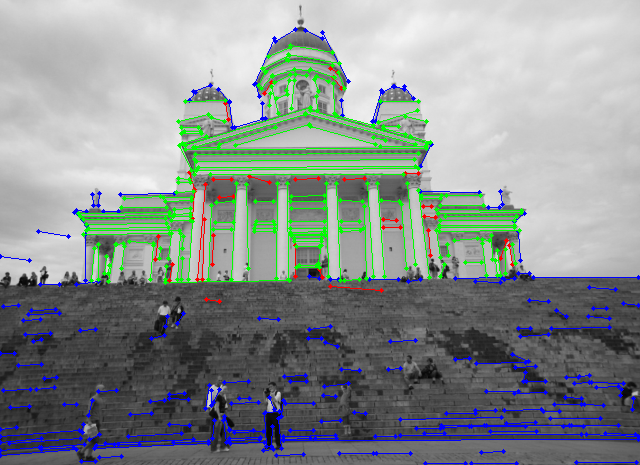}
        \includegraphics[width=0.46\linewidth]{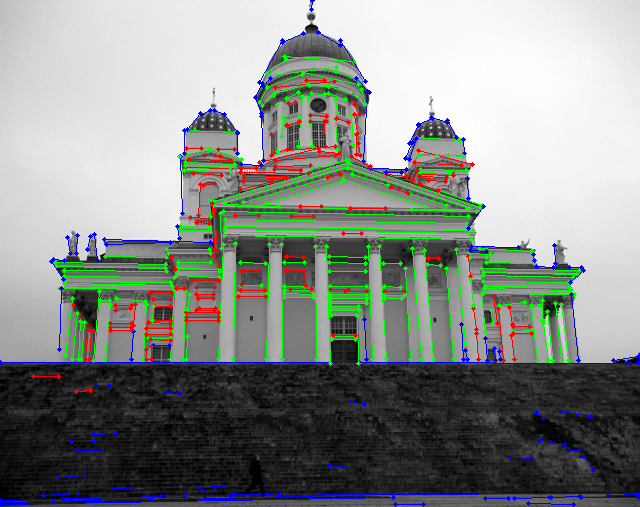}
        \caption{GT line assignations, shown as {\color{green}\texttt{MATCHED}}, {\color{red}\texttt{UNMATCHED}}, and {\color{blue}\texttt{IGNORED}}.}
    \end{subfigure}
    \hfill
    \begin{subfigure}{\linewidth}
        \centering
        \includegraphics[width=0.50\linewidth]{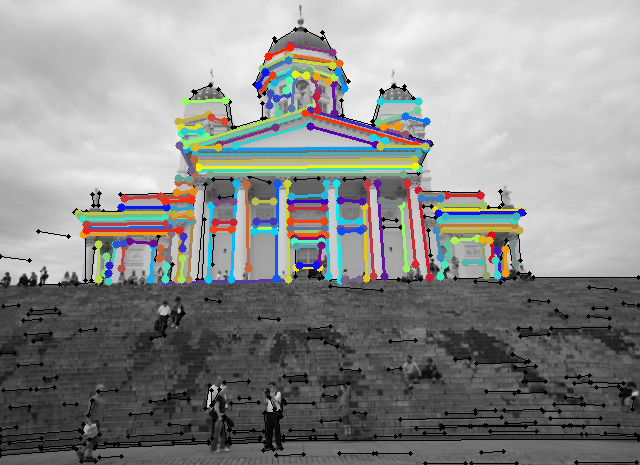}
        \includegraphics[width=0.46\linewidth]{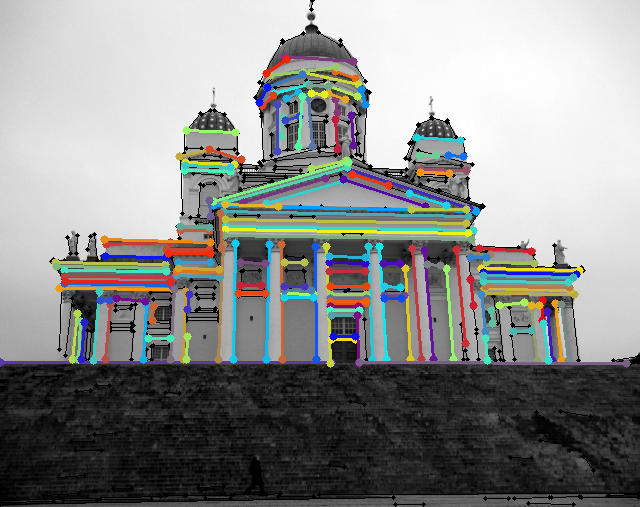}
        \caption{Each color identifies a match $(i, j) \in \mathcal{M}^{l}$ of the GT.}
    \end{subfigure}
    \caption{\textbf{Ground truth (GT) line assignations.} Examples of GT line matches. Note that blue lines are located in uncertain regions and depth discontinuities, and they are ignored during training.}
    \label{fig:gt-examples}
\end{figure}

\begin{figure*}
    \centering
    \begin{subfigure}{0.63\linewidth}
        \centering
        \includegraphics[width=1.0\linewidth]{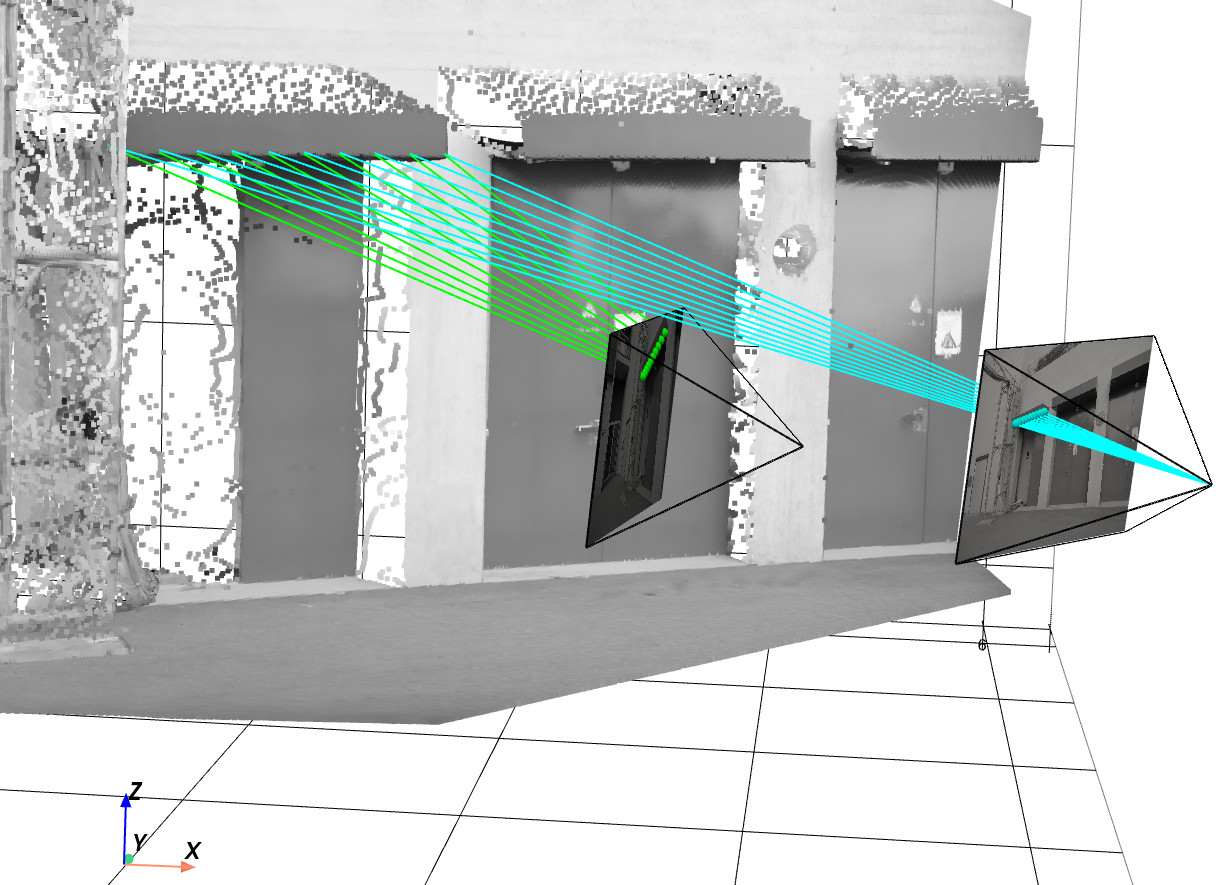}
        \caption{3D Visualization of a scene in ETH3D~\cite{Schops_2017_eth3d}, with the depth associated with each point and the camera poses.}
    \end{subfigure}
    \hfill
    \begin{subfigure}{0.33\linewidth}
        \centering
        \includegraphics[width=\linewidth]{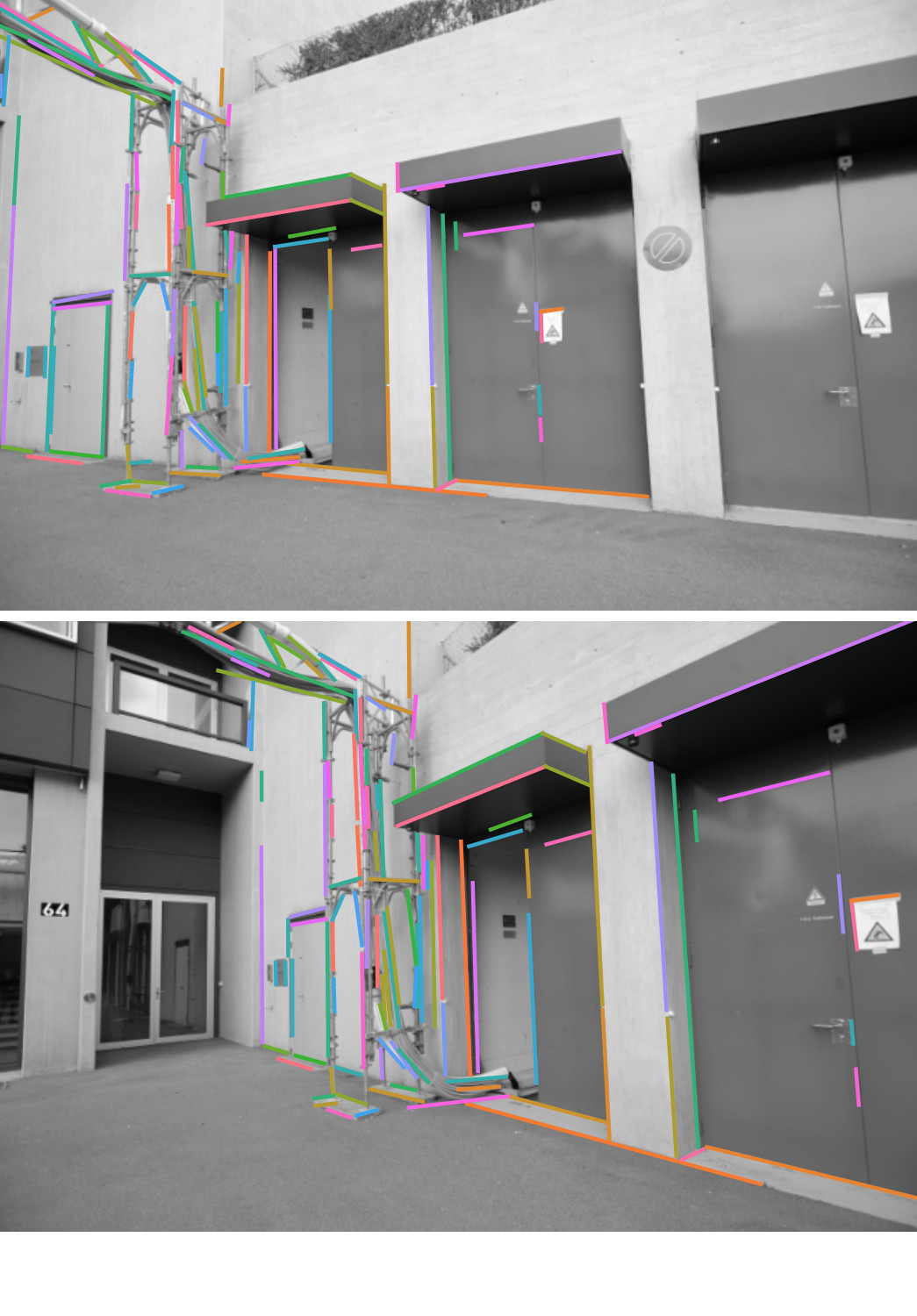}
        \vspace{-0.8cm}
        \caption{GT Matching result}
    \end{subfigure}
\caption{\textbf{Visualization of the ground truth (GT) generation.} To check if a pair of line segments $\mathbf{l}_i^{A}$ and  $\mathbf{l}_j^{B}$ correspond to each other, we sample points along the segment: cyan points in the right image of (a). Points are lifted using depth and re-projected in the second image: green points in the left image of (a). We use the number of points that lie close to the segment in the second image to build each entry $\mathbf{C}^B_{i, j}$ of the cost matrix. Together with the reciprocal matrix $\mathbf{C}^A$ we define an assignation problem whose solution is our GT shown in (b).}
\label{fig:gt-diagram}
\end{figure*}

\cref{fig:gt-examples} shows an example of the ground truth between two images. Line segments can be either \texttt{MATCHED} (green), \texttt{UNMATCHED} (red), or \texttt{IGNORED} (blue). The latter case happens when the depth along the line is too uncertain or when its reprojection in the other image is occluded. The generation process is also illustrated in \cref{fig:gt-diagram} for one pair of line segments. The advantage of the proposed method is that it recovers a large number of matches for each pair of images, providing a strong matching signal. In comparison to the method proposed in \cite{l2d2} which does robust 3D reconstruction of line segments, our method is faster and simpler. By avoiding the 3D line reconstruction step, we can train in larger scenes with potentially noisy depth, like the MegaDepth dataset~\cite{Li_2018_MegaDepth}.

\section{Additional Insights on GlueStick}
\label{sec:additional_insights}
In this section, we give extra insights and motivations for our design choices.

\noindent \textbf{Choice of the Line Segment Detector.}
In all our training and experiments, we used the Line Segment Detector (LSD)~\cite{gioi_2010_lsd} to extract line segments. 
For a certain application, such as indoor wireframe parsing~\cite{Huang_2018_wireframe}, learned methods largely overtake classic ones~\cite{dai2022fclip, suarez_2022_elsed, Xu_2021_letr, Xue_2020_hawp}. However, learned methods struggle to generalize this power to other contexts, tasks, or types of images. For this reason, we have chosen LSD as the generic method to train GlueStick. Furthermore, we believe that our line pre-processing, turning an unordered set of lines into a connected graph, is beneficial to make the endpoints more repeatable across views, thus potentially making LSD more repeatable.

We ran a small experiment to compute the line repeatability of different line detectors on the HPatches~\cite{hpatches_2017_cvpr} and ETH3D~\cite{Schops_2017_eth3d} datasets. We define line repeatability for a pair of images as the percentage between the number of line correspondences and the number of lines in the pair of images~\cite{mikolajczyk2005comparison}. We establish line correspondences between images with the protocol defined in \cref{sec:gt_generation} of the main paper and \cref{sec:gt_generation_app}.
\cref{tab:rep_exp} shows that the learned baseline F-Clip~\cite{dai2022fclip} obtains the highest repeatability, but detects few lines, due to the fact that it was trained on the ground truth lines of the Wireframe dataset~\cite{Huang_2018_wireframe}. On the contrary, LSD provides the best trade-off in terms of repeatability and number of lines. Thus, this good trade-off, as well as its low localization error and versatility, make LSD a very suitable choice for our approach.

\begin{table}
\centering
\scriptsize
\setlength{\tabcolsep}{3pt}
\begin{tabular}{lcccc}
    \toprule
    & \multicolumn{2}{c}{HPatches} & \multicolumn{2}{c}{ETH3D} \\
    \cmidrule(r{5pt}l{5pt}){2-3} 
    \cmidrule(r{5pt}l{5pt}){4-5} 
    & \#Lines  & Rep. (\%) & \#Lines  & Rep. (\%) \\
    \midrule
    LSD~\cite{gioi_2010_lsd} & 307 & 68.72 & 578 & 47.49 \\
    ELSED~\cite{suarez_2022_elsed} & 243 & 64.66 & 464 & 48.03 \\
    HAWP~\cite{Xue_2020_hawp} & 366 & 60.86 & 420 & 38.69 \\
    SOLD${}^2$~\cite{pautrat_2021_sold2} & 167 & 65.49 & 332 & 39.86 \\
    F-Clip~\cite{dai2022fclip} & 139 & 72.91 & 444 & 50.90 \\
    LETR~\cite{Xu_2021_letr} & 95 & 70.65 & 311 & 47.70 \\
    \bottomrule
\end{tabular}
\caption{\textbf{Line segment detection comparison.} We compare different line segment detectors in terms of their repeatability in the HPatches~\cite{hpatches_2017_cvpr} and ETH3D~\cite{Schops_2017_eth3d} datasets.}
\label{tab:rep_exp}
\end{table}

\begin{figure*}
    \centering
    \newcommand{\fvs}{0.1cm}
    \begin{subfigure}{0.235\linewidth}
        \centering
        \includegraphics[width=1.0\linewidth]{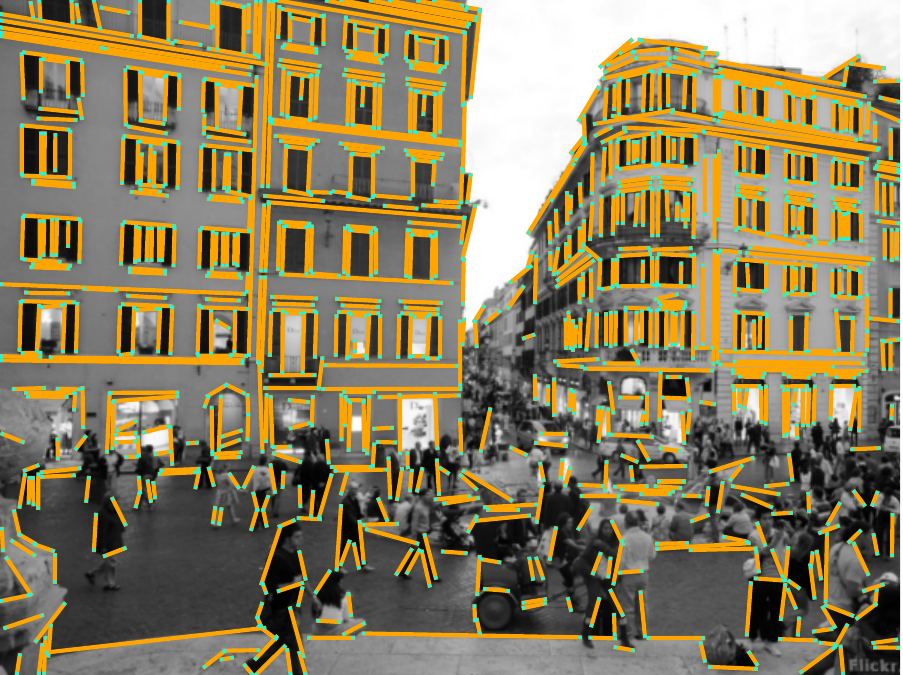}\vspace{\fvs}
        \includegraphics[width=1.0\linewidth]{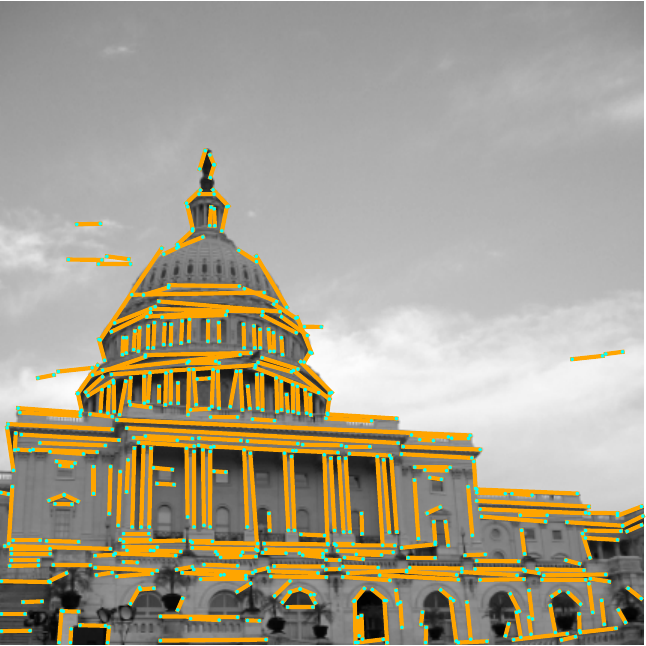}\vspace{\fvs}
        \includegraphics[width=1.0\linewidth]{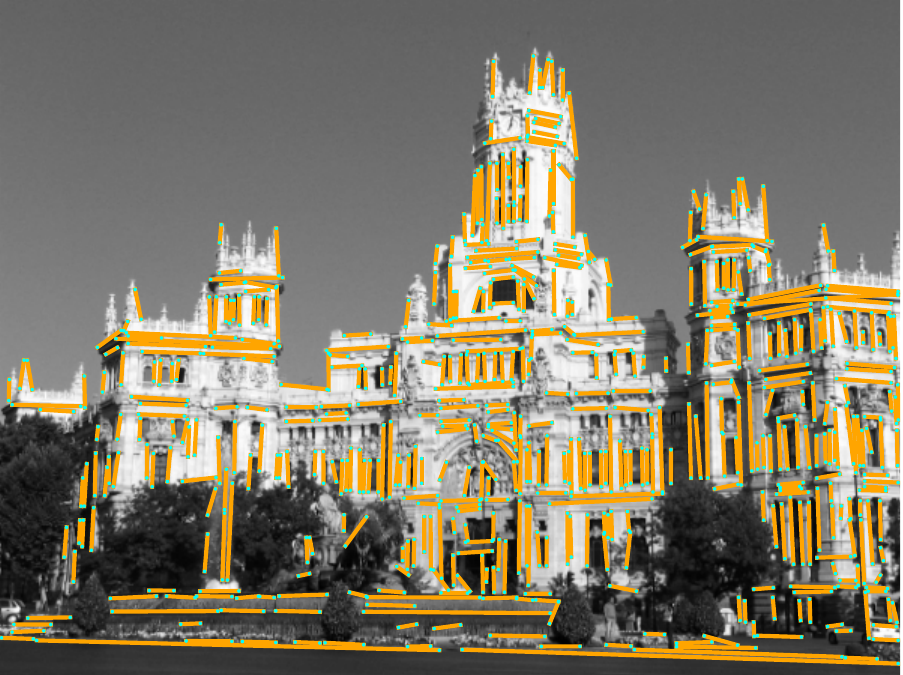}\vspace{\fvs}
        \includegraphics[width=1.0\linewidth]{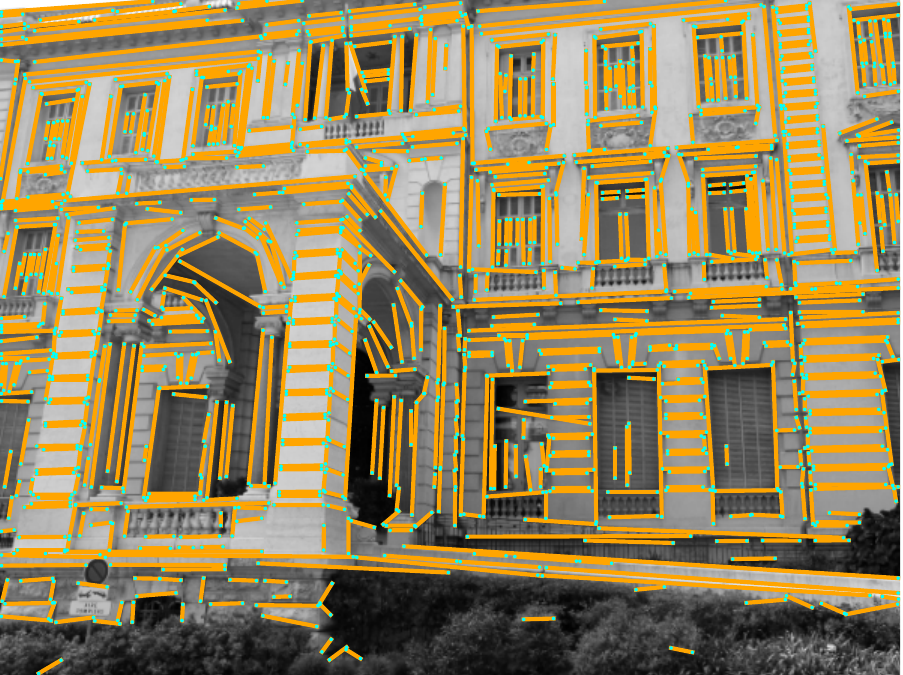}\vspace{\fvs}
        \includegraphics[width=1.0\linewidth]{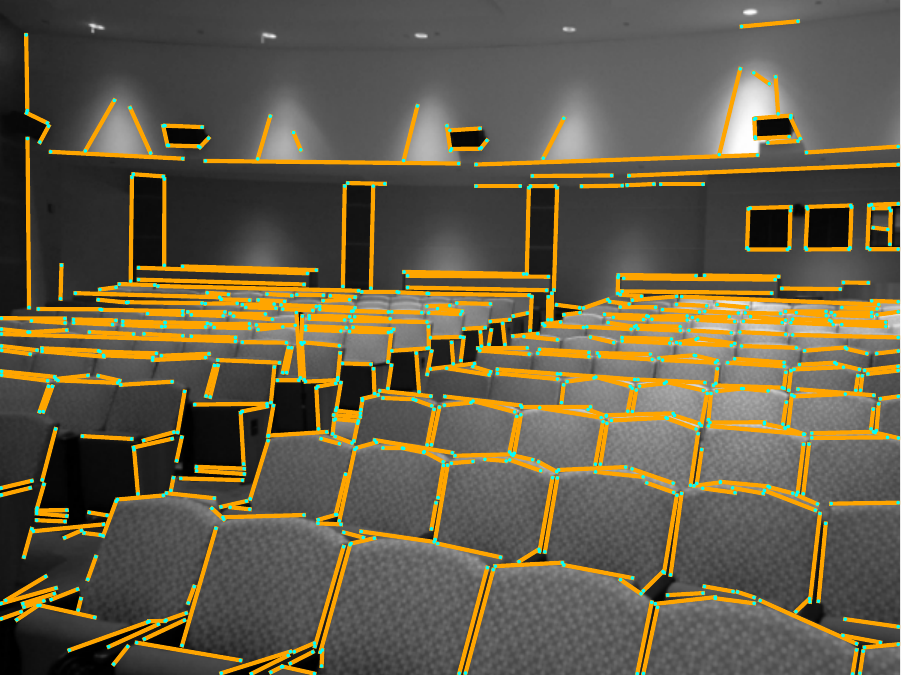}\vspace{\fvs}
        \includegraphics[width=1.0\linewidth]{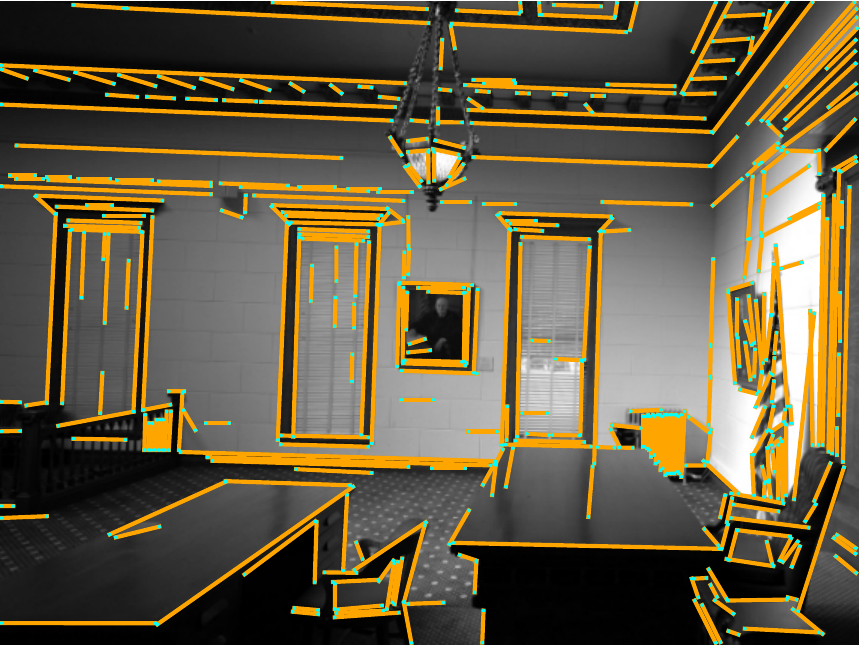}\vspace{\fvs}
        \caption{LSD~\cite{gioi_2010_lsd}}
    \end{subfigure}
    \hfill
    \begin{subfigure}{0.235\linewidth}
        \centering
        \includegraphics[width=1.0\linewidth]{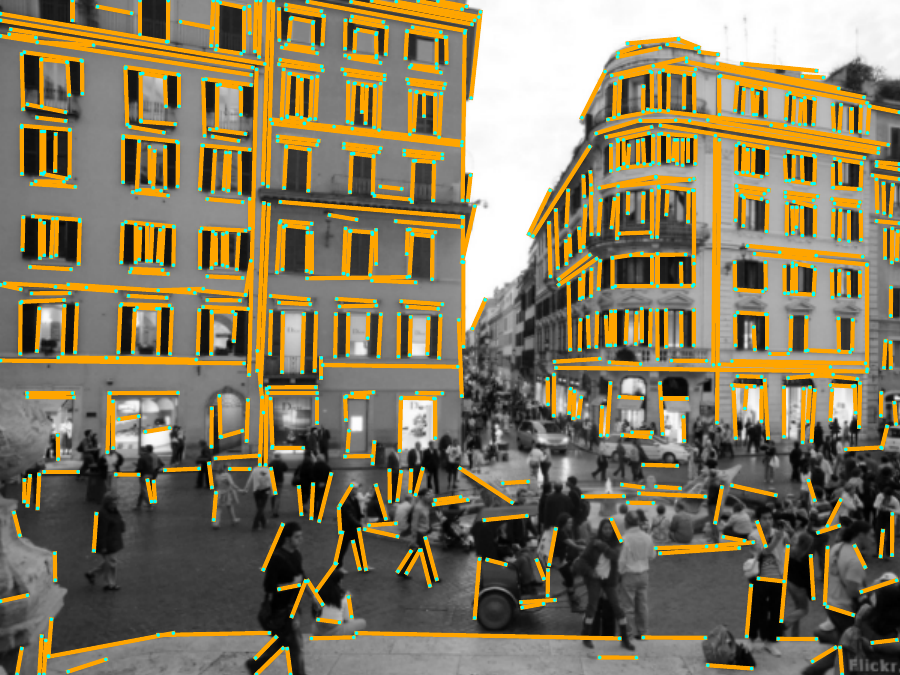}\vspace{\fvs}
        \includegraphics[width=1.0\linewidth]{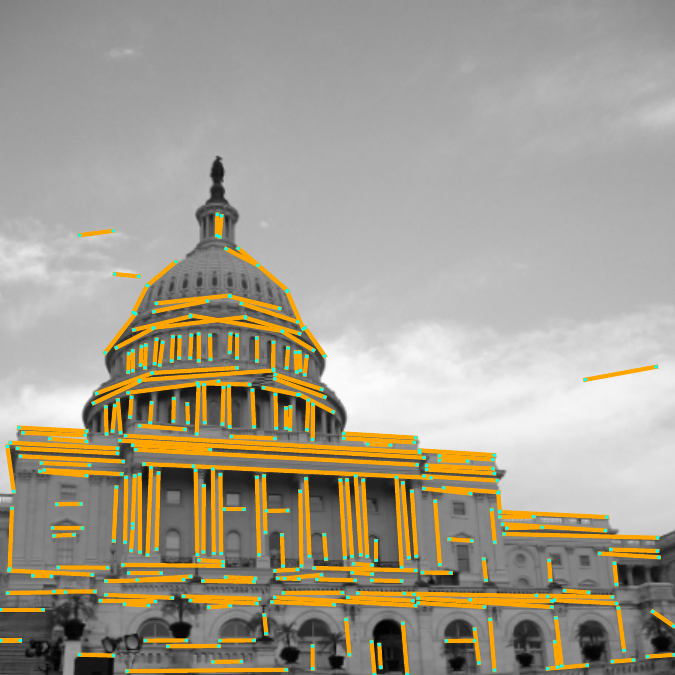}\vspace{\fvs}
        \includegraphics[width=1.0\linewidth]{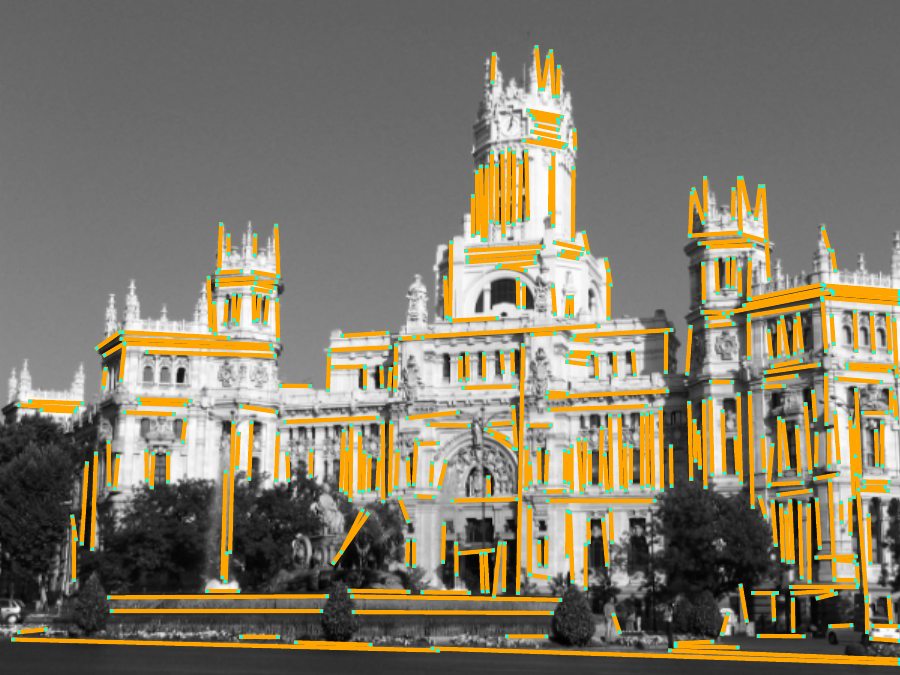}\vspace{\fvs}
        \includegraphics[width=1.0\linewidth]{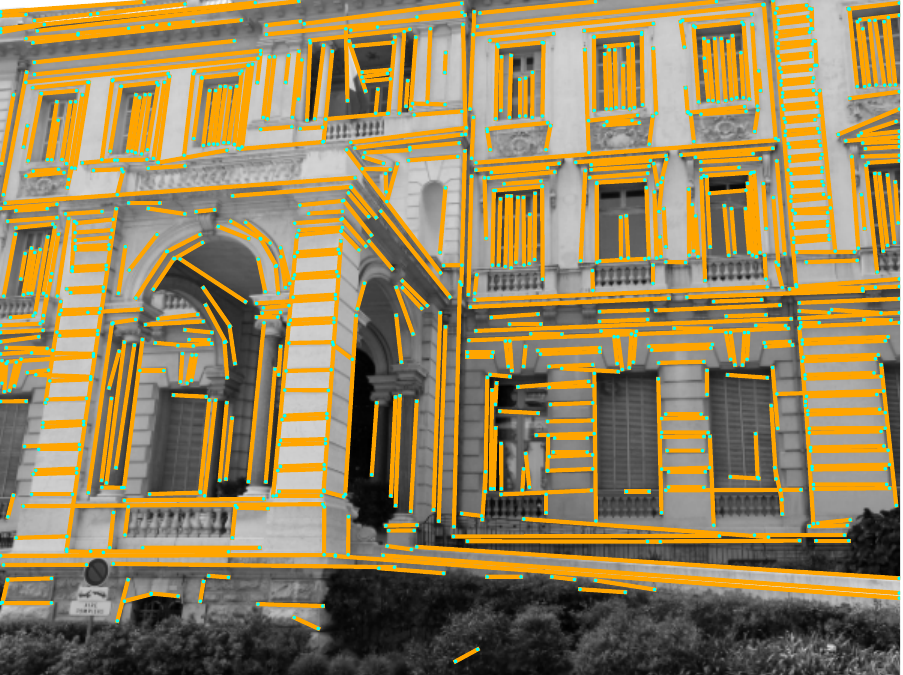}\vspace{\fvs}
        \includegraphics[width=1.0\linewidth]{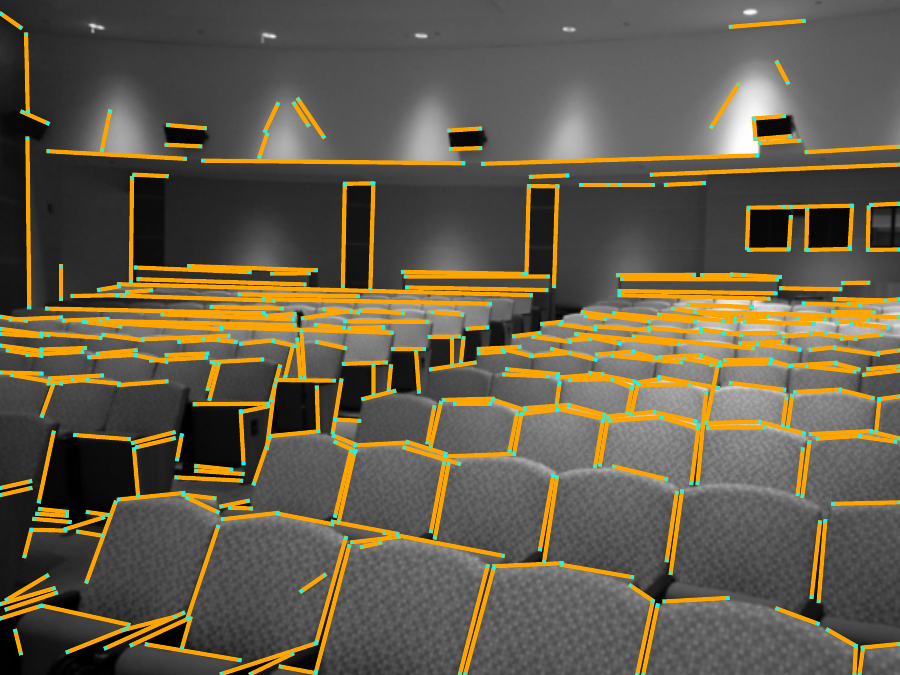}\vspace{\fvs}
        \includegraphics[width=1.0\linewidth]{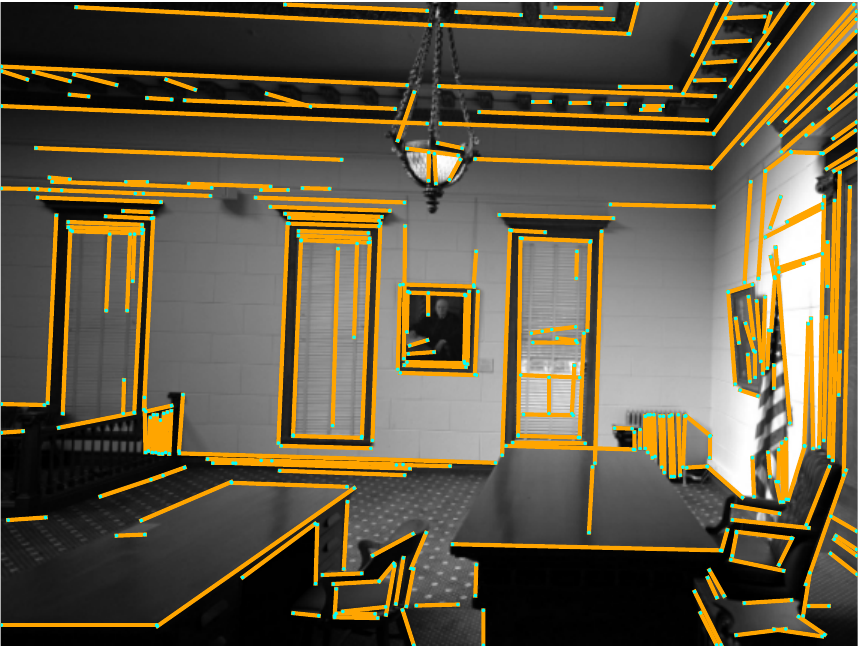}\vspace{\fvs}
        \caption{ELSED~\cite{suarez_2022_elsed}}
    \end{subfigure}
    \hfill
    \begin{subfigure}{0.235\linewidth}
        \centering
        \includegraphics[width=1.0\linewidth]{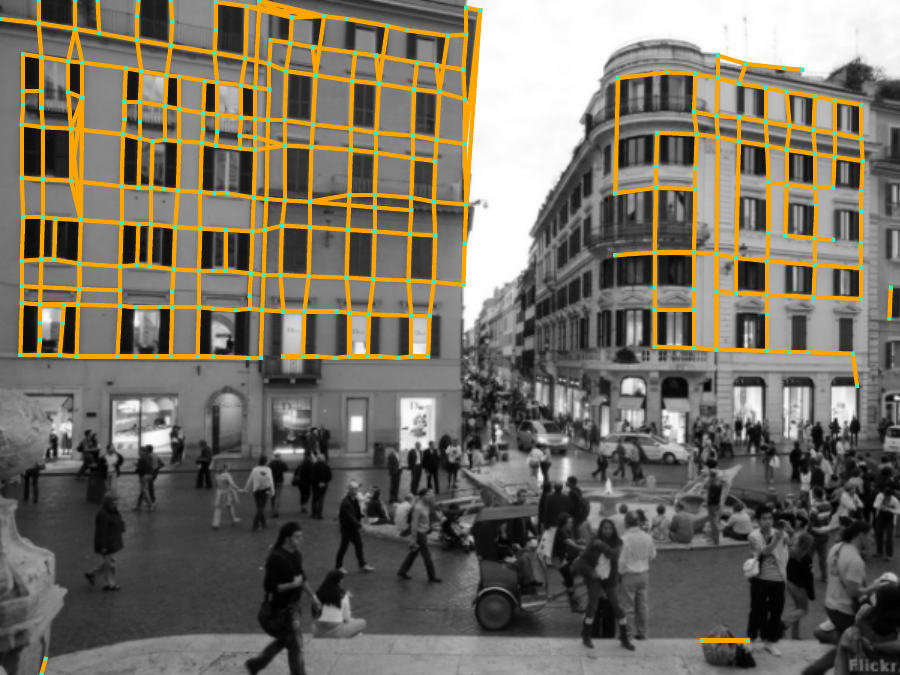}\vspace{\fvs}
        \includegraphics[width=1.0\linewidth]{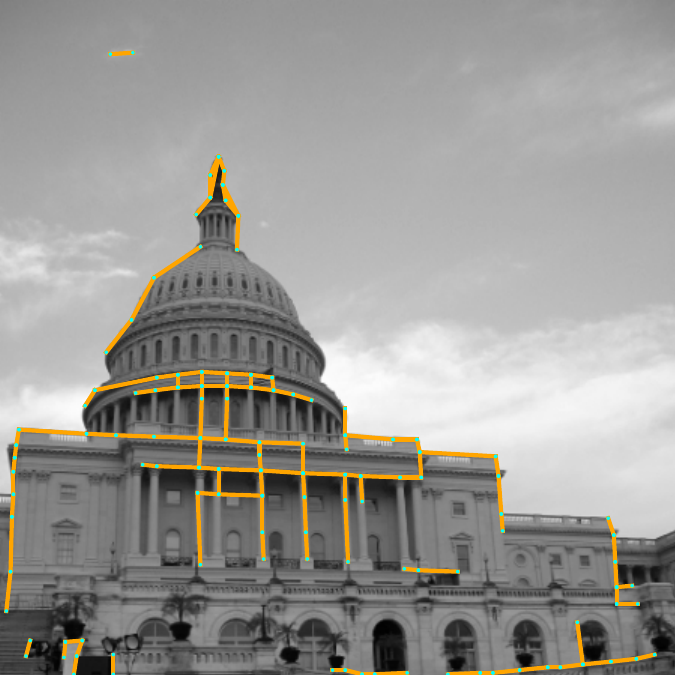}\vspace{\fvs}
        \includegraphics[width=1.0\linewidth]{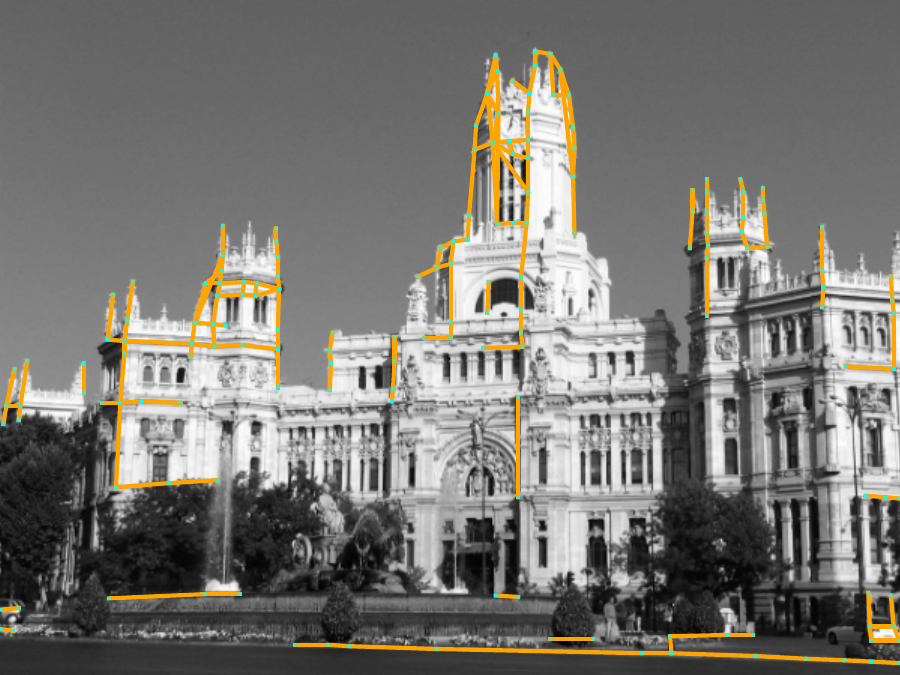}\vspace{\fvs}
        \includegraphics[width=1.0\linewidth]{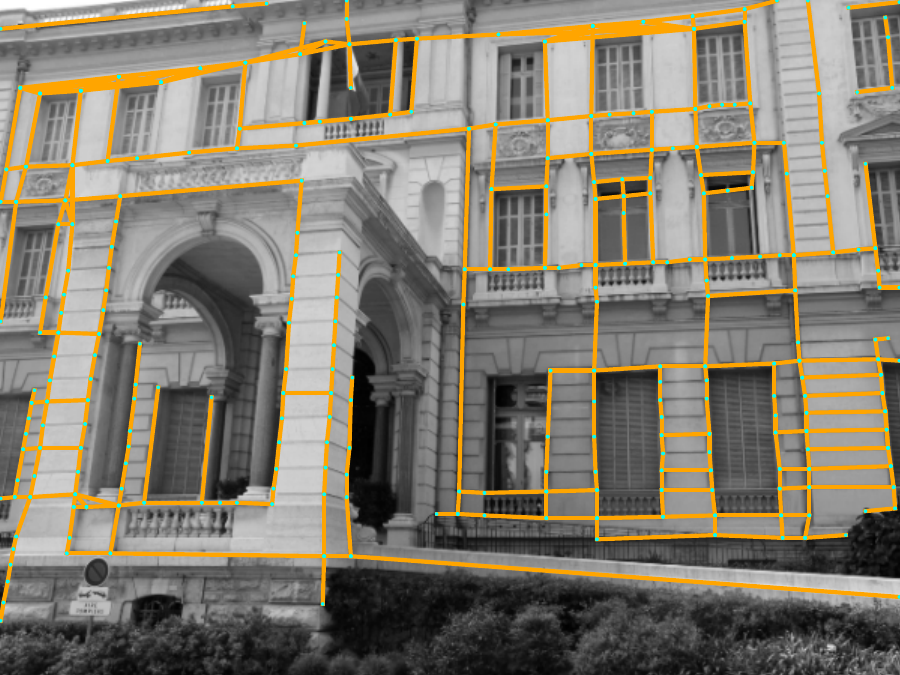}\vspace{\fvs}
        \includegraphics[width=1.0\linewidth]{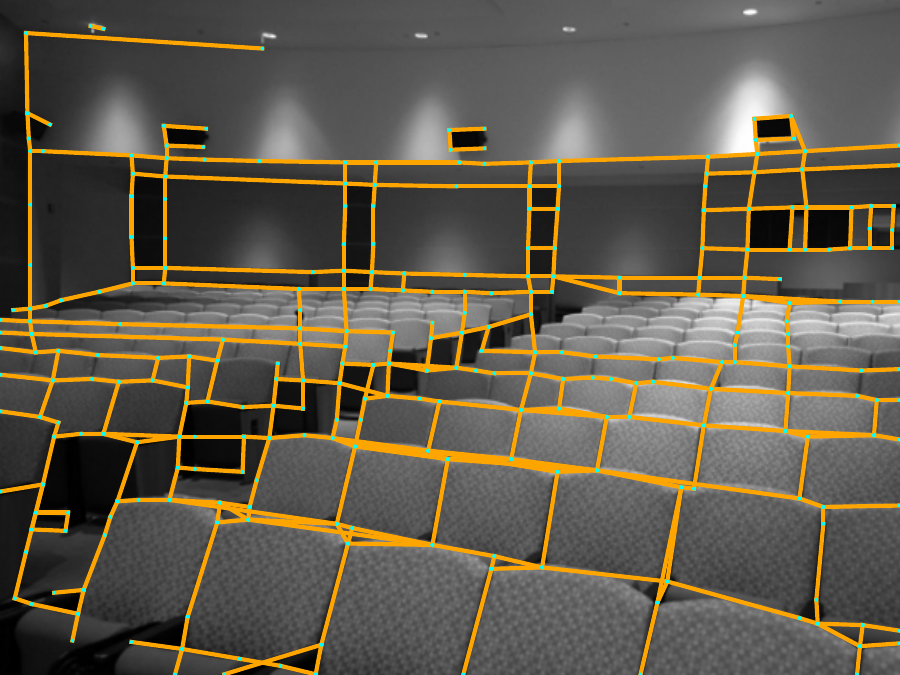}\vspace{\fvs}
        \includegraphics[width=1.0\linewidth]{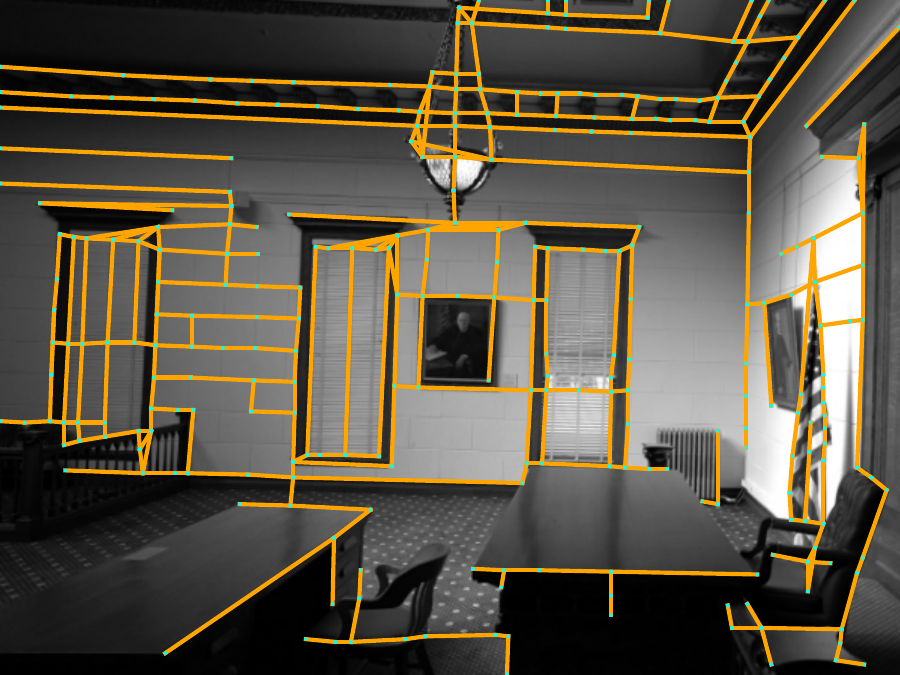}\vspace{\fvs}
        \caption{SOLD${}^{2}$\cite{pautrat_2021_sold2}}
    \end{subfigure}
    \hfill
    \begin{subfigure}{0.235\linewidth}
        \centering
        \includegraphics[width=1.0\linewidth]{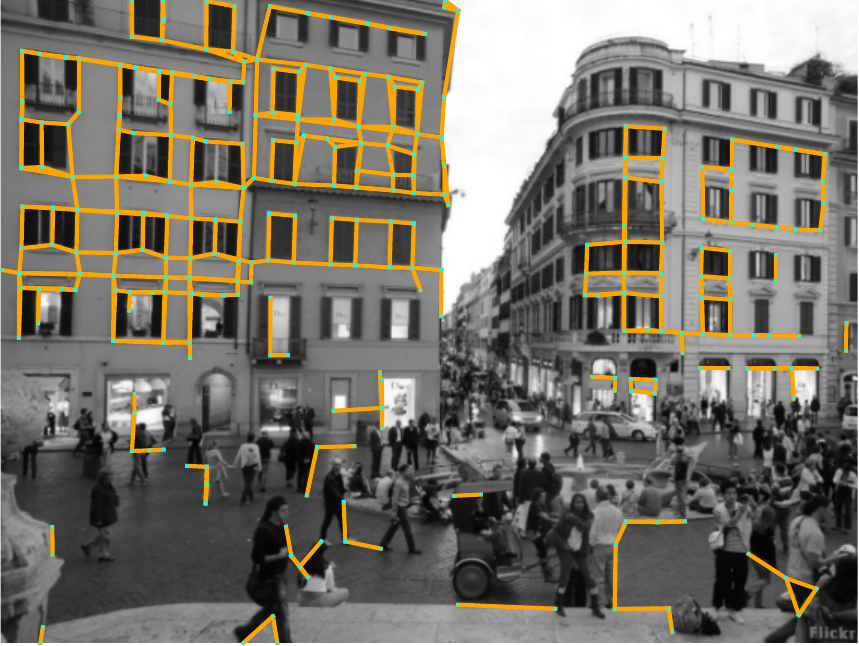}\vspace{\fvs}
        \includegraphics[width=1.0\linewidth]{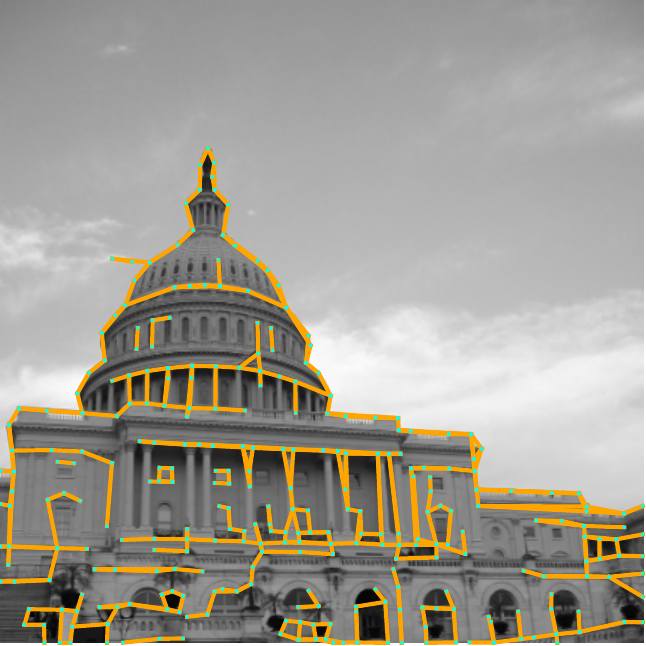}\vspace{\fvs}
        \includegraphics[width=1.0\linewidth]{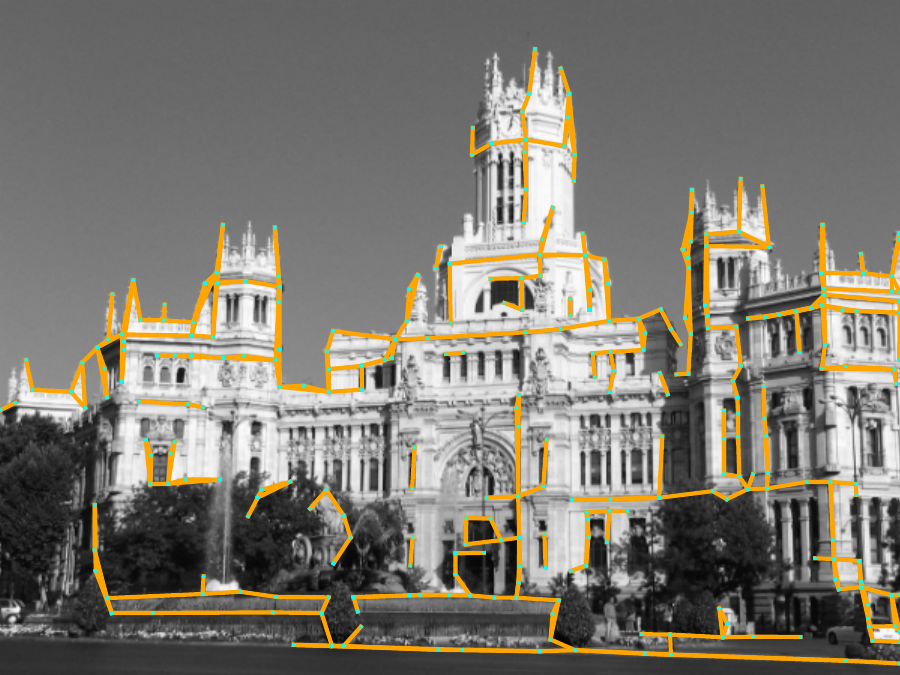}\vspace{\fvs}
        \includegraphics[width=1.0\linewidth]{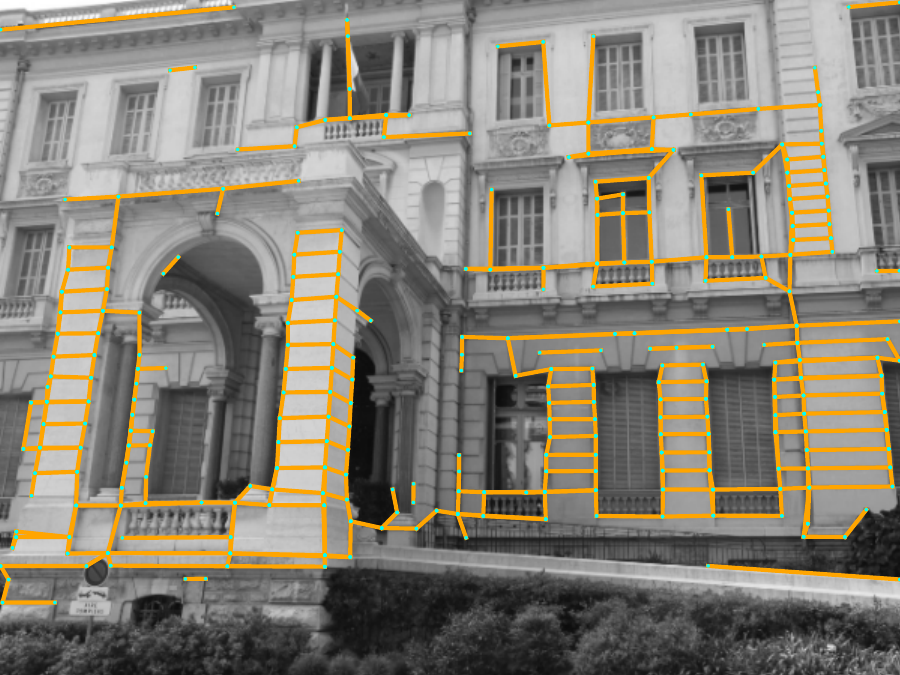}\vspace{\fvs}
        \includegraphics[width=1.0\linewidth]{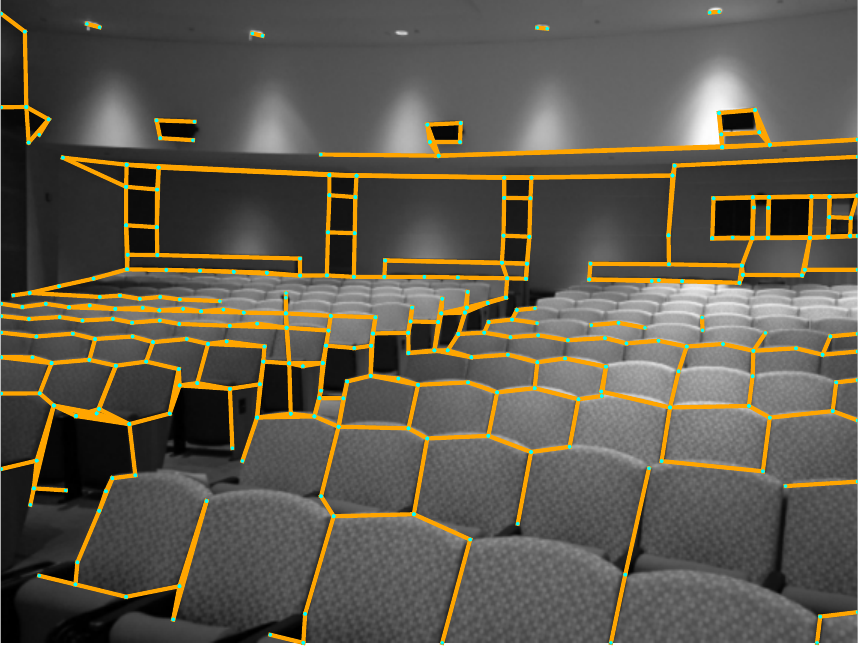}\vspace{\fvs}
        \includegraphics[width=1.0\linewidth]{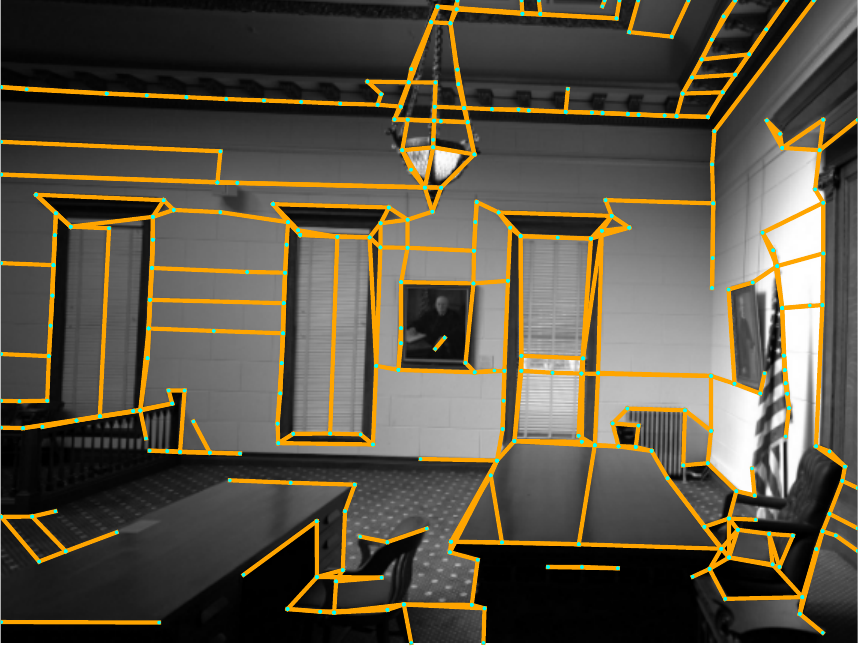}\vspace{\fvs}
        \caption{HAWP~\cite{Xue_2020_hawp}}
    \end{subfigure}
    \caption{\textbf{Comparison of line segment detectors.} Learned methods such as SOLD${}^{2}$~\cite{pautrat_2021_sold2} and HAWP~\cite{Xue_2020_hawp} may not generalize well in all situations, such as outdoors. Traditional ones such as LSD~\cite{gioi_2010_lsd} and ELSED~\cite{suarez_2022_elsed} produce a lot of overlapping segments and sometimes noisy ones. We decided to use LSD for its high versatility and accuracy.}
    \label{fig:line-detectors}
\end{figure*}

We provide in \cref{fig:line-detectors} additional visualizations of line segments for two traditional methods: LSD~\cite{gioi_2010_lsd} and ELSED~\cite{suarez_2022_elsed}, and two learned methods: SOLD${}^{2}$~\cite{pautrat_2021_sold2} and HAWP~\cite{Xue_2020_hawp}. While traditional ones sometimes detect noisy lines (for example in the sky), learned ones are often biased towards their training set and do not generalize very well to different settings, such as outdoor images.

\begin{figure}
    \centering
    \includegraphics[width=0.9\columnwidth]{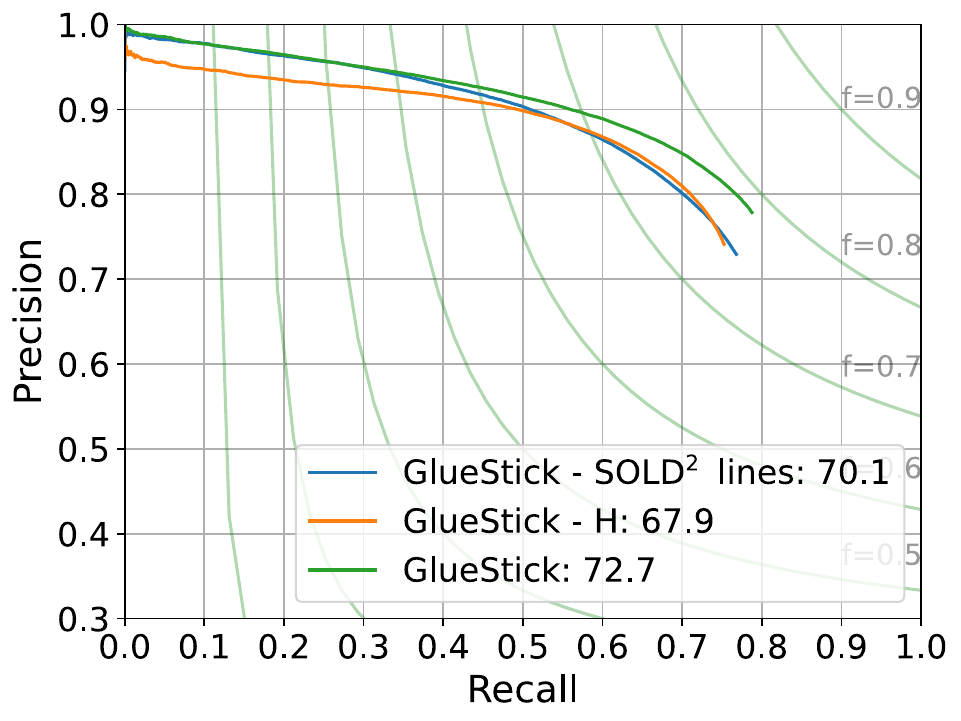}
    \caption{\textbf{Additional ablation study on the ETH3D dataset~\cite{Schops_2017_eth3d}.} We report the precision-recall curve of the line matching, as well as Average Precision (AP) in the legend. Our final \textit{GlueStick} model running with LSD~\cite{gioi_2010_lsd} lines is compared to its pre-trained version on homographies (\textit{GlueStick - H}), and the final model using SOLD${}^{2}$ lines~\cite{pautrat_2021_sold2} (\textit{GlueStick - SOLD${}^{2}$ lines}).}
    \label{fig:additional_ablation}
\end{figure}

However, it is important for GlueStick to generalize and perform well with other line segment detectors. In \cref{fig:additional_ablation} we run GlueStick using either LSD or SOLD${}^{2}$~\cite{pautrat_2021_sold2} lines, and we evaluate the precision-recall of both methods on the ETH3D dataset~\cite{Schops_2017_eth3d}. The latter are generic lines extracted by a deep network, with strong repeatability and low localization error~\cite{pautrat_2021_sold2}. 
It can be seen from the precision-recall curves that 1) our GlueStick model trained on LSD lines is able to generalize to other lines such as SOLD${}^{2}$~\cite{pautrat_2021_sold2}, and 2) the performance is slightly better with LSD lines. This is reasonable, since GlueStick was already trained on these lines. In summary, we chose LSD as base detector for downstream tasks since it remains one of the most accurate detector currently available, by directly relying on the image gradient at a sub-pixel level. 

\noindent \textbf{Effect of the Fine-tuning.}
Again in \cref{fig:additional_ablation}, we compare our final GlueStick model with its pre-trained version on homographies, \textit{GlueStick - H}. The plot shows that pre-training on homographies is already sufficient to get very high performance on ETH3D - better than the previous state-of-the-art line matchers. Fine-tuning on MegaDepth~\cite{Li_2018_MegaDepth} with real viewpoint changes can however further improve the robustness of our matcher, as demonstrated by the stronger performance of the final model.

\begin{figure}
    \centering
    \includegraphics[width=0.9\columnwidth]{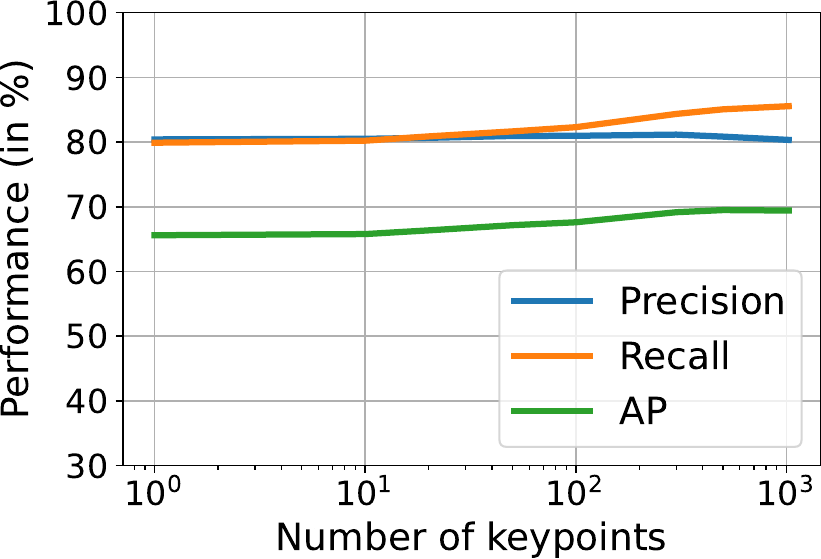}
    \caption{\textbf{Analysis of the dependence on keypoints.} We run GlueStick on 1000 image pairs warped by a homography (taken from the 1M distractor images of \cite{radenovic2018}), with varying numbers of keypoints, and report the precision, recall and Average Precision (AP) of the line matching. GlueStick robustly matches lines even when few or no keypoints are present.}
    \label{fig:kp_dependence}
\end{figure}

\noindent \textbf{Dependence on Point Matches.}
When jointly matching two kinds of features, one caveat is often that one type of feature takes the lead and the other relies mainly on the first one. While we know from SuperGlue~\cite{sarlin_2020_superglue} that point-only matching is already very strong on its own, we show here that our architecture is very robust to the absence of keypoints and that line-only matching is still possible. We ran an evaluation of the precision, recall, and average precision (AP) of the line matching on 1000 validation images of our homography dataset (images taken from the 1M distractor images of \cite{radenovic2018}), and tested different maximum numbers of keypoints per image. The results showed in \cref{fig:kp_dependence}, highlight that our line matching is extremely robust to the lack of keypoints. The precision remains indeed constant, and the recall and AP are decreasing by at most 5\% when switching from 1000 keypoints to no keypoints. Thus, this study confirms that our matcher can be used in texture-less areas where no keypoints are present, and is still able to match lines with high accuracy.

\begin{figure}
    \centering
    \includegraphics[width=0.9\columnwidth]{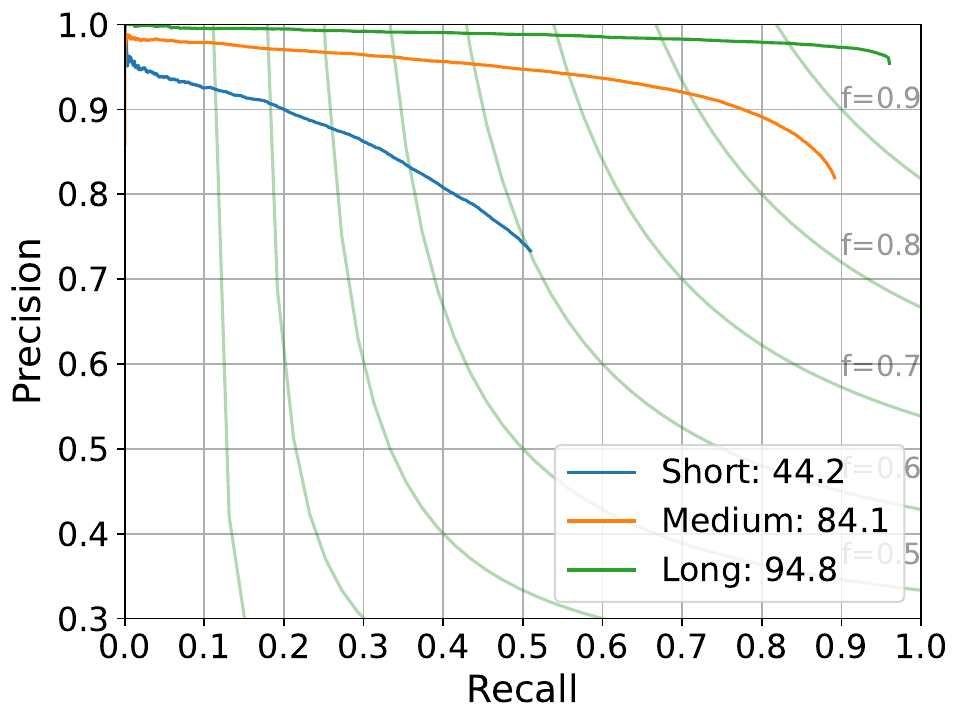}
    \caption{\textbf{Analysis of the impact of line length.} We run GlueStick on the ETH3D dataset~\cite{Schops_2017_eth3d} and evaluate separately the matching of Short, Medium, and Long lines. The best performance is obtained for long lines, as they are more stable than short ones. GlueStick can thus match long lines even with an endpoint representation.}
    \label{fig:line_length}
\end{figure}

\noindent \textbf{Impact of the Line Length.}
While we adopted a line representation based on the endpoints, one may wonder whether GlueStick can handle very long lines, and how it performs with respect to the line length. We studied this on the ETH3D~\cite{Schops_2017_eth3d} by categorising lines into three categories of length (in pixels): \textit{Short} ($[0, 50)$), \textit{Medium} ($[50, 150)$), and \textit{Long} ($[150, +\infty)$). The results are shown in \cref{fig:line_length}. It can be seen that the best performance is obtained for long lines, showing that GlueStick is still able to match lines even without context in the middle of the line. This result is due to the fact that long lines are more stable across views, while short ones are often noisy and not very repeatable.

\begin{figure}
    \centering
    \includegraphics[width=0.9\columnwidth]{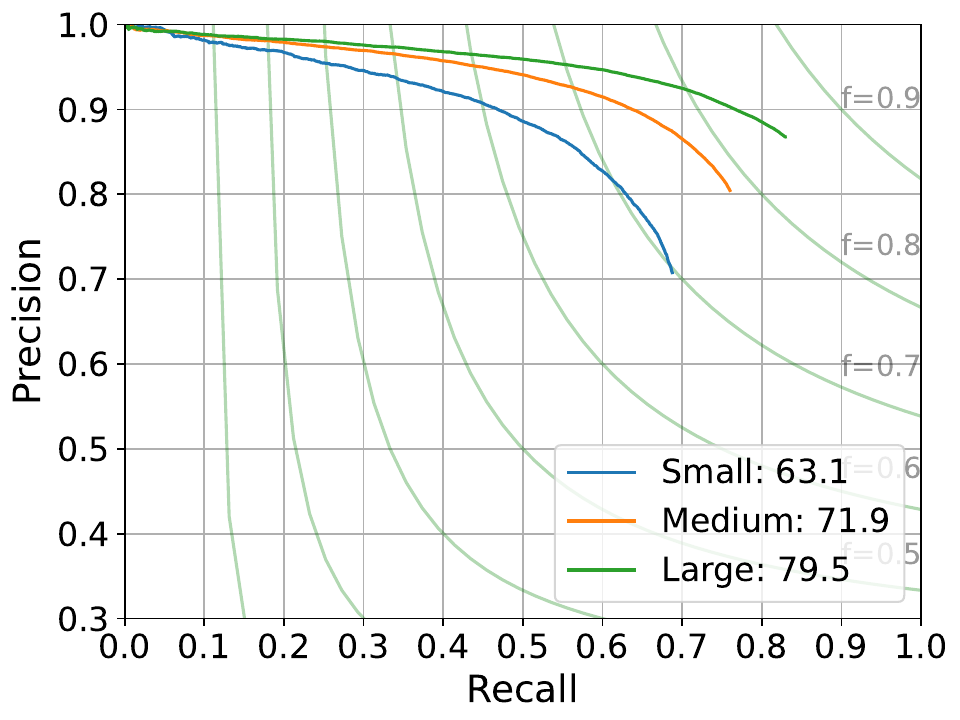}
    \caption{\textbf{Analysis of the impact of the image overlap.} We run GlueStick on the ETH3D dataset~\cite{Schops_2017_eth3d} and classify image pairs into three categories: Small, Medium, and Large overlap. While the performance of GlueStick decreases with smaller overlaps, it is able to maintain a high performance for all kinds of overlaps.}
    \label{fig:image overlap}
\end{figure}

\noindent \textbf{Robustness to Small Image Overlaps.}
The image overlap and scale changes between images can play a large role in matching. To study the effect of image overlap on GlueStick, we revisited our line matching experiment on the ETH3D dataset~\cite{Schops_2017_eth3d} and separated the pairs of images into three categories of image overlap: \textit{Small} ($[0, 0.33)$), \textit{Medium} ($[0.33, 0.66)$), and \textit{Large} ($[0.66, 1]$). Overlap is defined as the proportion of pixels falling into the other image after reprojection. It is computed symmetrically between the two images, and the minimum of the two values is kept. Results are available in \cref{fig:image overlap}. While the performance naturally decreases with smaller overlaps, GlueStick maintains a strong performance on such hard cases.

\section{Experimental Details}
\label{sec:experimental_details}
In this section, we first provide additional baselines to the experiment on ScanNet for homography and relative pose estimation. Secondly, we give details and visualizations of the pure rotation estimation between two image pairs, and its application to image stitching. Thirdly, we provide a comparison of methods on homography estimation on the HPatches dataset~\cite{hpatches_2017_cvpr}. Finally, we give the full results of visual localization on the 7Scenes dataset~\cite{7scenes}, though the performance on most scenes is already saturated.

\subsection{Relative Pose through Homographies}

The experiment in Tab. 1 of the main paper was meant to evaluate the quality of homographies retrieved from points, lines or points+lines features. Homographies were evaluated by decomposing into the corresponding relative pose and evaluating the latter on the ScanNet dataset~\cite{dai2017scannet}. For the sake of completeness, we also report here the results that would be obtained with a more efficient method to obtain the relative pose: the 5-point algorithm to obtain the essential matrix~\cite{DBLP:conf/cvpr/Nister03}, later decomposed as a relative pose. \cref{tab:scannet_rel_pose_E} demonstrates that the quality of relative poses retrieved through essential matrices is much higher than with homographies - as could be expected. GlueStick remains nevertheless the top performing method among all baselines. Note that we use here the outdoor models for all methods, for fairness reasons as GlueStick was only trained on outdoor data.

\begin{table}
    \centering
    \scriptsize
    \setlength{\tabcolsep}{5pt}
    \begin{tabular}{clcc}
        \toprule
         & & Pose error ($\downarrow$) & Pose AUC ($\uparrow$) \\
        \midrule
        \multirow{3}{*}{Pose from H} & SuperGlue (SG)~\cite{sarlin_2020_superglue} & 18.1 & 15.6 / 29.8 / 39.4 \\
         & LoFTR~\cite{sun_2021_loftr} & 16.8 & 15.8 / 30.9 / 41.4 \\
         & GlueStick & \textbf{14.1} & \textbf{19.3 / 35.4 / 46.0} \\
        \midrule
        \multirow{3}{*}{Pose from E} & SuperGlue (SG)~\cite{sarlin_2020_superglue} & 8.6 & 30.5 / 46.0 / 54.1 \\
         & LoFTR~\cite{sun_2021_loftr} & 11.7 & 23.6 / 39.6 / 48.4 \\
         & GlueStick & \textbf{8.4} & \textbf{30.9 / 46.8 / 55.1} \\
        \bottomrule
    \end{tabular}
    \caption{\textbf{Using essential matrices instead of homographies on ScanNet~\cite{dai2017scannet}.} While our experiment on ScanNet is meant to evaluate homographies, we display here the results that would be obtained when using essential matrices to get a relative pose, instead of homographies. We report the median pose error in degrees, as well as the AUC at 10$^\circ$ / 20$^\circ$ / 30$^\circ$ error. Essential matrices are naturally more robust and obtain better results when evaluated on relative pose estimation.}
    \label{tab:scannet_rel_pose_E}
\end{table}

\subsection{Pure Rotation Estimation}

In this section, we describe the details of the pure rotation algorithm used to perform the experiment of \cref{sec:rotation_rel_pose} of the main paper. Inside a Hybrid RANSAC~\cite{camposeco2018hybrid}, we design minimal and least square solvers to estimate the rotation based on points, lines, or a combination of both.

\begin{figure*}
        \centering
        \begin{subfigure}{0.33\linewidth}
            \centering
            \includegraphics[width=\linewidth]{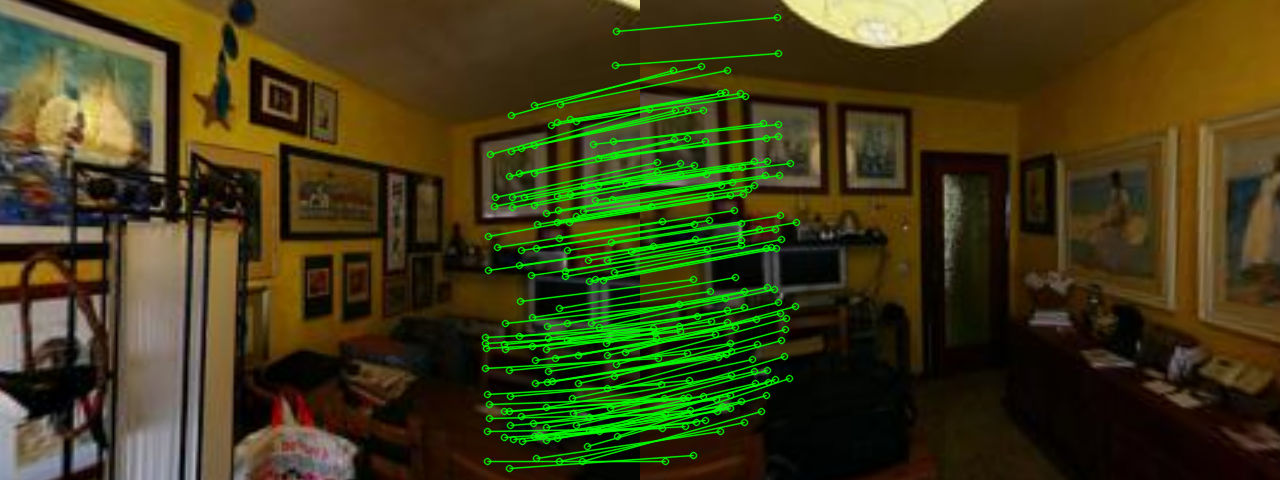}
            \includegraphics[width=\linewidth]{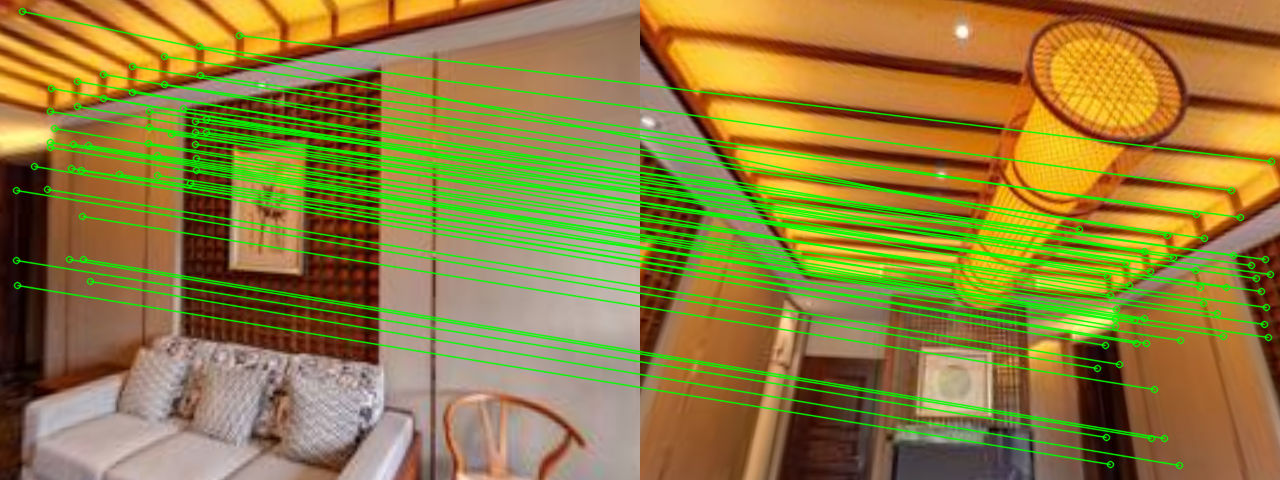}
            \includegraphics[width=\linewidth]{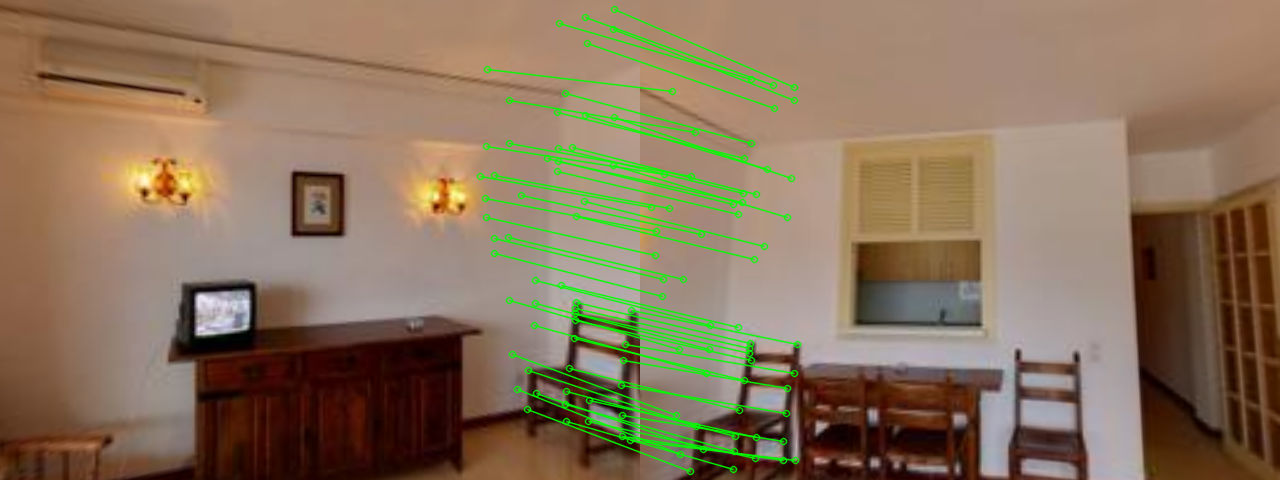}
            \includegraphics[width=\linewidth]{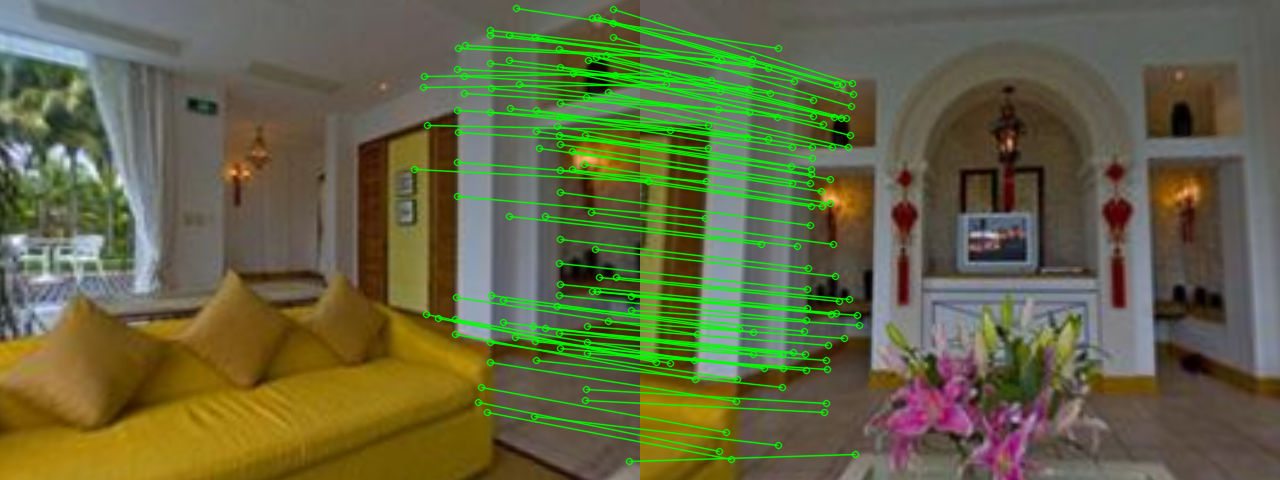}
            \includegraphics[width=\linewidth]{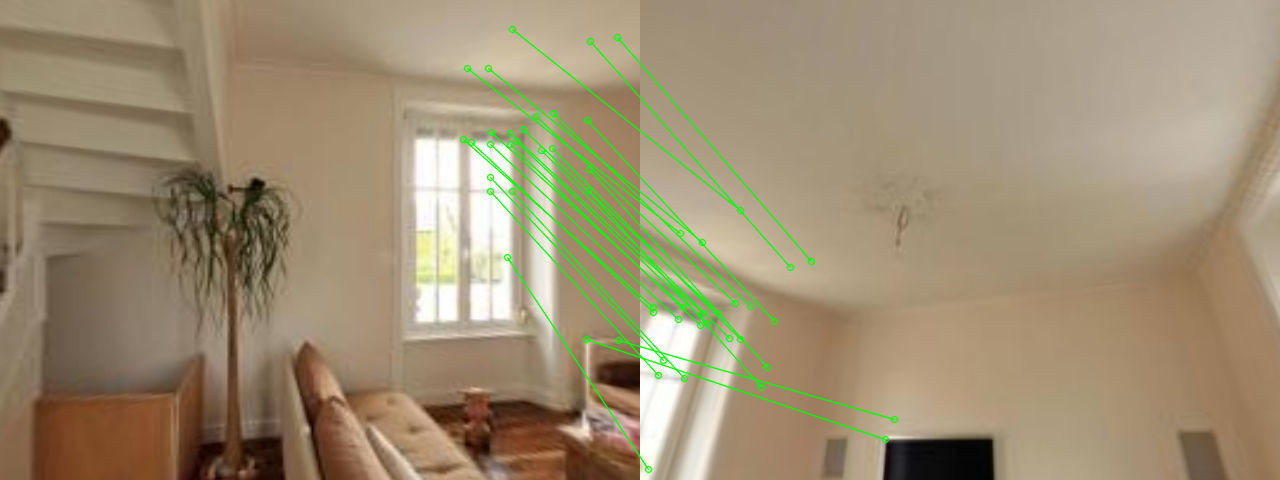}
            \caption{\small Point matches}
        \end{subfigure}
        \hfill
        \begin{subfigure}{0.33\linewidth}
            \centering
            \includegraphics[width=\linewidth]{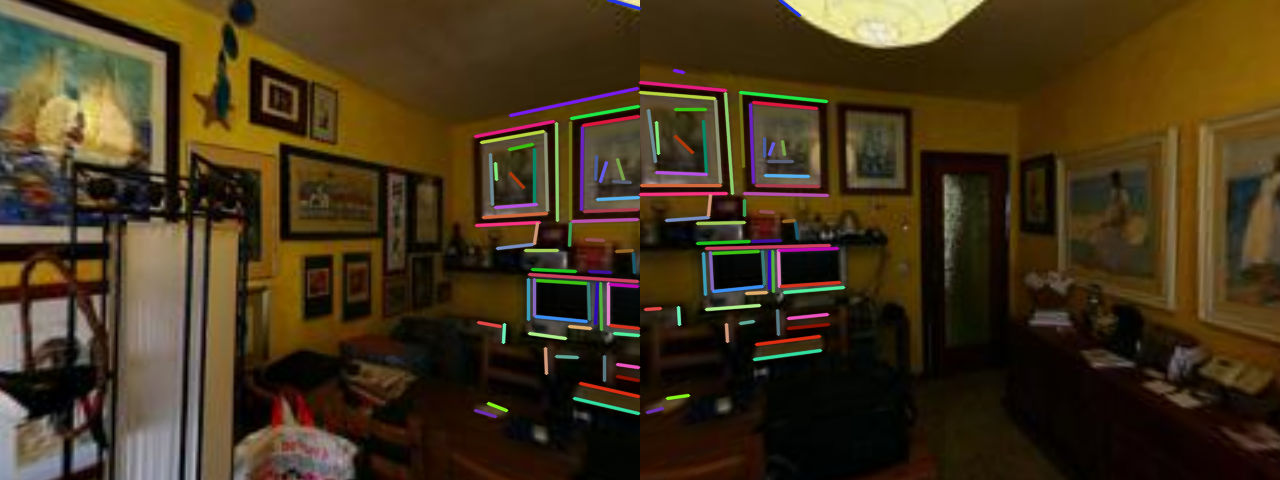}
            \includegraphics[width=\linewidth]{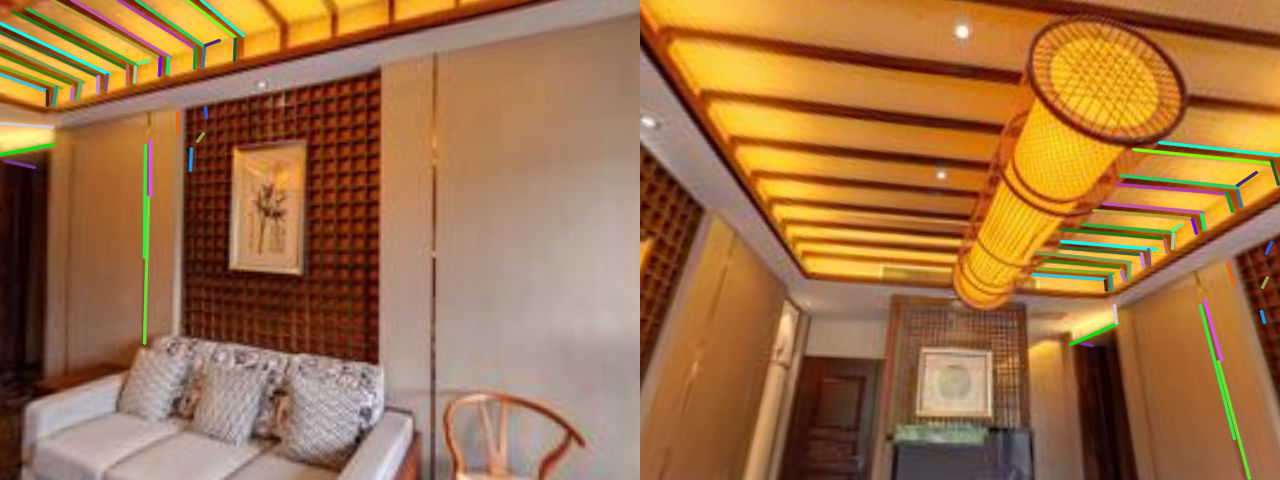}
            \includegraphics[width=\linewidth]{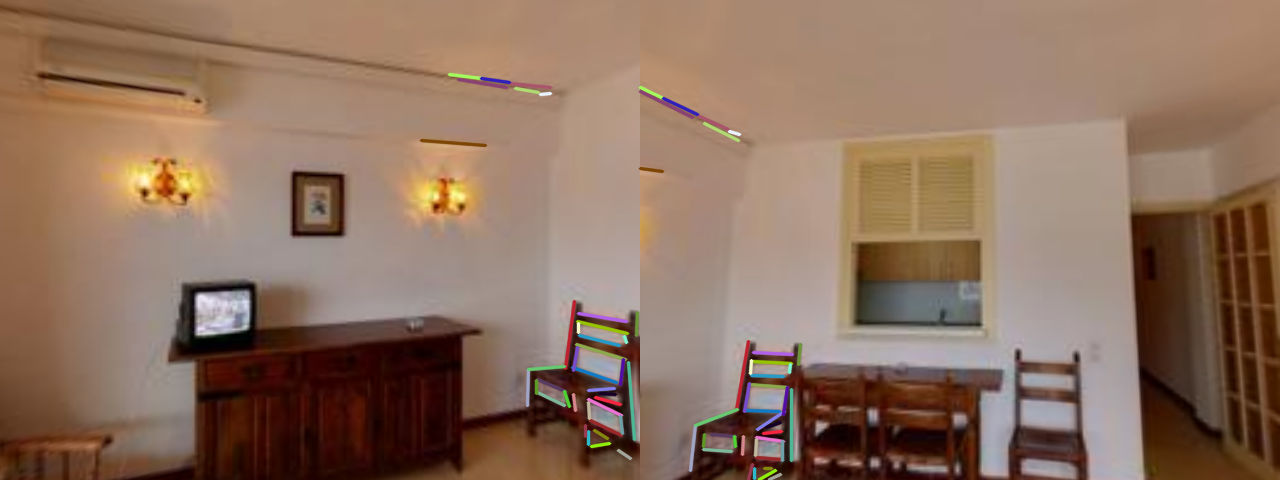}
            \includegraphics[width=\linewidth]{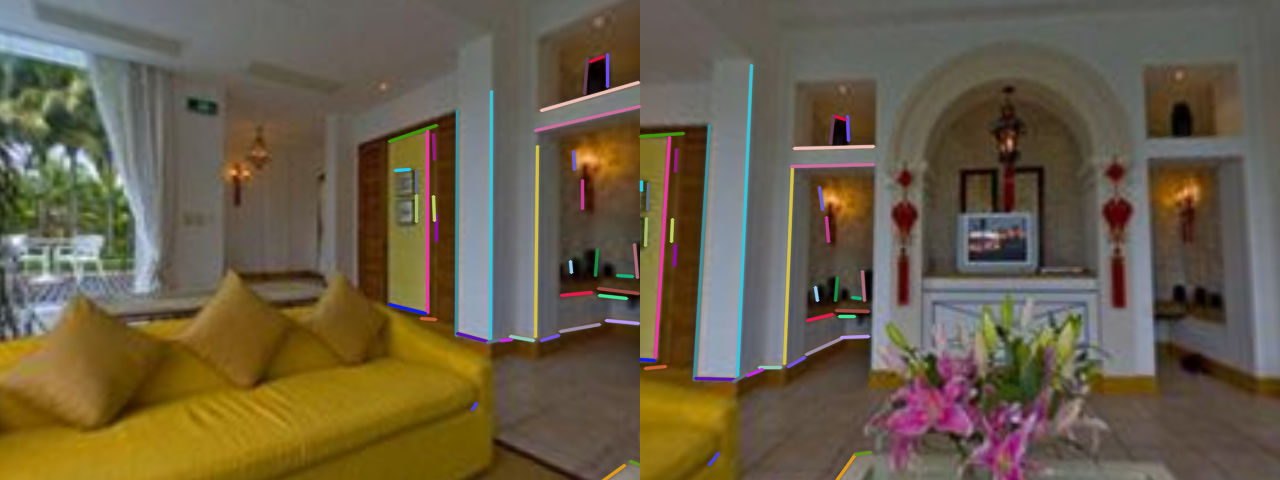}
            \includegraphics[width=\linewidth]{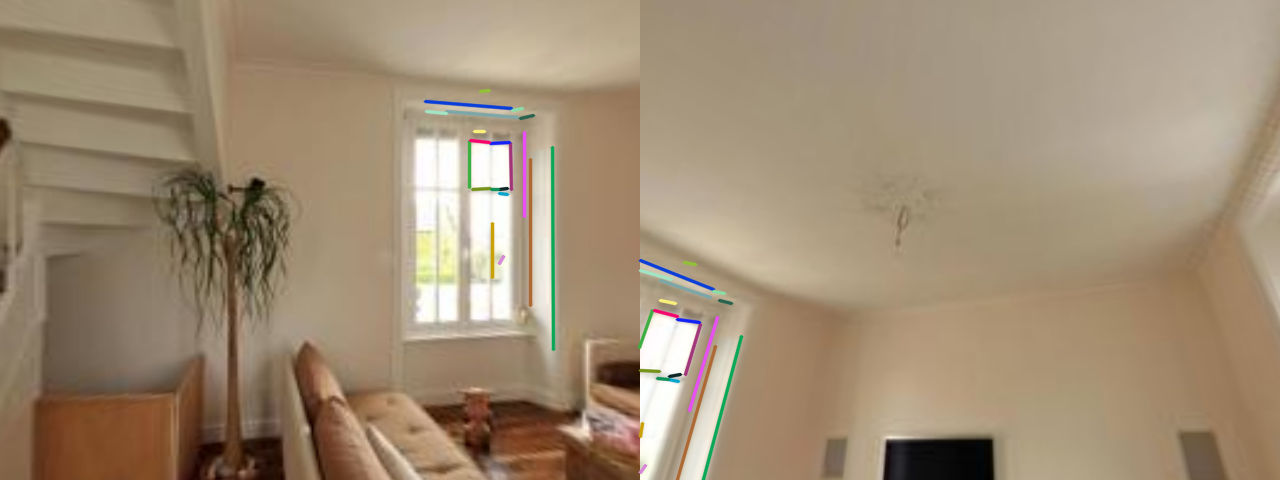}
            \caption{\small Line matches}
        \end{subfigure}    
        \hfill
        \begin{subfigure}{0.33\linewidth}
            \centering
            \includegraphics[width=\linewidth]{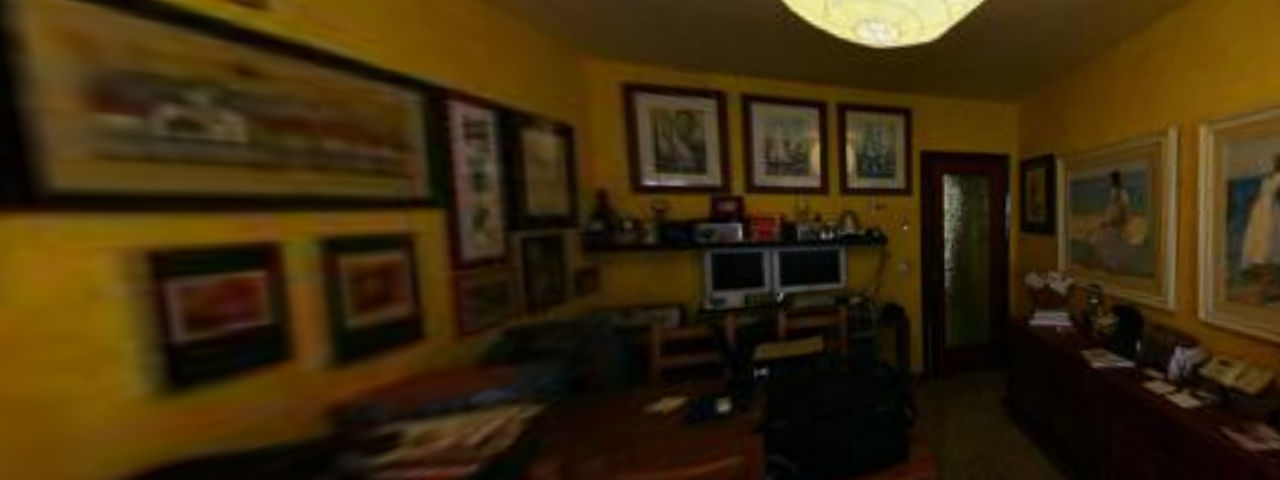}
            \includegraphics[width=\linewidth]{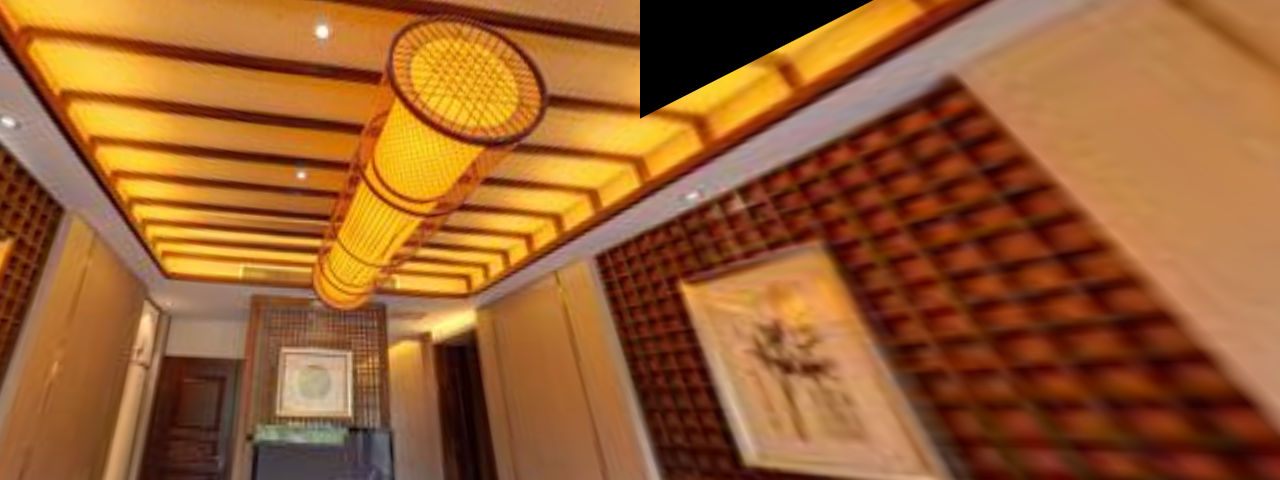}
            \includegraphics[width=\linewidth]{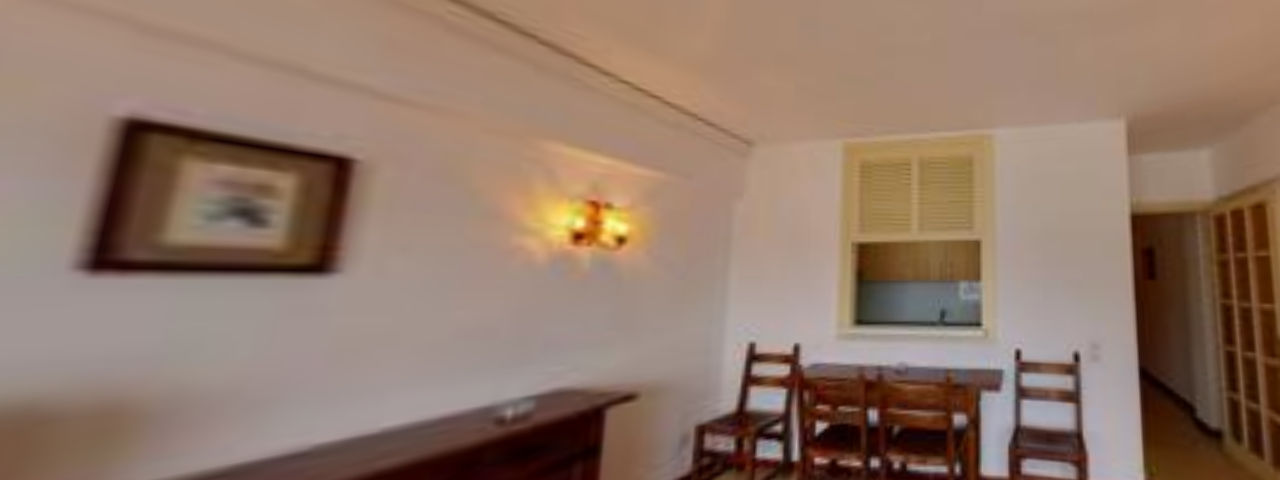}
            \includegraphics[width=\linewidth]{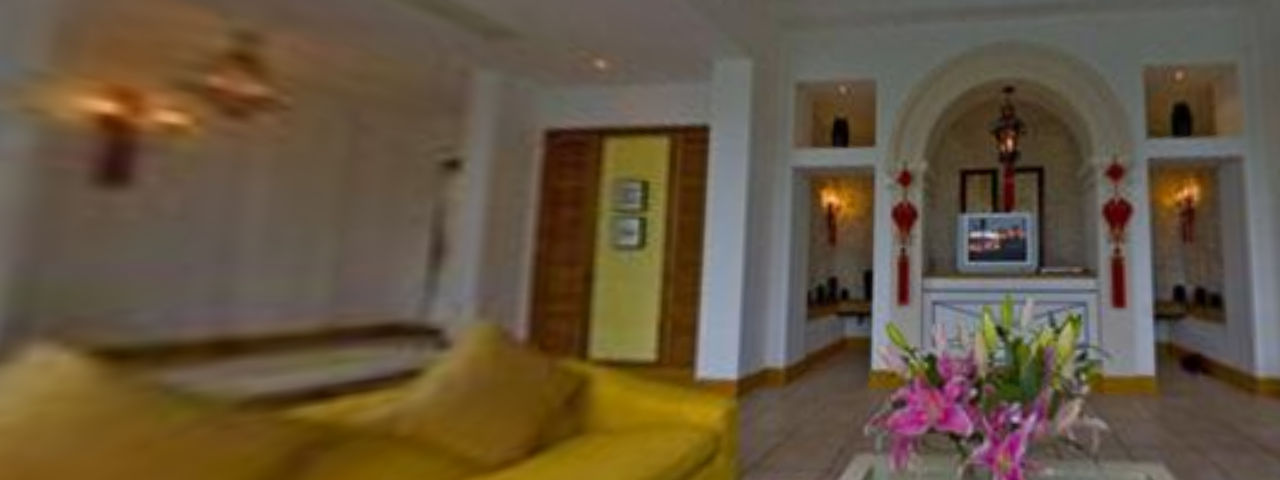}
            \includegraphics[width=\linewidth]{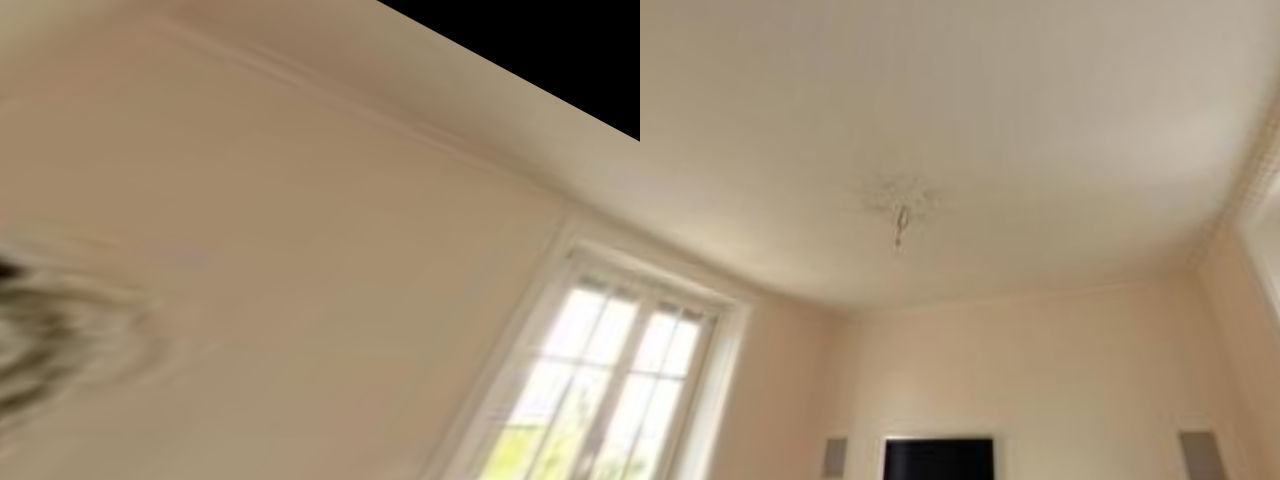}
            \caption{\small Image stitching results}
        \end{subfigure}        
\caption{\textbf{Examples of GlueStick matches on image pairs of SUN360~\cite{xiao2012recognizing}.} We provide the point and line matches, as well as the stitching of the two images using the resulting matches.}
\label{fig:pure-rotation-examples}
\end{figure*}

Images are cropped from the panorama images of SUN360~\cite{xiao2012recognizing} and projected with a calibration matrix $\mathbf{K} \in \mathbb{R}^{3\times3}$. \cref{fig:pure-rotation-examples} shows some examples. The relation between both images is defined by:
\begin{equation}
    \mathbf{x}^{B} = \mathbf{K} \mathbf{R} \mathbf{K}^{-1} \mathbf{x}^{A},
\end{equation}
where $\mathbf{R} \in SO(3)$ is the rotation matrix between the cameras. Thus, we can calibrate the features detected on each image, multiplying them by $\mathbf{K}^{-1}$. This way, we only have to robustly estimate the 3 Degrees-of-Freedom (DoF) of the rotation.

Point features are sampled uniformly, and lines are sampled proportionally to the square root of their length to give priority to larger lines. The probability of choosing one type of feature is proportional to its number of matches. For example, if a pair has 60 point matches and 40 line matches, the probability of choosing a point is 60\% and 40\% for lines.

The minimal solver randomly chooses 2 feature matches (point-point, point-line, or line-line) and estimates the rotation based on them. This can be seen as aligning 2 sets of 3D vectors. Homogeneous points are already 3D vectors going from the camera center to the plane $Z=1$. To get a vector from a line segment with homogeneous endpoints $\mathbf{x}_s$ and $\mathbf{x}_e$ we use its line-plane normal $\mathbf{n} = \mathbf{x}_s \times \mathbf{x}_e$. To make the method invariant to the order of the endpoints, we force the normals of the segments to have a positive dot product. This simple heuristic works as long as the sought rotation is less than 180$^\circ$.
Therefore, we can obtain a 3D vector from any of the types of features. For two (or more) correspondences, the optimal rotation can then be found with SVD using the Kabsch algorithm~\cite{kabsch_1976_solution}.

We provide a more extensive table of results than in the main paper in \cref{tab:rot_exp}, as well as visual examples of the point and line matches obtained by GlueStick, and the resulting image stitching output in \cref{fig:pure-rotation-examples}.

\begin{table}
\centering
\scriptsize
\setlength{\tabcolsep}{3pt}
\begin{tabular}{lcccccccc}
    \toprule
    & \multicolumn{2}{c}{Rotation error ($\downarrow$)} & \multicolumn{6}{c}{AUC ($\uparrow$)} \\
    \cmidrule(r{5pt}l{5pt}){2-3} 
    \cmidrule(r{5pt}l{5pt}){4-9} 
    & Avg   & Med & 0.25º  & 0.5º  & 1º & 2º  & 5º  & 10º  \\
    \midrule
    LineTR~\cite{syoon_2021_linetr} & 60.37 & 10.80 & 0.239 & 0.307 & 0.366 & 0.414 & 0.455 & 0.474 \\ 
    LBD~\cite{zhang_2013_lbd} & 19.57 & 0.054 & 0.593 & 0.681 & 0.736 & 0.768 & 0.790 & 0.799 \\ 
    SOLD${}^2$~\cite{pautrat_2021_sold2} & 23.74 & 0.308 & 0.232 & 0.384 & 0.515 & 0.609 & 0.682 & 0.713 \\ 
    L2D2~\cite{l2d2} & 18.31 & 0.056 & 0.578 & 0.667 & 0.725 & 0.765 & 0.795 & 0.808 \\ 
    GlueStick-L & 8.79 & 0.070 & 0.579 & 0.701 & 0.780 & 0.830 & 0.869 & 0.885 \\ 
    \midrule
    SG~\cite{sarlin_2020_superglue} & \textbf{0.135} & 0.052 & 0.730 & 0.860 & 0.929 & 0.964 & 0.985 & \textbf{0.992} \\ 
    GlueStick-P & 0.230 & 0.052 & 0.729 & 0.860 & 0.928 & 0.963 & 0.984 & 0.991 \\ 
    \midrule
    PL-Loc~\cite{syoon_2021_linetr} & 0.338 & 0.050 & 0.733 & 0.859 & 0.927 & 0.961 & 0.982 & 0.989 \\ 
    GlueStick-PL & 0.327 & \textbf{0.039} & \textbf{0.789} & \textbf{0.890} & \textbf{0.943} & \textbf{0.970} & \textbf{0.986} & 0.991 \\ 
    \bottomrule
\end{tabular}
\caption{\textbf{Pure rotation estimation on SUN360~\cite{xiao2012recognizing}.} We estimate a rotation based on point-only, line-only, or points+lines matches. We report the average and median rotation error in degrees, as well as the Area Under the Curve (AUC) at 0.25 / 0.5 / 1 / 2 / 5 / 10 degrees error.}
\label{tab:rot_exp}
\end{table}

\subsection{Homography Estimation in HPatches}
HPatches~\cite{hpatches_2017_cvpr} is one of the most frequently used datasets to evaluate image matching. It contains 108 sequences where the scene contains only one dominant plane, with 6 images per sequence. Each sequence has either illumination or viewpoint changes.
Similarly as in ~\cite{sarlin_2020_superglue}, we compute a homography from point and/or line correspondences and RANSAC, and compute the Area Under the Curve (AUC) of the reprojection error of the four image corners. We report the results for thresholds 3 / 5 / 10 pixels. We also compute the precision and recall of the ground truth (GT) matches obtained by the GT homography.

\begin{figure}
	\centering
	\includegraphics[width=1.0\columnwidth]{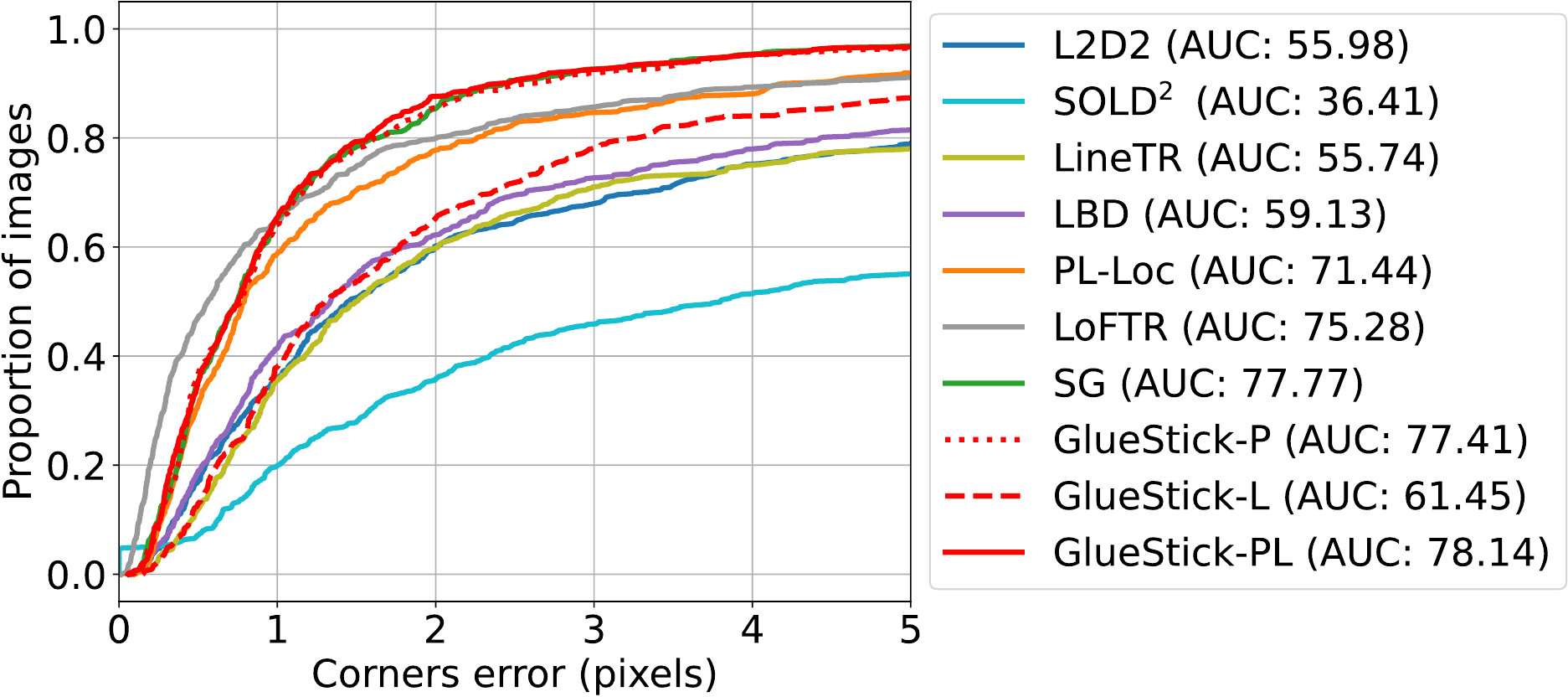}
	\caption{\textbf{HPatches~\cite{hpatches_2017_cvpr} cumulative error curve.} We report the percentage of images where the homography is correctly predicted for various pixel error thresholds.}
	\label{fig:hpatches_exp}
\end{figure}

We report the results in \cref{tab:hpatches_exp} and \cref{fig:hpatches_exp}. HPatches is clearly saturated and the precision/recall metrics are already very high for point-based methods. Note the strong performance of GlueStick on line matching, with an increase by nearly 10\% in precision compared to the previous state of the art. Regarding point-based methods and homography scores, GlueStick obtains a very similar performance as the previous point matchers SuperGlue~\cite{sarlin_2020_superglue} and LoFTR~\cite{sun_2021_loftr}, and ranks first (with a very small margin) in terms of homography estimation.
In addition to the fact that HPatches is saturated, it contains also very few structural lines that could have been useful to refine the homography fitting. Thus, the improvement brought by line segments is not significant here.

\begin{table}
	\centering
	\scriptsize
	\setlength{\tabcolsep}{3pt}

 	\resizebox{\linewidth}{!}{%

        \begin{tabular}{clccccccc}
		\toprule
		& & \multicolumn{3}{c}{AUC ($\uparrow$)} & \multicolumn{2}{c}{ Points ($\uparrow$)} & \multicolumn{2}{c}{ Lines ($\uparrow$)} \\
		\cmidrule(r{5pt}l{5pt}){3-5} 
		\cmidrule(r{5pt}l{5pt}){6-7} 
		\cmidrule(r{5pt}l{5pt}){8-9} 
		& & 3px  & 5px  & 10px & P & R & P & R \\
		\midrule
		\multirow{5}{*}{L}
		& L2D2~\cite{l2d2} & 43.73 & 55.98 & 69.39 & - & - & 55.55 & 38.76 \\ 
		& SOLD${}^2$~\cite{pautrat_2021_sold2} & 26.60 & 36.41 & 48.82 & - & - & 80.57 & 77.43 \\ 
        & LineTR~\cite{syoon_2021_linetr} & 42.90 & 55.74 & 69.26 & - & - & 78.78 & 58.74 \\ 
		& LBD~\cite{zhang_2013_lbd} & 46.82 & 59.13 & 71.82 & - & - & 82.73 & 56.38 \\ 
		& GlueStick-L & 46.61 & 61.45 & 76.32 & - & - & \textbf{90.27} & 76.69 \\ 
		\midrule
		\multirow{3}{*}{P} 
        & SuperGlue~\cite{sarlin_2020_superglue} & 66.21 & 77.77 & 88.05 & \textbf{98.85} & 97.44 & - & - \\ 
		& LoFTR~\cite{sun_2021_loftr} & 66.15 & 75.28 & 84.54 & 97.60 & \textbf{99.38} & - & - \\
		& GlueStick-P & 65.88 & 77.41 & 87.72 & \textbf{98.85} & 97.08 & - & - \\ 
		\midrule
		\multirow{2}{*}{P+L} 
		& PL-Loc~\cite{syoon_2021_linetr} & 60.03 & 71.44 & 83.08 & 90.80 & 77.60 & 80.33 & 50.35 \\ 
		& GlueStick-PL & \textbf{66.88} & \textbf{78.14} & \textbf{88.12} & 98.00 & 94.86 & 89.54 & \textbf{80.44} \\ 
		\bottomrule
	\end{tabular}
        }
	\caption{\textbf{Homography estimation in HPatches~\cite{hpatches_2017_cvpr}.} We report the Area Under the Curve (AUC) of the cumulative error curve generated by the re-projection error of the four image corners at different thresholds (3px, 5px, 10px), as well as the precision (P) and recall (R) of the matches.}
	\label{tab:hpatches_exp}
\end{table}

\subsection{Visual Localization on the Full 7Scenes Dataset}
As stated in the main paper, 7Scenes~\cite{7scenes} is a rather small-scale dataset for visual localization, which is already largely saturated for point-based methods. Adding line segments into the pipeline can improve the results only in a few scenes such as Fire, Office and mostly on Stairs. We demonstrate this in \cref{tab:7scenes_full}, where we used the same setup as described in the main paper. While all methods obtain close results on such a saturated dataset, GlueStick is slightly ahead of the baselines, and largely outperforms them on the most challenging scene, Stairs.

\begin{table*}
    \centering
    \scriptsize
    \setlength{\tabcolsep}{6pt}
    \begin{tabular}{ccccccccc}
        \toprule
         & \multicolumn{3}{c}{Points} & \multicolumn{5}{c}{Points + Lines} \\
        \cmidrule(r{5pt}l{5pt}){2-4} \cmidrule(r{5pt}l{5pt}){5-9}
         & SuperGlue~\cite{sarlin_2020_superglue} & LoFTR~\cite{sun_2021_loftr} & GlueStick - P & SOLD${}^{2}$~\cite{pautrat_2021_sold2} & LineTR~\cite{syoon_2021_linetr} & L2D2~\cite{l2d2} & SG + Endpts & GlueStick - PL \\
        \midrule
        Chess & \textbf{2.4} / 0.81 / 94.5 & 2.5 / 0.86 / 93.8 & \textbf{2.4} / \textbf{0.80} / 94.3 & \textbf{2.4} / 0.82 / 94.4 & \textbf{2.4} / 0.81 / 94.5 & \textbf{2.4} / 0.83 / 94.5 & \textbf{2.4} / 0.82 / \textbf{94.6} & \textbf{2.4} / 0.82 / 94.5 \\
        Fire & 1.9 / 0.76 / 96.4 & 1.7 / \textbf{0.66} / 96.8 & 2.0 / 0.78 / 96.6 & \textbf{1.6} / 0.69 / 96.8 & \textbf{1.6} / 0.69 / 97.0 & \textbf{1.6} / 0.69 / 96.4 & 1.7 / 0.69 / 97.2 & 1.7 / 0.69 / \textbf{97.4} \\
        Heads & 1.1 / 0.74 / 99.0 & 1.1 / 0.78 / 98.2 & 1.1 / 0.74 / 99.2 & \textbf{1.0} / \textbf{0.72} / \textbf{99.4} & 1.1 / 0.75 / 99.1 & \textbf{1.0} / 0.73 / 99.3 & 1.1 / 0.74 / 99.2 & \textbf{1.0} / 0.74 / \textbf{99.4} \\
        Office & 2.7 / 0.83 / 83.9 & 2.7 / 0.83 / 82.0 & 2.7 / 0.83 / 83.6 & \textbf{2.6} / 0.80 / \textbf{84.7} & \textbf{2.6} / \textbf{0.79} / 84.4 & \textbf{2.6} / 0.81 / 83.9 & \textbf{2.6} / \textbf{0.79} / 84.4 & \textbf{2.6} / \textbf{0.79} / 84.6 \\
        Pumpkin & 4.0 / 1.05 / 62.0 & \textbf{3.9} / 1.12 / \textbf{62.4} & \textbf{3.9} / \textbf{1.04} / 62.2 & 4.0 / 1.07 / 60.2 & 4.0 / 1.08 / 61.5 & 4.0 / 1.05 / 61.3 & 4.0 / 1.06 / 61.2 & 4.0 / 1.06 / 61.5 \\
        Red kitchen & 3.3 / \textbf{1.12} / 72.5 & 3.3 / 1.14 / \textbf{73.8} & 3.3 / \textbf{1.12} / 72.8 & \textbf{3.2} / 1.15 / 72.6 & 3.3 / 1.15 / 72.6 & \textbf{3.2} / 1.14 / 72.9 & \textbf{3.2} / 1.14 / 72.8 & \textbf{3.2} / 1.13 / 73.0 \\
        Stairs &  4.7 / 1.25 / 53.4 & 4.4 / 0.95 / 53.9 & 4.4 / 1.21 / 55.4 & 3.2 / 0.83 / 75.8 & 3.7 / 1.02 / 66.6 & 4.1 / 1.15 / 55.8 & 3.1 / 0.81 / 75.6 & \textbf{2.9} / \textbf{0.79} / \textbf{79.7} \\
        \midrule
        Total & 2.9 / 0.94 / 80.2 & 2.8 / 0.91 / 80.1 & 2.8 / 0.93 / 80.6 & 2.6 / 0.87 / 83.4 & 2.7 / 0.90 / 82.2 & 2.7 / 0.91 / 80.6 & 2.6 / \textbf{0.86} / 83.6 & \textbf{2.5 / 0.86 / 84.3} \\
        \bottomrule
    \end{tabular}
    \caption{\textbf{Visual localization on the full 7Scenes dataset~\cite{7scenes}.} We report the median translation error (cm) / median rotation error (deg) / pose AUC at a 5 cm / 5 deg threshold. Most scenes are already saturated for point methods, and lines can hardly make a difference.}
    \label{tab:7scenes_full}
\end{table*}

\section{Qualitative Examples}
\label{sec:qualitative_examples}

\subsection{Feature Matches}
\begin{figure*}
    \centering
    \setlength{\tabcolsep}{4pt}
    \begin{tabular}{ccc}
        \rotatebox{90}{\phantom{xxxx}LBD~\cite{zhang_2013_lbd}} &
        \includegraphics[width=0.47\textwidth]{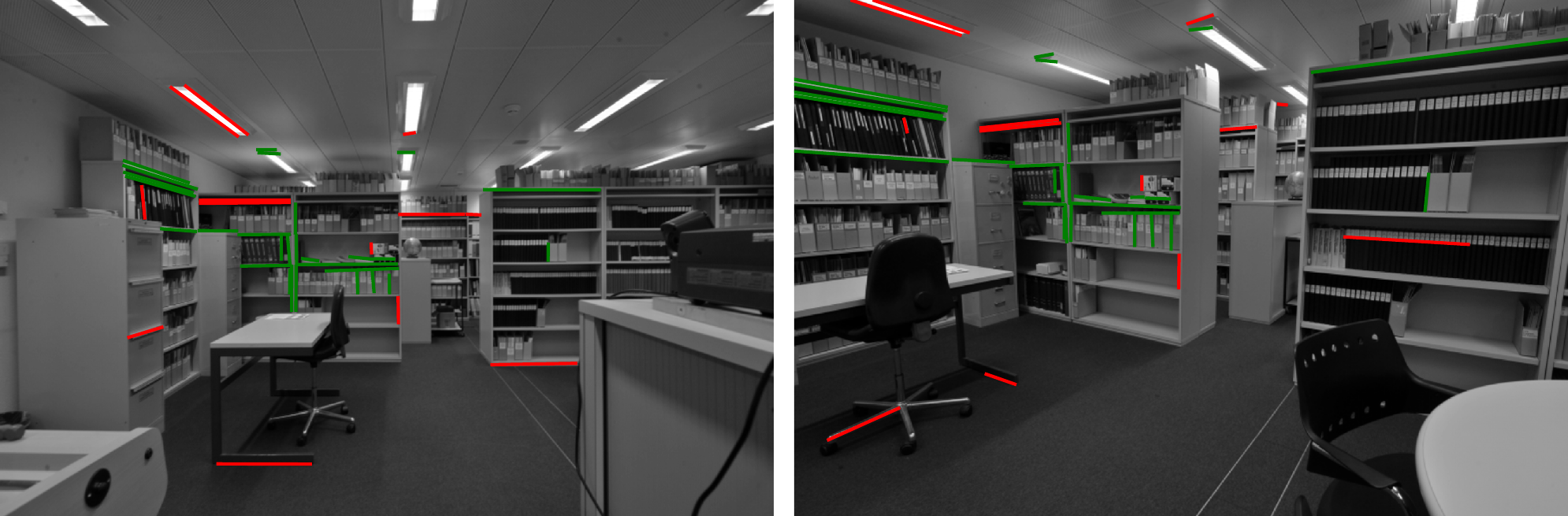} &
        \includegraphics[width=0.47\textwidth]{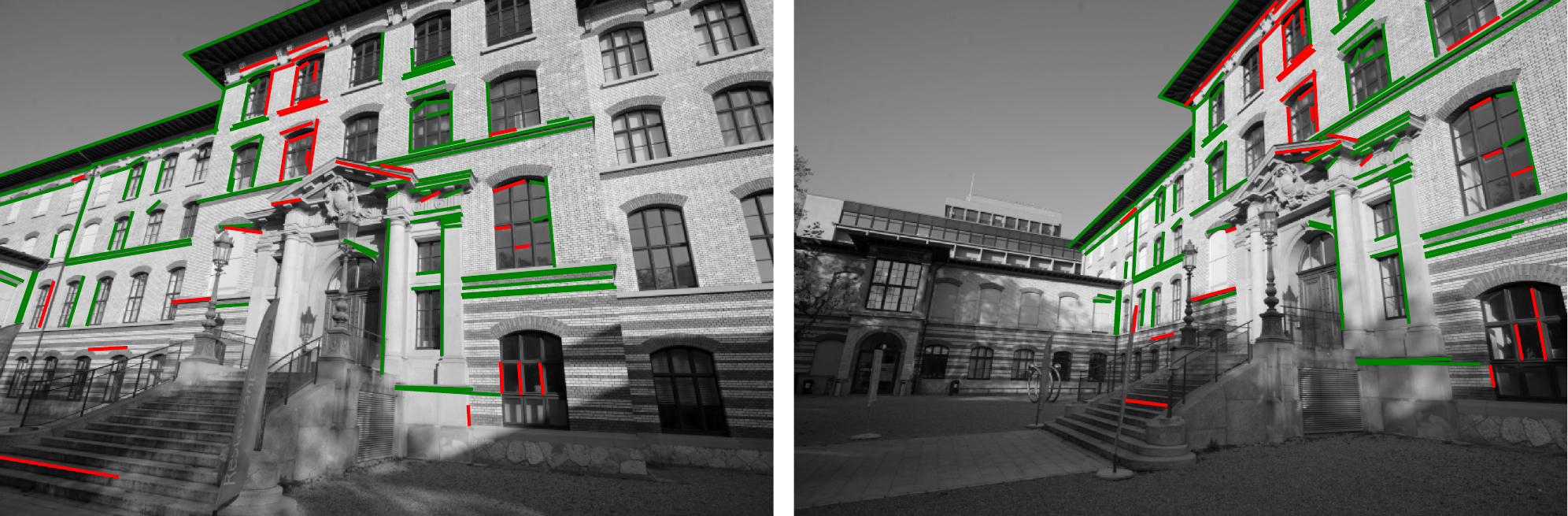} \\
        \rotatebox{90}{\phantom{xxx}SOLD${}^{2}$\cite{pautrat_2021_sold2}} &
        \includegraphics[width=0.47\textwidth]{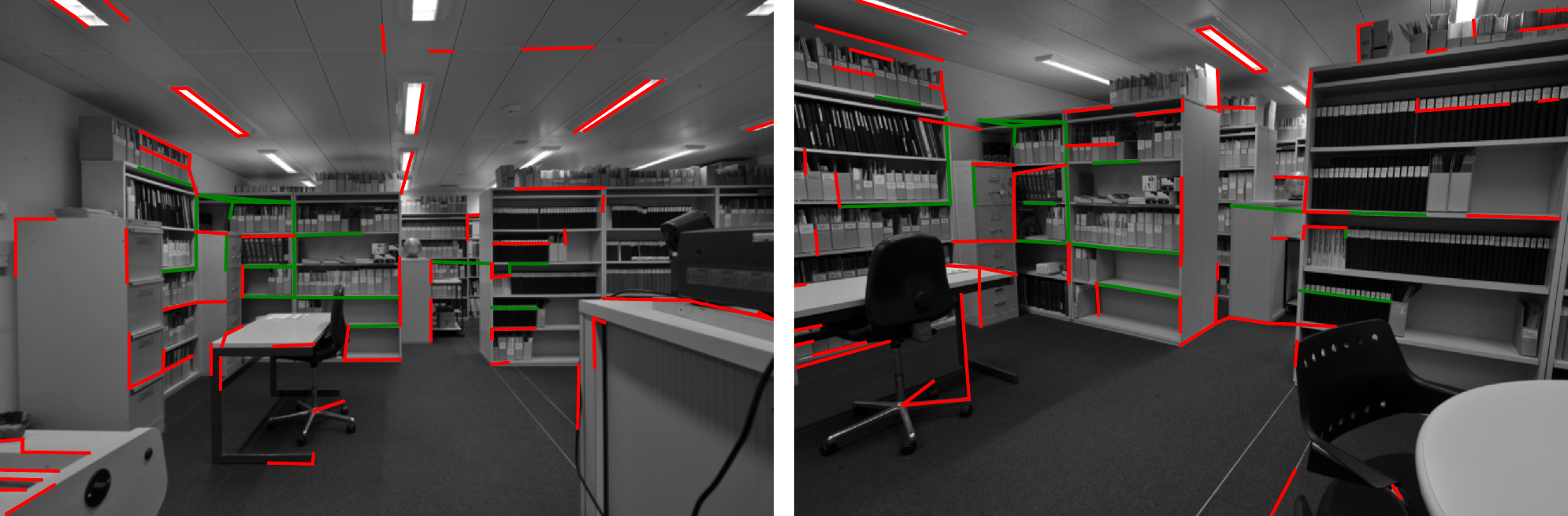} &
        \includegraphics[width=0.47\textwidth]{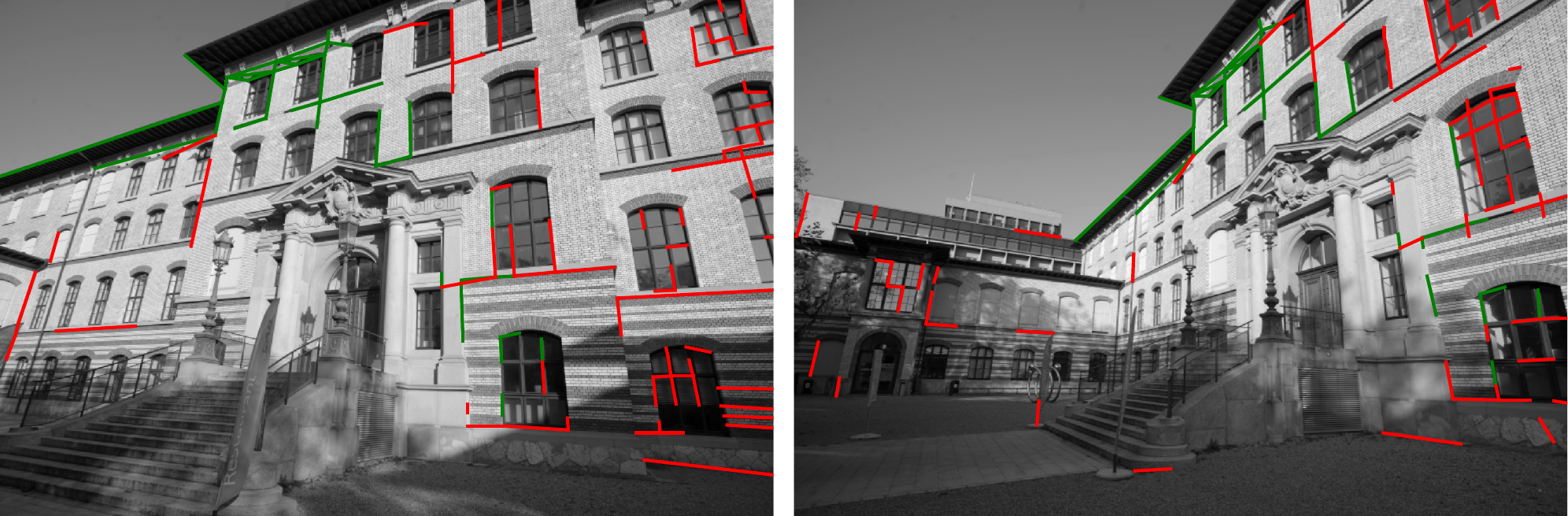} \\
        \rotatebox{90}{\phantom{xxx}LineTR~\cite{syoon_2021_linetr}} &
        \includegraphics[width=0.47\textwidth]{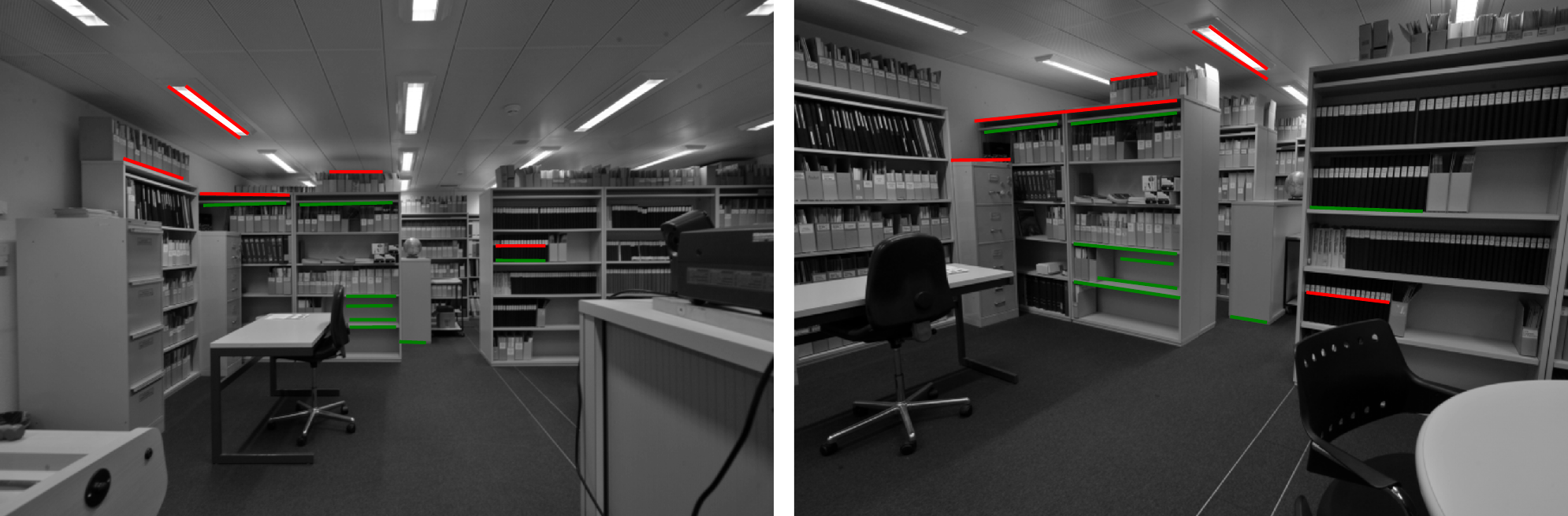} &
        \includegraphics[width=0.47\textwidth]{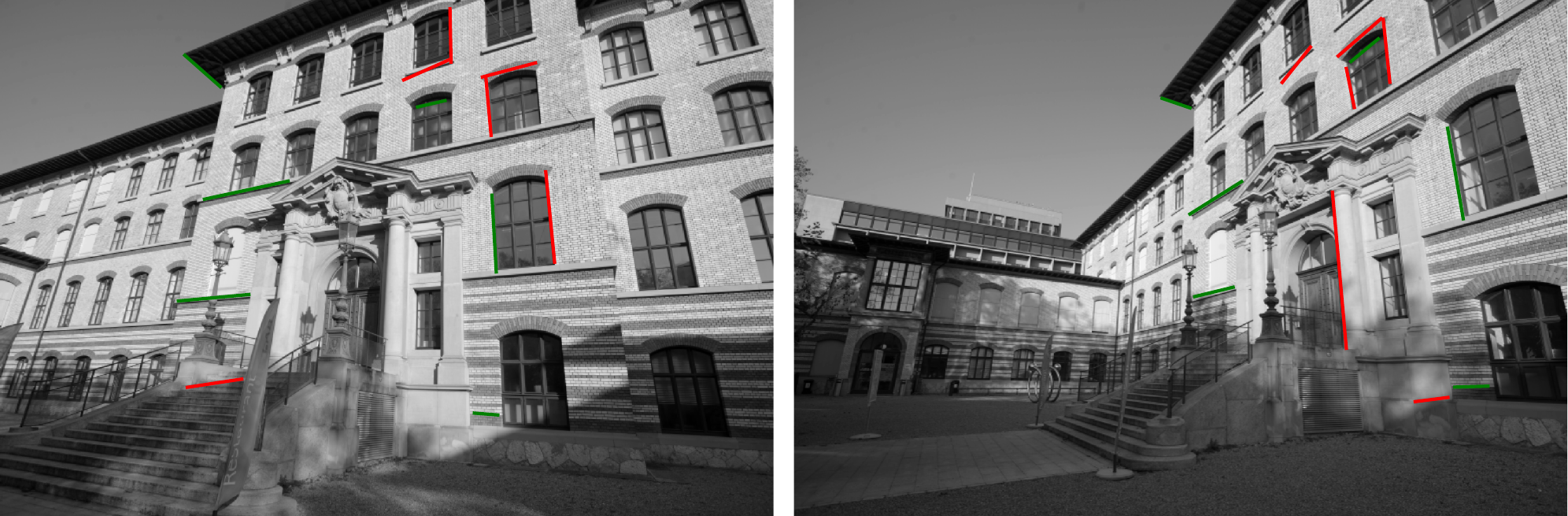} \\
        \rotatebox{90}{\phantom{xxxx}L2D2~\cite{l2d2}} &
        \includegraphics[width=0.47\textwidth]{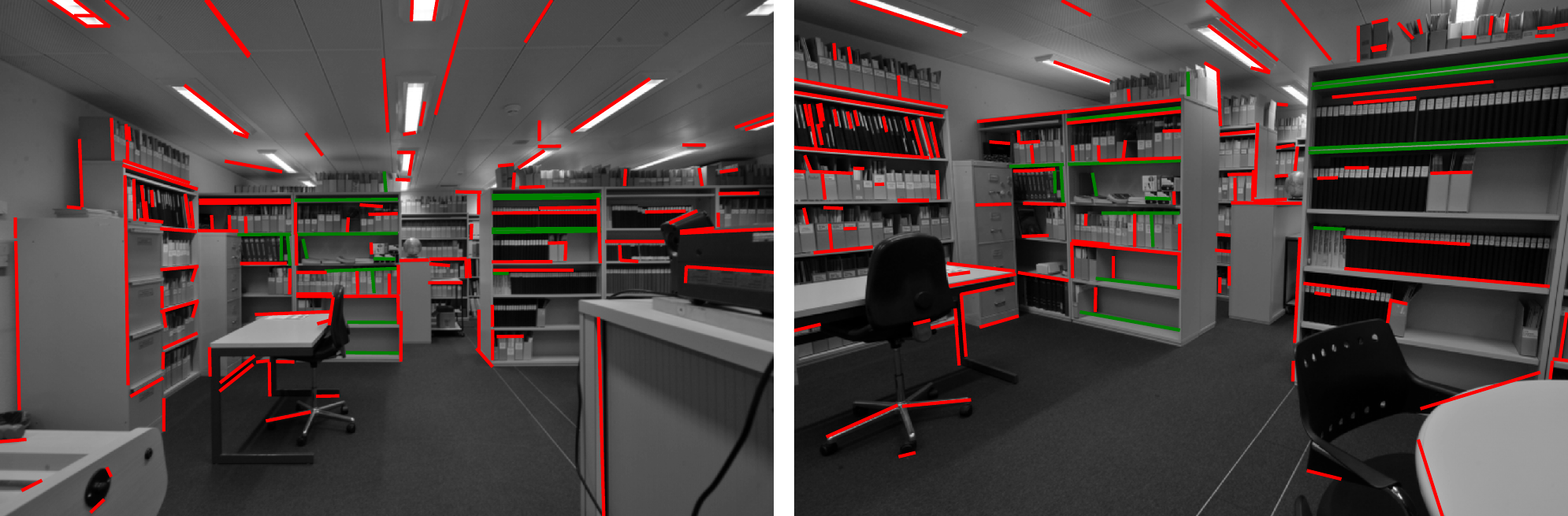} &
        \includegraphics[width=0.47\textwidth]{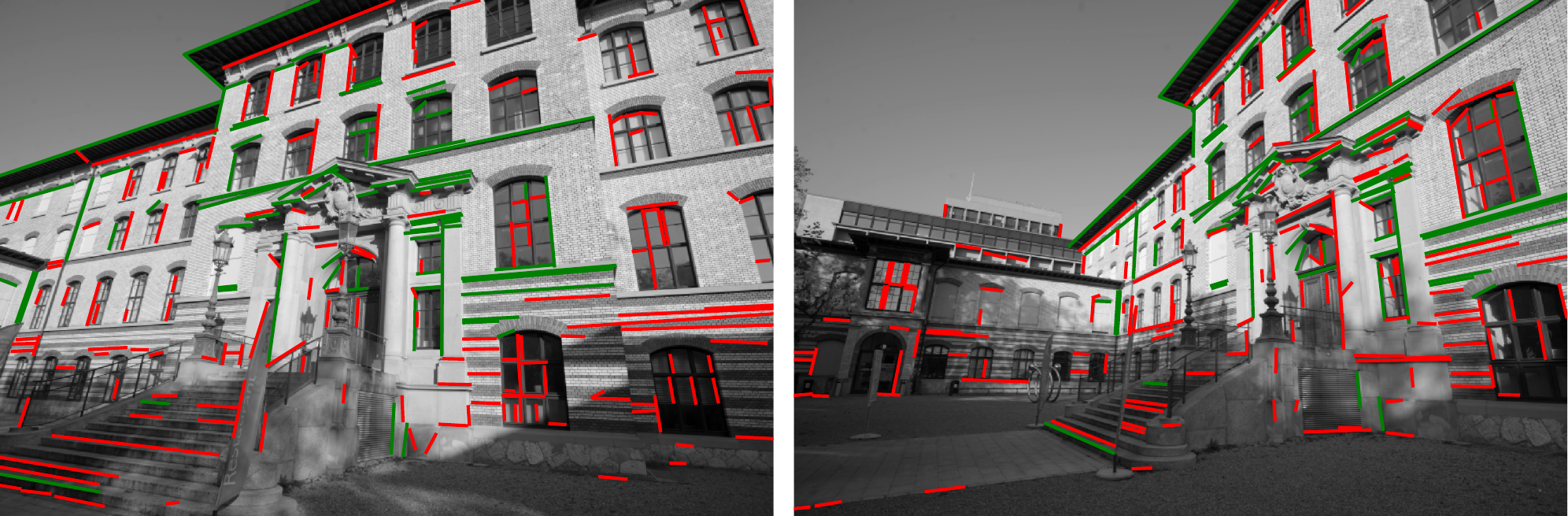} \\
        \rotatebox{90}{\phantom{xxx}GlueStick} &
        \includegraphics[width=0.47\textwidth]{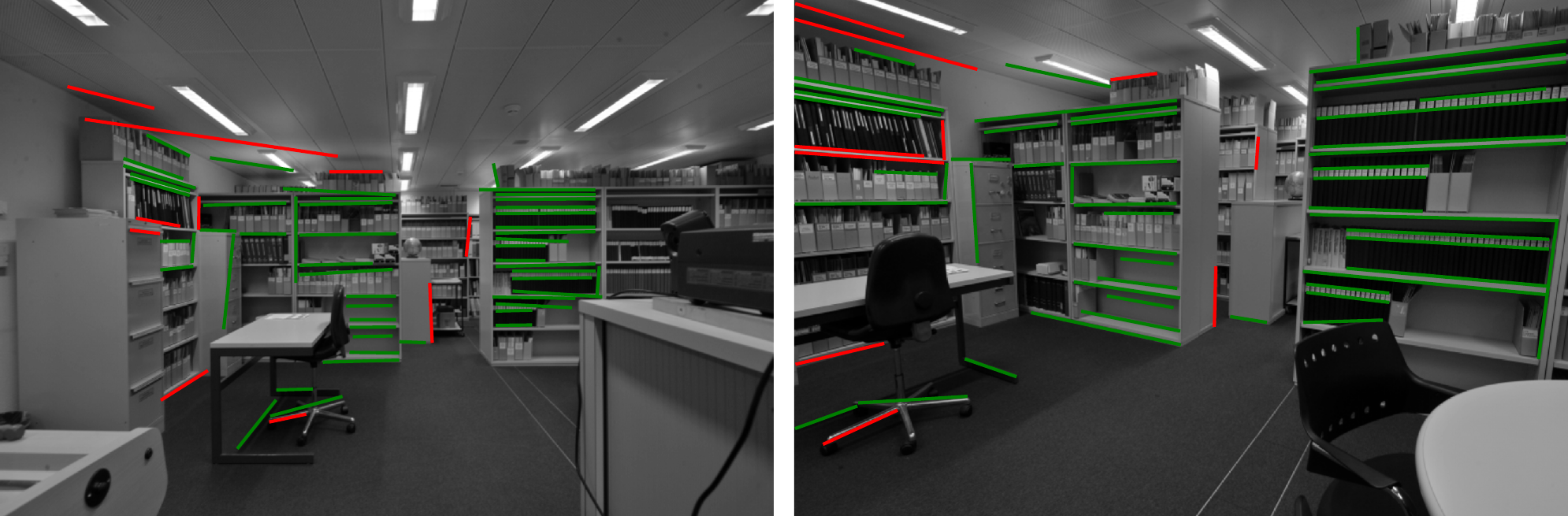} &
        \includegraphics[width=0.47\textwidth]{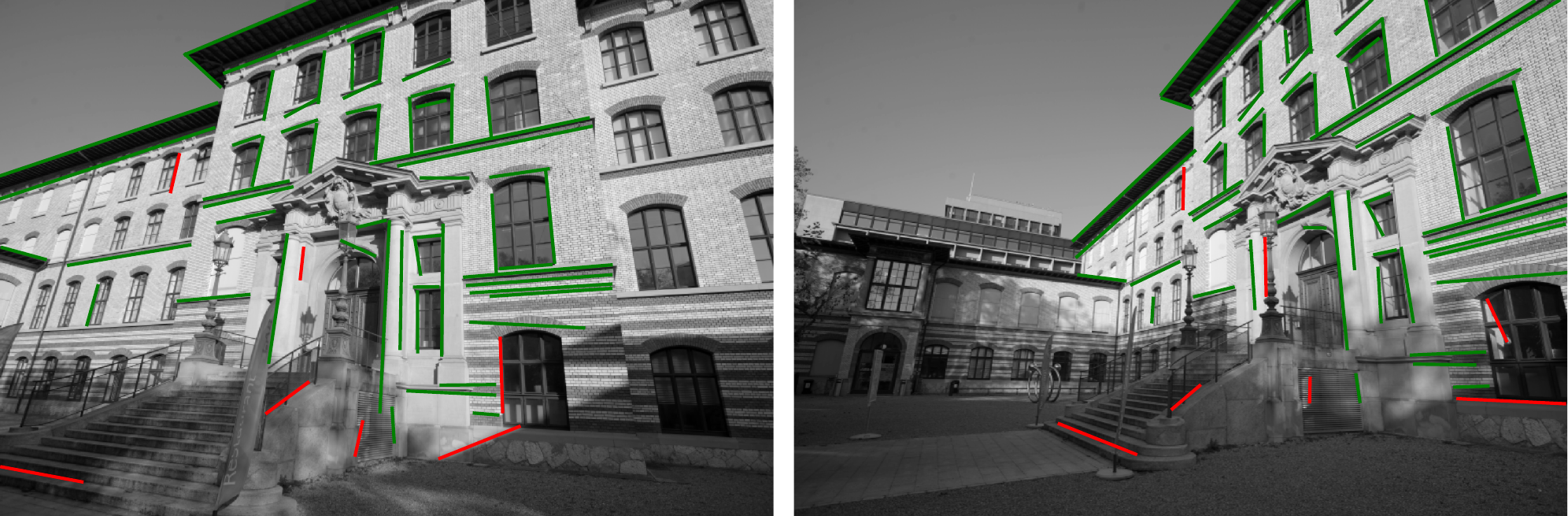} \\
    \end{tabular}
    \caption{\textbf{Line matches examples on ETH3D~\cite{Schops_2017_eth3d}.} We display correct line matches in green and incorrect ones in red for several state-of-the-art line matchers.}
    \label{fig:examples}
\end{figure*}

\cref{fig:examples} displays some examples of line matching on the ETH3D dataset~\cite{Schops_2017_eth3d}. We plot in green the correct matches and in red the incorrect ones. Thanks to its spatial reasoning in the GNN and context-awareness, GlueStick is consistently matching more lines and with a higher precision than previous works. This is in particular true for scenes with repeated structures, such as the one in the right column, where the descriptors of SOLD${}^{2}$\cite{pautrat_2021_sold2} and L2D2\cite{l2d2} do not have context from neighboring lines, and can only match a few lines.

\subsection{Visualization of the Camera Pose Estimation}
\begin{figure*}
    \centering
    \small
    \begin{tabular}{cc}
        hloc~\cite{sarlin2019coarse,hloc} & GlueStick \\
        \includegraphics[width=0.47\textwidth]{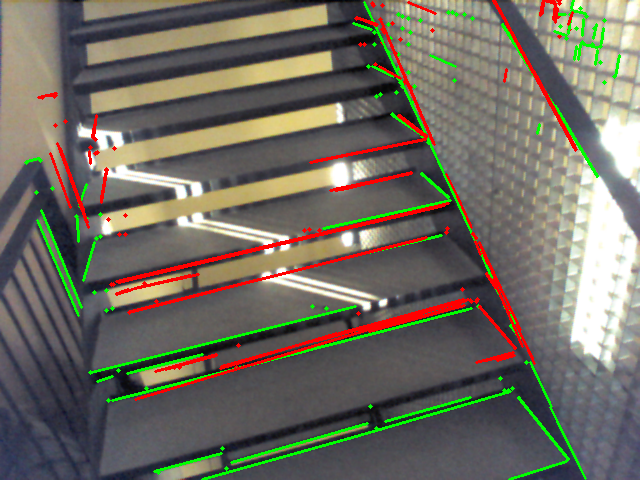} &
        \includegraphics[width=0.47\textwidth]{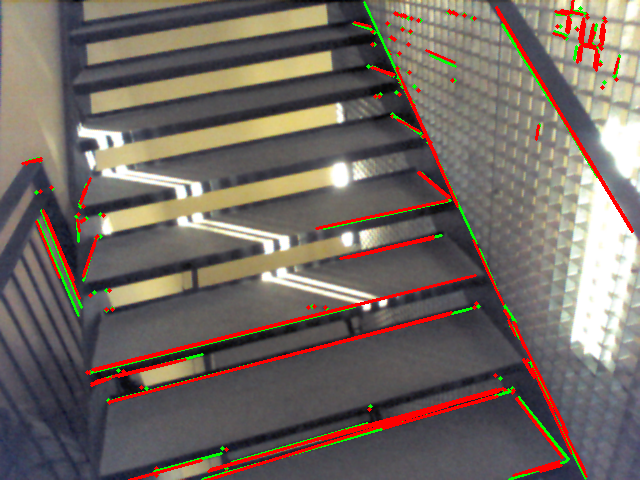} \\
        \includegraphics[width=0.47\textwidth]{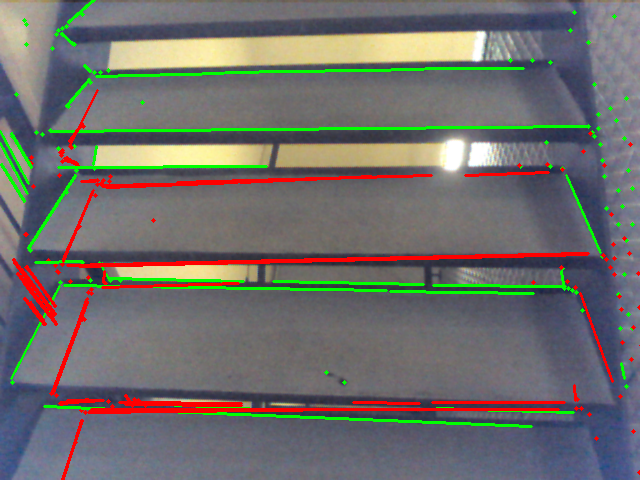} &
        \includegraphics[width=0.47\textwidth]{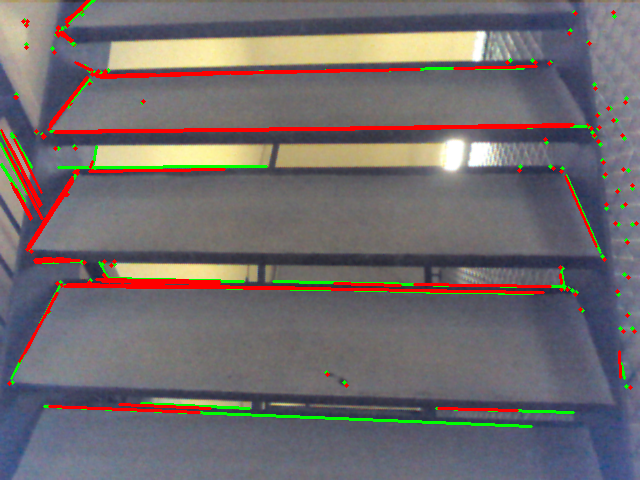}
    \end{tabular}
    \caption{\textbf{Camera pose estimation visualizations.} We compare the originally detected keypoints and lines (in green) with the re-projected points and lines using the estimated camera pose (in red), for hloc~\cite{sarlin2019coarse,hloc} with SuperPoint~\cite{detone_2018_superpoint} + SuperGlue~\cite{sarlin_2020_superglue} and our method. The reprojections of GlueStick align almost perfectly with the 2D detections, showing a high quality estimated pose.}
    \label{fig:line_loc_vis}
\end{figure*}

We visualize the reprojection of points and lines on the scene Stairs of the 7Scenes dataset~\cite{7scenes} in \cref{fig:line_loc_vis}. We plot in green the points and lines that were originally detected in 2D, and re-project in red the corresponding 3D features using the estimated camera pose. The reprojections of GlueStick are almost perfectly aligned compared to the ones of hloc~\cite{sarlin2019coarse,hloc}, highlighting the quality of the poses retrieved by our method.

\subsection{Failure Cases and Limitations}
While jointly matching keypoints and lines in the same matching network helps disambiguating many challenging scenarios, GlueStick may still underperform in some scenarios. We list in the following some limitations of our method, and report some failure cases.

\noindent \textbf{Limitations.}
Currently, the main performance bottleneck of GlueStick lies in the line segment detection. While this field has seen great advances in recent years, existing line detectors are still not as repeatable and accurate as point features, making the line matching more challenging. Partially occluded lines are also a potential issue for GlueStick, as it represents lines with their two endpoints. However, we observed a surprisingly good robustness of GlueStick to partially occluded lines, probably thanks to the neighboring points and lines that are not occluded. Note that it is also possible to equip GlueStick with a similar mechanism as in SOLD${}^{2}$~\cite{pautrat_2021_sold2}, by sampling several points along the line segments, and matching them with the Needleman-Wunsch algorithm. We tried this option and observed a small increase in performance (notably in areas with occluded lines), but at the cost of higher running time. Therefore, we did not incorporate this feature in our final method.

Another issue is that points and lines are still detected with different methods for now. Thus, three networks / algorithms need to be run to detect and describe keypoints, detect lines, and finally match them. Jointly detecting and describing points and lines would be an interesting future direction of research. Furthermore, the extraction of discrete features such as points and lines is usually non-differentiable, such that one cannot get a fully differentiable pipeline going from the feature extraction to their matching. Enabling such end-to-end training could potentially make features better specialized for matching.

Finally, our current supervision requires ground truth correspondences of points and lines across images (usually obtained through reprojection with depth and camera poses). Other supervision signals such as epipolar constraints and using two-view geometry would be an interesting direction of improvement in the future.

\begin{figure*}
    \centering
    \small
    \begin{tabular}{c}
        \includegraphics[width=0.8\textwidth]{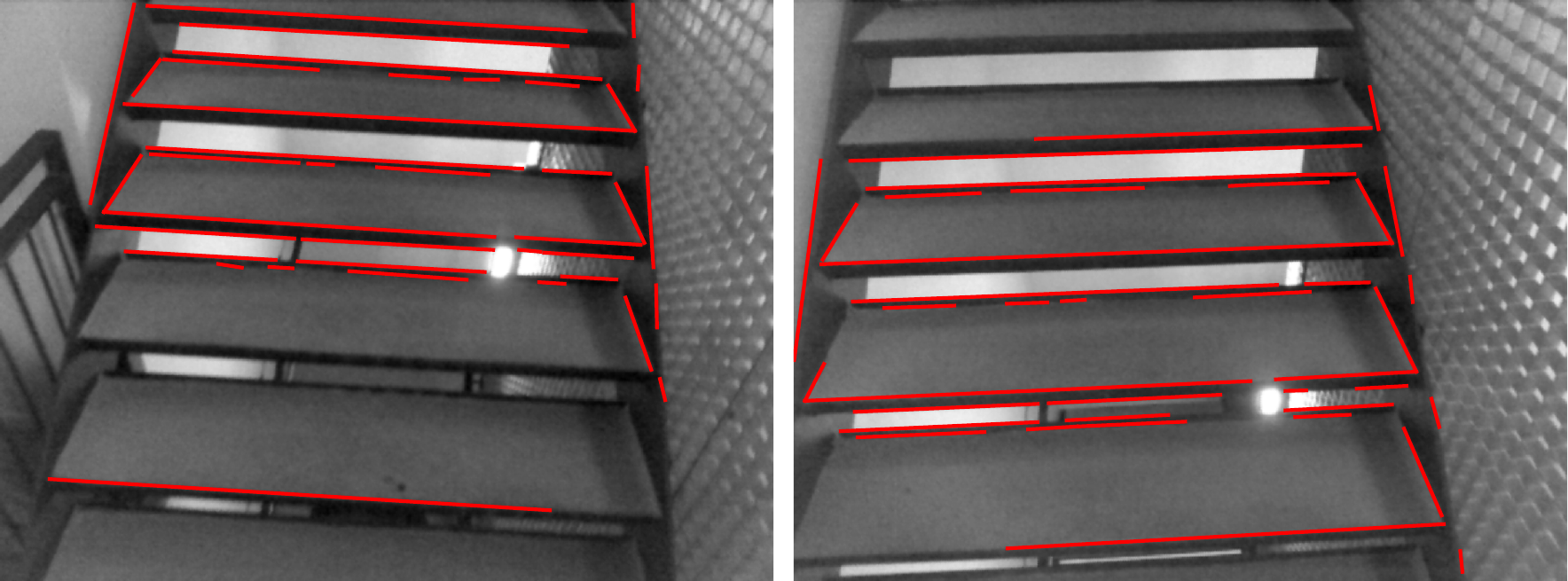} \\
        (a) Repeated patterns: GlueStick finds consistent line matches, but displaced by one step. \vspace{0.3cm} \\
        \includegraphics[width=0.8\textwidth]{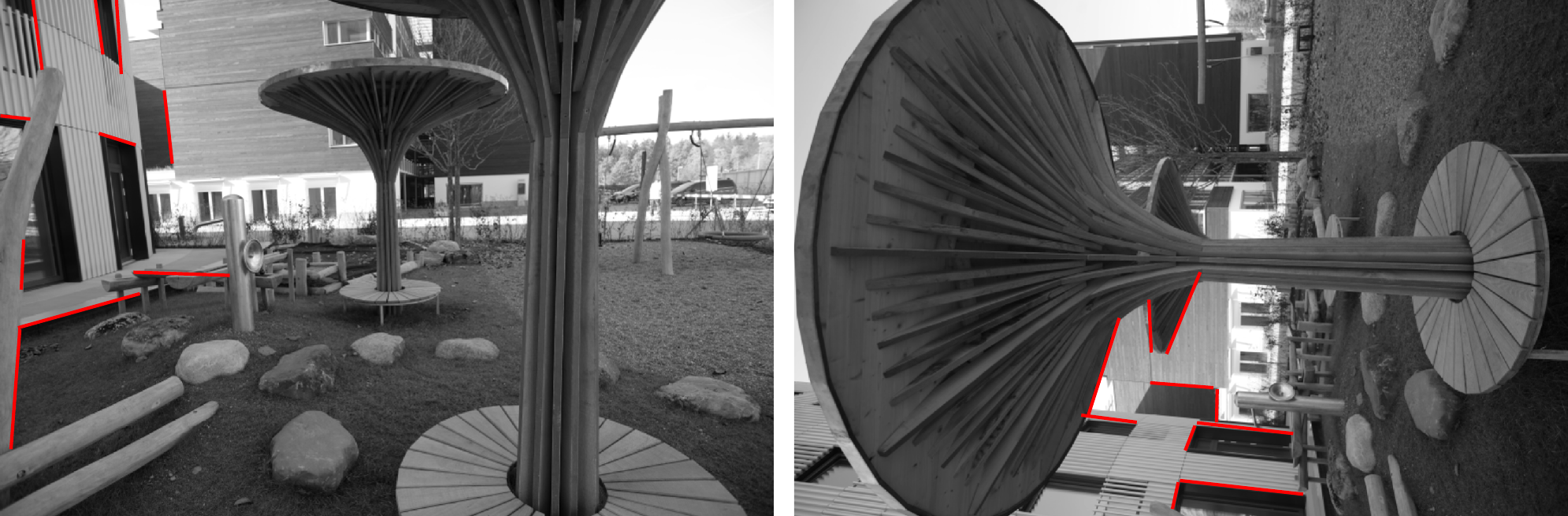} \\
        (b) GlueStick is not trained on large rotations. \vspace{0.3cm} \\
        \includegraphics[width=0.8\textwidth]{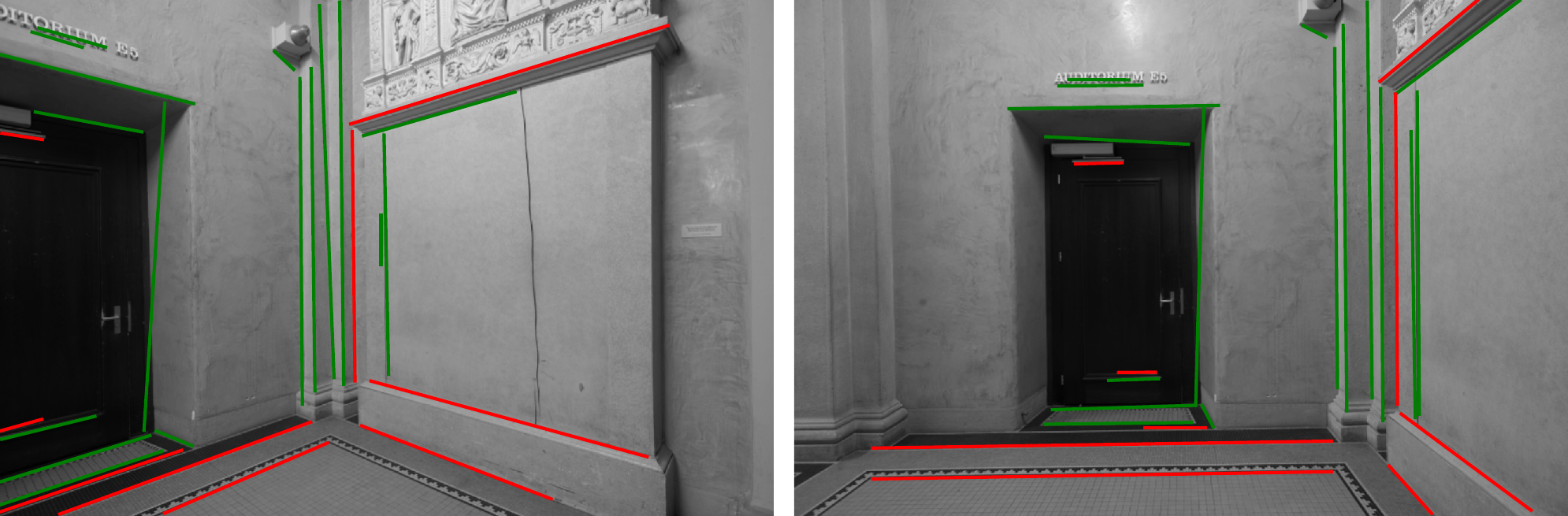} \\
        (c) The lack of texture / symmetric structures make visual descriptors / spatial descriptors less reliable. \\
    \end{tabular}
    \caption{\textbf{Failure cases.} We display correct line matches in green and wrong ones in red. GlueStick may still fail or underperform in some situations, such as (a) perfectly repeated patterns that are hard to disambiguate, (b) large rotation (e.g. $> 45^{\circ}$), and (c) lack of texture and symmetric structures.}
    \label{fig:failure_cases}
\end{figure*}

\noindent \textbf{Failure cases.}
We display in \cref{fig:failure_cases} a few examples of scenarios where GlueStick may still fail or underperform. First, in scenes with repeated patterns, GlueStick is able to find a consistent matching, but can be displaced by one pattern if there is no additional hint to disambiguate the transform between the two images. This is for example the case on 7Scenes Stairs~\cite{7scenes}, when the camera is only seeing several steps, and it is unclear which step should be matching with which one in the other image.

Secondly, GlueStick has not been trained for large rotations beyond $45^{\circ}$ and often fails in these scenarios. The pre-training with homographies was done with rotations lower than $45^{\circ}$, and real viewpoint changes are rarely with such rotations. Nonetheless, a simple fix is to rotate one of the two images by $0^{\circ}, 90^{\circ}, 180^{\circ}$ and $270^{\circ}$, match it with the other image, and keep the best matching among the four.

Finally, a challenging scenario happens when the images have low texture in combination with symmetric structures. The former makes visual descriptors less reliable, while the latter makes it harder to disambiguate matches from the spatial context. The performance is then degraded in such situations. Having access to sequential data and feature tracking  may help solving such cases.

\section{Attention visualization}
\label{sec:attention_visualization}
\begin{figure*}
    \centering
    \small
    \begin{tabular}{cc}
        Attention from a keypoint & Attention from a line endpoint \\
        \includegraphics[width=0.48\textwidth]{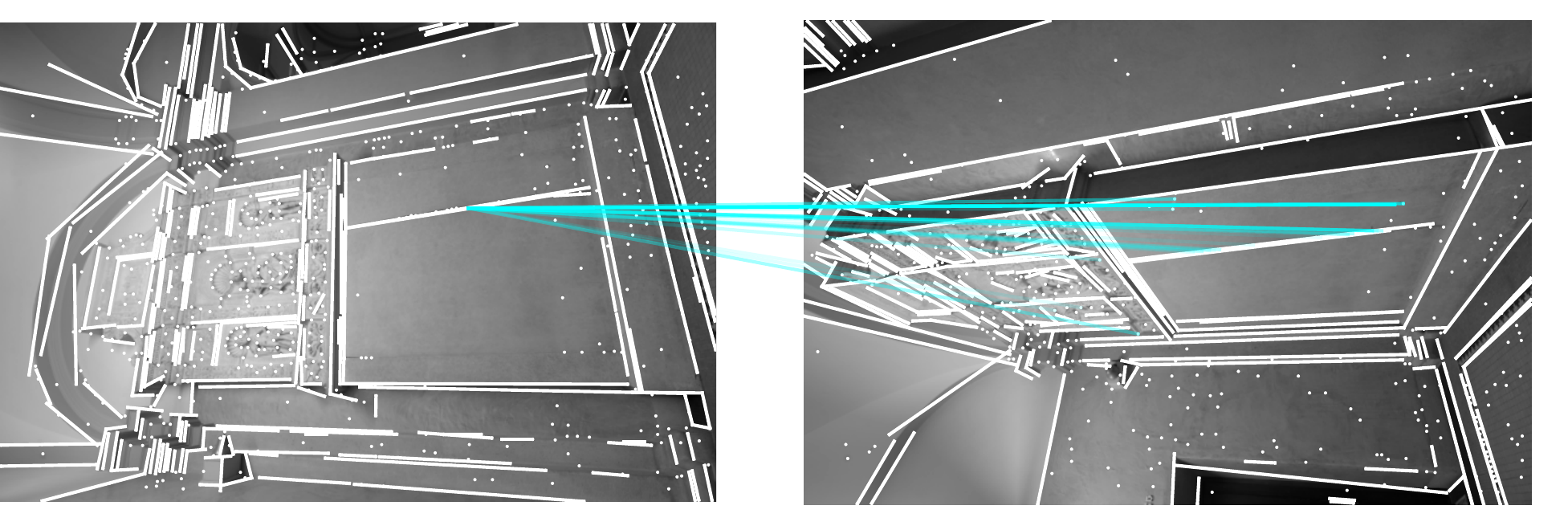}
        & \includegraphics[width=0.48\textwidth]{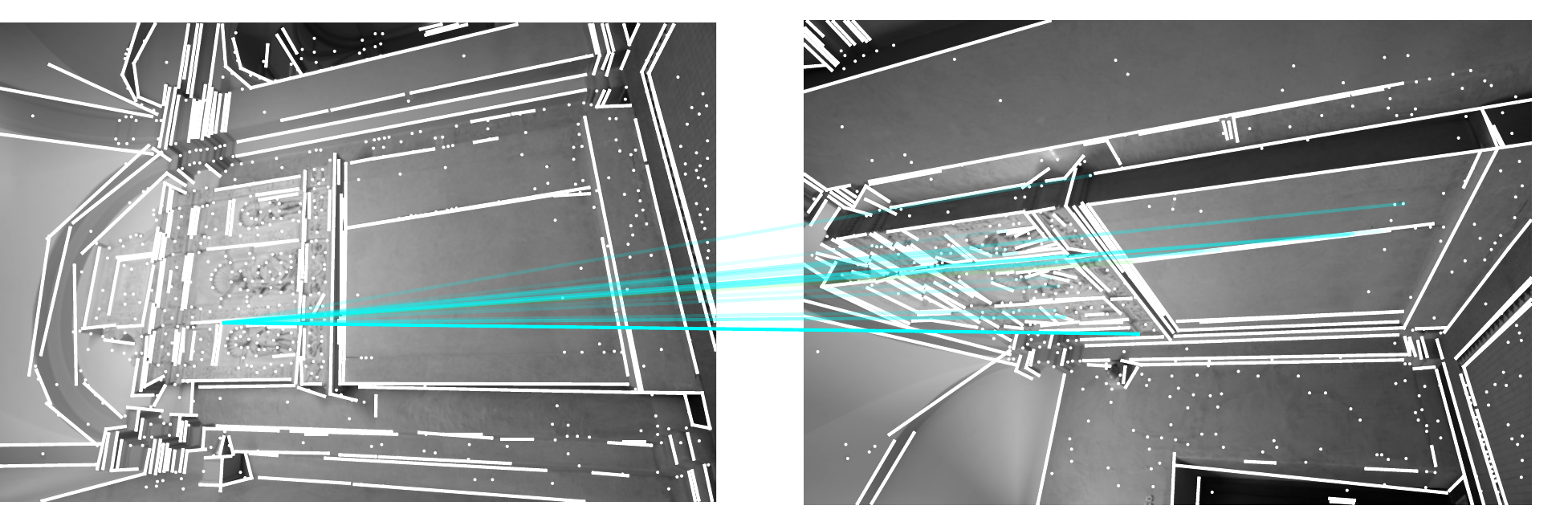} \\
        \multicolumn{2}{c}{Cross layer 2} \vspace{0.2cm} \\
        \includegraphics[width=0.48\textwidth]{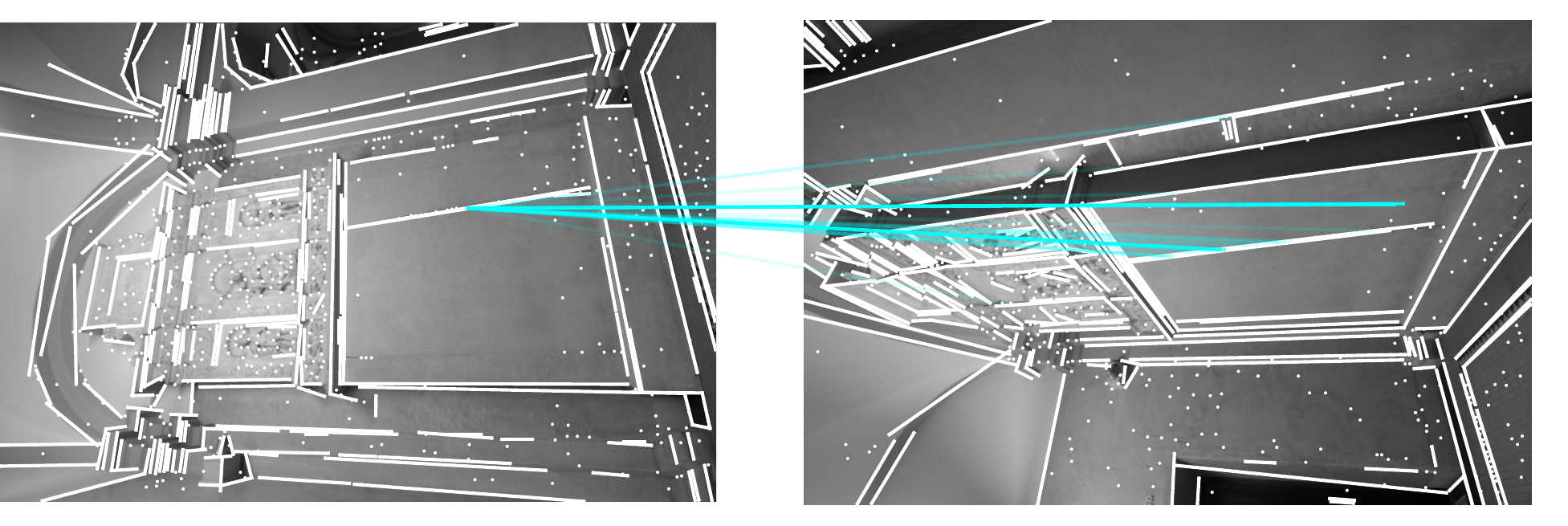}
        & \includegraphics[width=0.48\textwidth]{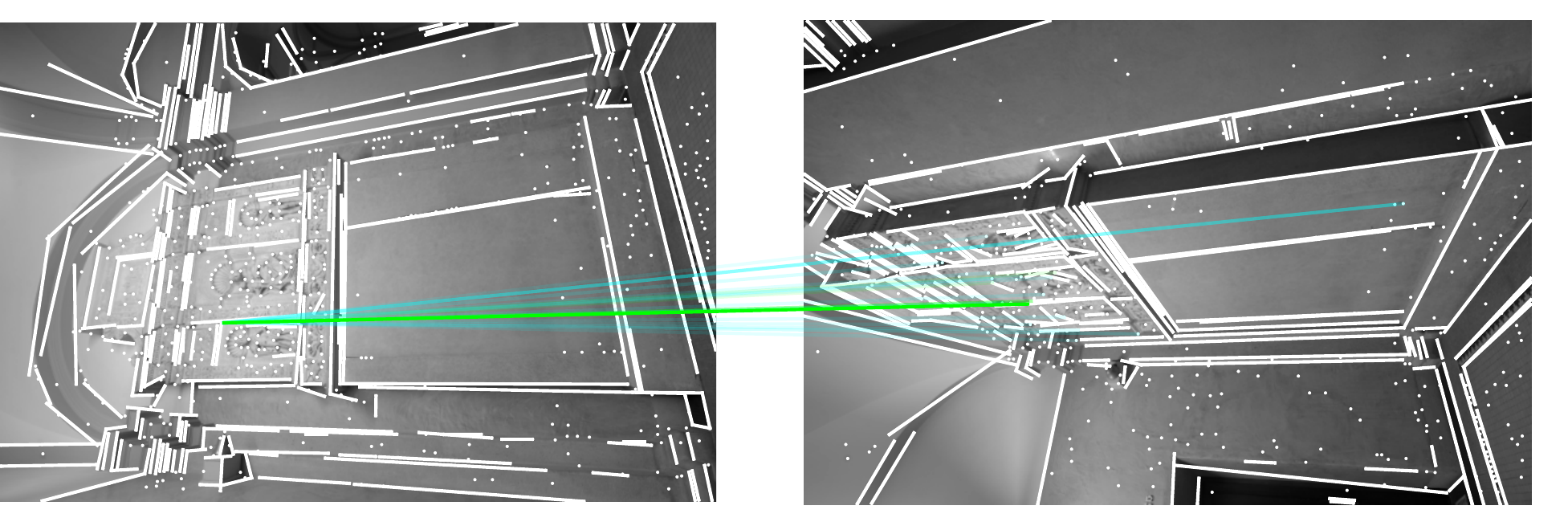} \\
        \multicolumn{2}{c}{Cross layer 4} \vspace{0.2cm} \\
        \includegraphics[width=0.48\textwidth]{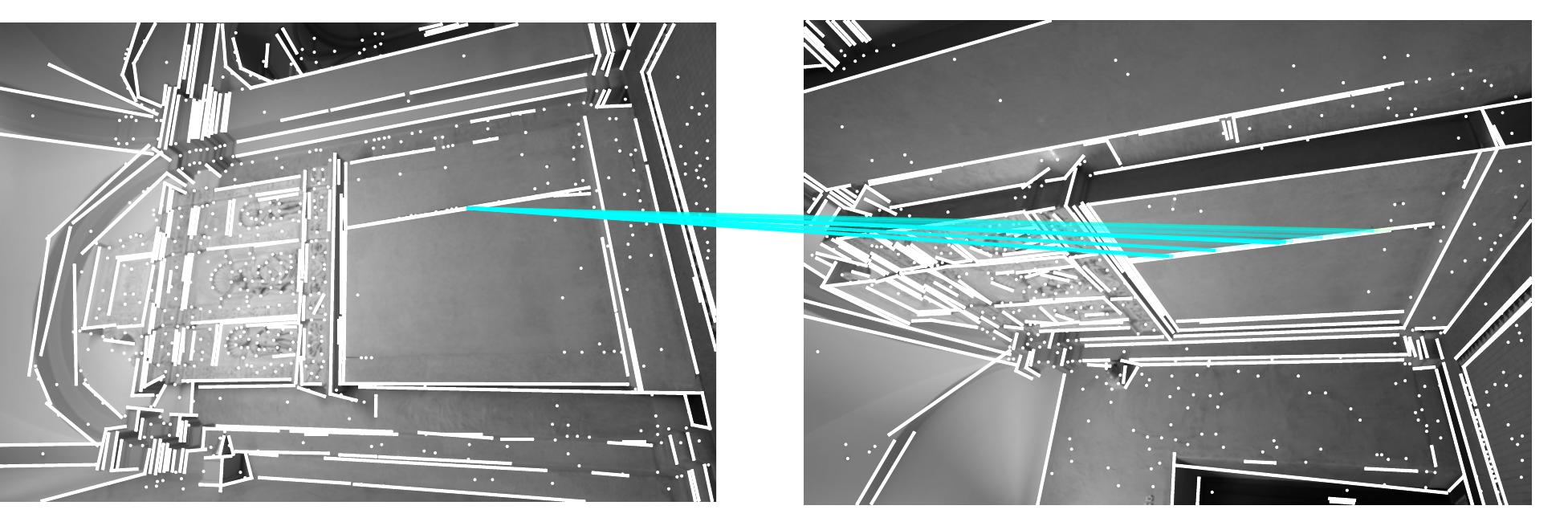}
        & \includegraphics[width=0.48\textwidth]{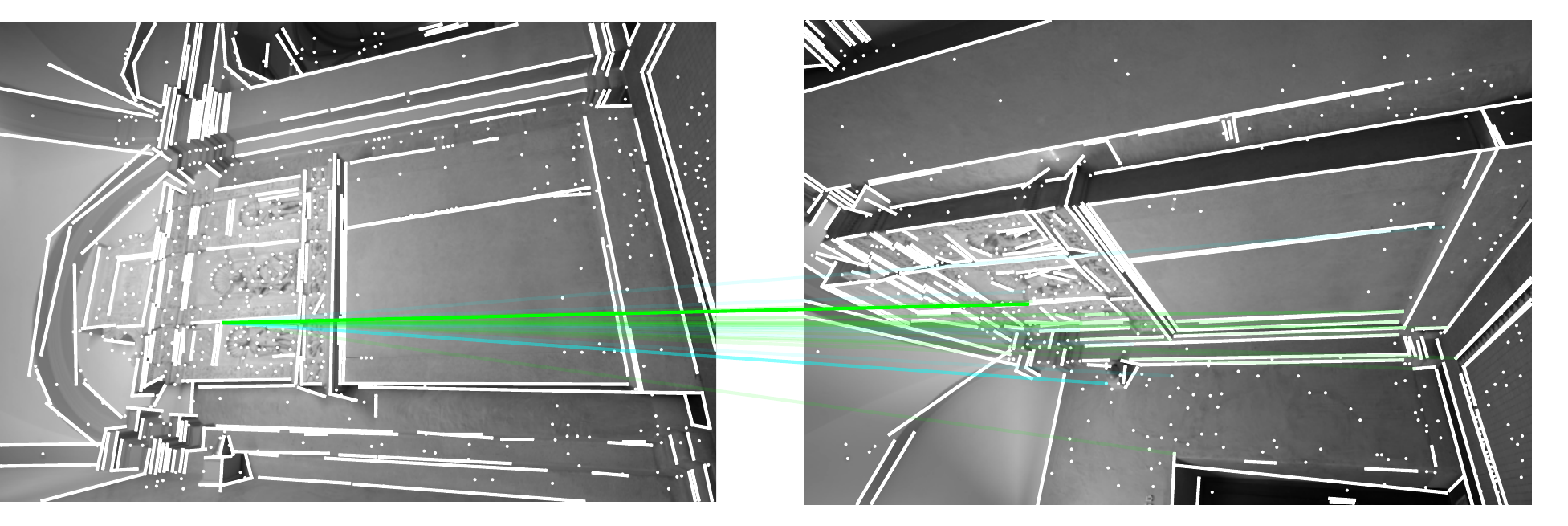} \\
        \multicolumn{2}{c}{Cross layer 6} \vspace{0.2cm} \\
        \includegraphics[width=0.48\textwidth]{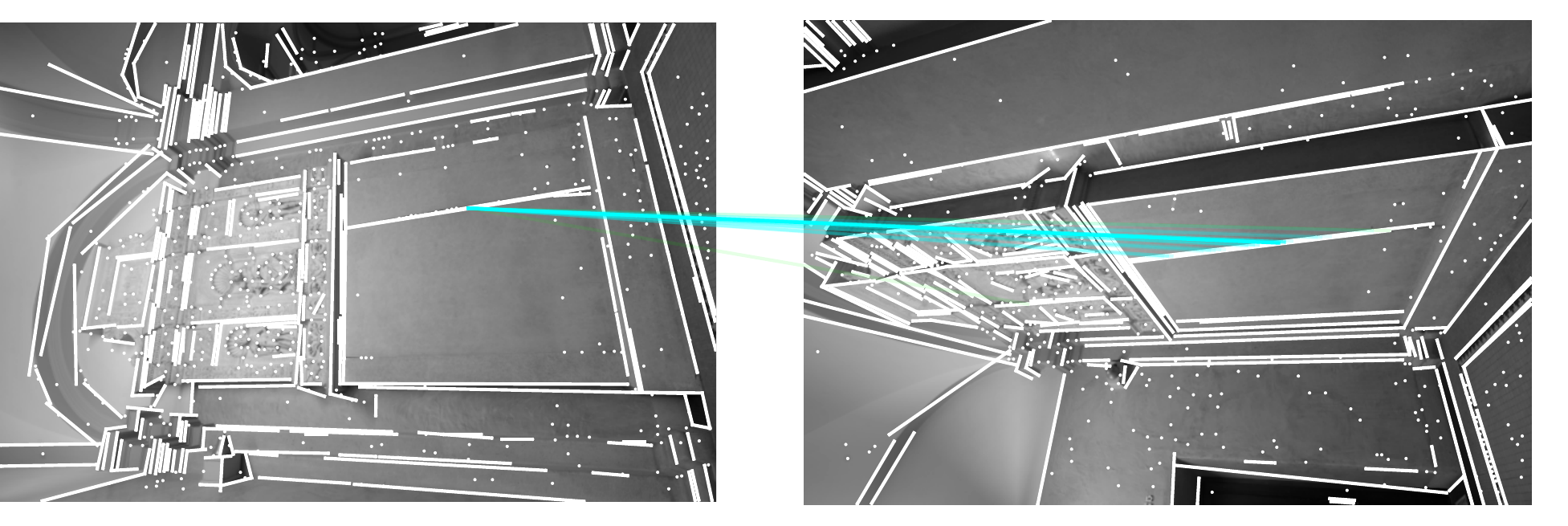}
        & \includegraphics[width=0.48\textwidth]{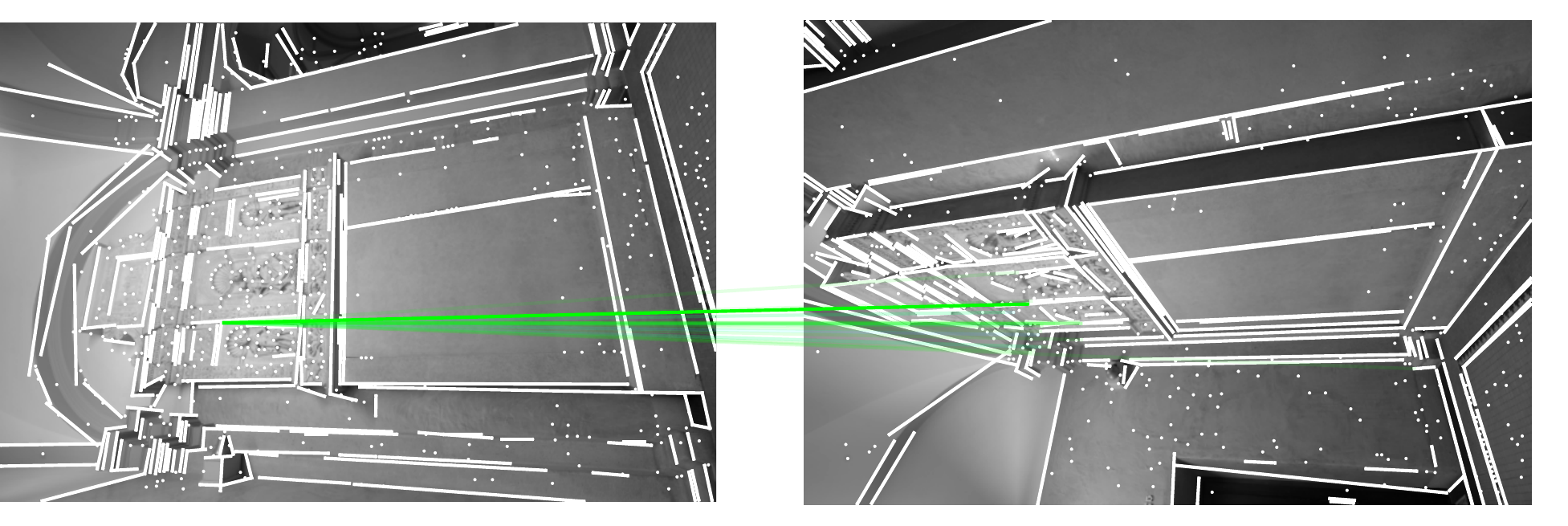} \\
        \multicolumn{2}{c}{Cross layer 8} \vspace{0.25cm} \\
        \includegraphics[width=0.47\textwidth]{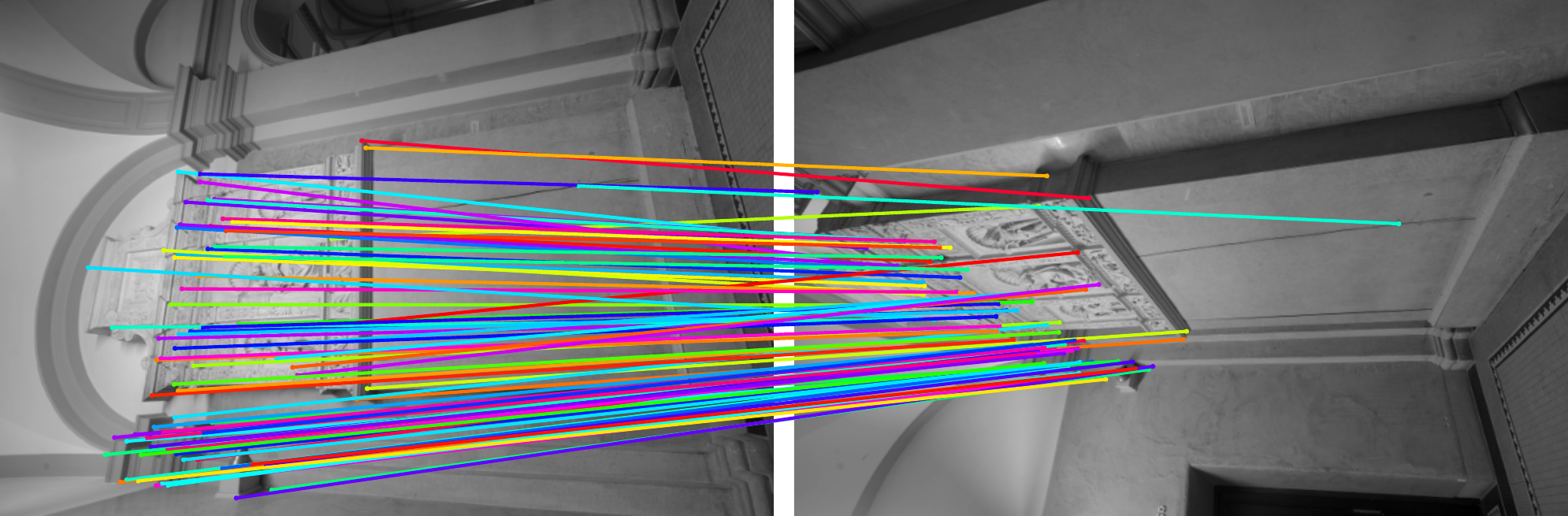}
        & \includegraphics[width=0.47\textwidth]{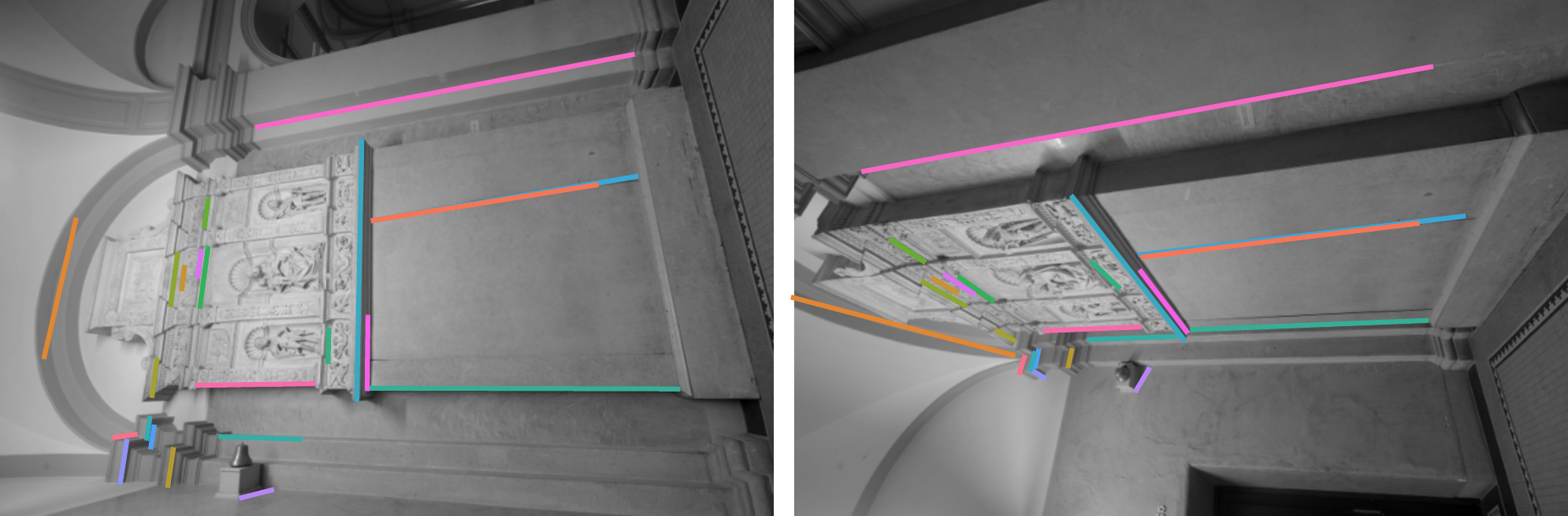} \\
        Point matches & Line matches
    \end{tabular}
    \caption{\textbf{Cross attention visualization.} We plot in the first column the attention from a keypoint and in the right column the attention of a line endpoint, for various layers in the Graph Neural Network. We compute here the average attention across all heads and keep the top 20 activated nodes. More opaque lines means higher attention, green matches are connected to a line endpoint in the second image, and cyan matches are connected to an isolated keypoint. The last row pictures the final matches.}
    \label{fig:attention_visualization}
\end{figure*}

We display the attention for some nodes in \cref{fig:attention_visualization}. This visualization is obtained by taking the attention matrix at various cross layers, averaging it across all heads, and taking the top 20 activated nodes. Green lines are used for nodes with connectivity greater than 0 (i.e. line endpoints), and cyan for nodes that are isolated keypoints. It can be seen from the left column that keypoint attention is leveraging the line structure to look for the right points along the line. In the right column, we can see that line endpoints can benefit from both keypoint and line endpoint attention. The attention is initially looking broadly at the image, before gradually focusing on the corresponding node in the other image. Thus, both points and line endpoints can complement each other to disambiguate the matching process.

~

~

~

{\small
\bibliographystyle{ieee_fullname}
\bibliography{arxiv}
}

\end{document}